%% file: cvpr.tex
\documentclass[10pt,twocolumn,letterpaper]{article}

\usepackage[pagenumbers]{cvpr} 

\usepackage{graphicx}
\usepackage{amsmath}
\usepackage{amssymb}
\usepackage{booktabs}
\usepackage{bm}
\usepackage{multirow}
\usepackage{xcolor}
\usepackage[export]{adjustbox}
\usepackage{makecell}
\usepackage{enumitem}

\definecolor{CLcolor}{rgb}{0.0, 0.8, 0.75}
\newcommand{\CL}[1]{\textcolor{CLcolor}{\textbf{CL}:~#1}}

\usepackage[pagebackref,breaklinks,colorlinks]{hyperref}

\usepackage[capitalize]{cleveref}
\crefname{section}{Sec.}{Secs.}
\Crefname{section}{Section}{Sections}
\Crefname{table}{Table}{Tables}
\crefname{table}{Tab.}{Tabs.}

\def\cvprPaperID{3972} 
\def\confName{CVPR}
\def\confYear{2022}

\begin{document}

\title{HVH: Learning a Hybrid Neural Volumetric Representation for\\Dynamic Hair Performance Capture}

\author{
Ziyan Wang$^{1,3}$~~~~~
Giljoo Nam$^{3}$~~~~~
Tuur Stuyck$^{3}$~~~~~ \\
Stephen Lombardi$^{3}$~~~~~
Michael Zollhöfer$^{3}$~~~~~
Jessica Hodgins$^{1,2}$~~~~~
Christoph Lassner$^{3}$
\vspace{0.1cm} \\ 
$^{1}$Carnegie Mellon University~~~
$^{2}$Meta AI~~~
$^{3}$Reality Labs Research
}

\include{defs}

\maketitle

\begin{abstract}

Capturing and rendering life-like hair is particularly challenging due to its fine geometric structure, the complex physical interaction and its non-trivial visual appearance. Yet, hair is a critical component for believable avatars.
In this paper, we address the aforementioned problems: 1)~we use a novel, volumetric hair representation that is composed of thousands of primitives. Each primitive can be rendered efficiently, yet realistically, by building on the latest advances in neural rendering. 2)~To have a reliable control signal, we present a novel way of tracking hair on the strand level. To keep the computational effort manageable, we use guide hairs and classic techniques to expand those into a dense hood of hair. 3)~To better enforce temporal consistency and generalization ability of our model, we further optimize the 3D scene flow of our representation with multiview optical flow, using volumetric raymarching.
Our method can not only create realistic renders of recorded multi-view sequences, but also create renderings for new hair configurations by providing new control signals.
We compare our method with existing work on viewpoint synthesis and drivable animation and achieve state-of-the-art results.
Please check out our project website at \hyperlink{https://ziyanw1.github.io/hvh/}{https://ziyanw1.github.io/hvh/}.
\end{abstract}

\input{intro}
\input{related_work}

\input{method}
\input{exp}
\input{discuss}

{\small
\bibliographystyle{ieee_fullname}
\bibliography{egbib}
}

\newpage
\input{appendix}

\end{document}

%% file: defs.tex
 \newcommand{\ZW}[1]{{\color{blue}{\bf ZW: #1}}}
 
 \newcommand{\SL}[1]{{\color{olive}{\bf SL: #1}}}
 \newcommand{\GN}[1]{{\color{orange}{\bf GN: #1}}}
 
 \definecolor{tscolor}{RGB}{153,51,255}
 \newcommand{\TS}[1]{{\color{tscolor}{\bf TS: #1}}}

 \definecolor{mzcolor}{RGB}{255,50,00}
 \newcommand\MZ[1] {\textbf{\textcolor{mzcolor}{MZ: #1}}}

 \newcommand{\CR}[1]{#1}
 
 \newcommand{\JKH}[1]{{\color{blue}{\bf JKH: #1}}}

 \newcommand{\bp}{\mathbf{p}}
 \newcommand{\bc}{\mathbf{c}}
 \newcommand{\bft}{\mathbf{f}}
 \newcommand{\bz}{\mathbf{z}}
 \newcommand{\bv}{\mathbf{v}}
 \newcommand{\br}{\mathbf{r}}
 \newcommand{\bV}{\mathbf{V}}
 \newcommand{\bR}{\mathbb{R}}

 \newcommand{\mL}{\mathcal{L}}
 \newcommand{\mR}{\mathcal{R}}
 \newcommand{\plh}{\mkern-1.5mu{\times}\mkern-2mu}
 
 \newcommand{\wt}[1]{\widetilde{#1}}
 
 \newcommand{\ttt}[1]{\small\texttt{#1}}

 \newcommand{\IGNORE}[1]{}

%% file: intro.tex
\section{Introduction}

Although notable progress has been made towards the realism of human avatars, cephalic hair is still one of the hardest parts of the human body to capture and render: with usually more than a hundred-thousand components, with complex physical interaction among them and with complex interaction with light, which is extraordinarily hard to model. However, it is an important part of our appearance and identity: hair styles can convey everything from religious beliefs to mood or activity. Hence, hair is critically important to make virtual avatars believable and universally usable. 

Previous work on mesh based representations~\cite{steve_meshvae, tewari2017mofa, tran2018nonlinear3dmm, li2017flame, xiang2020monoclothcap, saito2021scanimate, bagautdinov2021driving} has shown promising results on modeling the face and skin. However, they suffer when modeling hair, because meshes are not well suited for representing hair geometry. Recent volumetric representations~\cite{steve_nvs, mildenhall2020nerf} have high DoF which allows modeling of a changing geometric structure. They have achieved impressive results in 3D scene acquisition and rendering from multi-view photometric information. Compared to other geometric representations like multi-plane images~\cite{szeliski1998stereo, zhou2018mpi, mildenhall2019localmpi, attal2020matryodshka, broxton2020immersive} or point-based representations~\cite{aliev2020npr, wiles2020synsin, meshry2019neural, Lassner_pulsar, ruckert2021adop}, volumetric representations support a larger range of camera motion for view extrapolation and do not suffer from holes when rendering dynamic geometry like point-based representations. Furthermore, they can be learned from multi-view RGB data using differentiable volumetric ray marching, without additional MVS methods.

However, one major flaw of volumetric representations is their cubic memory complexity. 
%
This problem is particularly significant for hair, where high resolution is a requirement.
NeRF~\cite{mildenhall2020nerf} circumvents the $O(n^3)$ memory complexity problem by parameterizing a volumetric radiance field using an MLP. Given the implicit form, the MLP-based implicit function is not limited by spatial resolution.
A hierarchical structure with a coarse and fine level radiance function is used and an importance resampling based on the coarse level radiance field is utilized for boosting sample resolution. Although promising empirical results have been shown, they come with at the advance of high rendering time and the quality is still limited by the coarse level sampling resolution. Another limitation of NeRFs is that they were initially designed for static scenes. There is some recent work~\cite{tretschk2021nrnerf, park2021nerfies, li2021nsff, li2021neural, xian2021space, pumarola2021dnerf, Wang_2021_nvnerf, park2021hypernerf, yuan2021star} that extends the original NeRF concept to modeling dynamic scenes. However, they are still limited to relatively small motions, do not support drivable animation or are not efficient for rendering.


We present a hybrid representation: by using many volumetric primitives, we focus the resolution of the model onto the relevant regions of the 3D space. For each of the volumes, we construct a neural representation that captures the local appearance of the hair in great detail, similar to~\cite{liu2020nsvf, Wang_2021_nvnerf, kilonerf, steve_mvp} . However, without explicitly modeling the dynamics and structure of hair, it would be hard for the model to learn these properties solely through the indirect supervision of the multi-view appearance.
Given that the model learns to position primitives in an unsupervised manner, the model is also prone to overfitting as a result of not incorporating any temporal consistency during training. We address the problem of spatio-temporal modeling of dynamic upper head and hair by explicitly modeling hair dynamics at the coarse level and by enforcing temporal consistency of the model by multi-view optical flow at the fine level.

Procedurally, we first perform hair strand tracking at a coarse level by lifting multi-view optical flow to a 3D scene flow. To constrain the hair geometry and reduce the impact of the noise in multi-view optical flow, we also make sure the tracked hair strands preserve geometric properties like shape, length and curvature across time. As a second step, we attach volumes to hair strands to model the dynamic scene which can be optimized using differentiable volumetric raymarching. The volumes that are attached to the hair strands are regressed using a decoder that takes per-hair-strand features and a global latent code as input and is aware of the hair specific structure. Additionally, we further enforce fine 3D flow consistency by rendering the 3D scene flow of our model into 2D and compare it with the corresponding ground truth optical flow. This step is essential for making the model generalize better to unseen motions.
To summarize, the contributions of this work are
\begin{itemize}
    \item A hybrid neural volumetric representation that binds volumes to guide hair strands for hair performance capture.
    \item A hair tracking algorithm that utilizes multiview optical flow and per-frame hair strand reconstruction while preserving specific geometric properties like hair strand length and curvature.
    \item A volumetric ray marching algorithm on 3D scene flow which enables optimization of the position and orientation of each volumetric primitive through multiview 2D optical flow. 
    \item A hair specific volumetric decoder for hair volume regression and with awareness of hair structure. 
\end{itemize}


%% file: related_work.tex
\section{Related Work}
In this section, we discuss the most closely related classical hair dynamic and shape modeling methods.
We then discuss learning-based approaches that use either volumetric or non-volumetric scene representations for spatio-temporal modeling.

\IGNORE{
\noindent\textbf{Image-based Hair Modeling} has long been an interesting and challenging problem due to complicated hair geometry, a massive number of hair strands, severe self-occlusion and collisions, as well as view-dependent appearance. Comparing to tracking of other parts like the body~\cite{loper2015smpl, bagautdinov2021driving}, face~\cite{tran2018nonlinear3dmm, tewari2017mofa}, head~\cite{steve_meshvae, li2017flame}, hair is relatively hard to track as a result of its gigantic state space, it is hard to get a geometric template for tracking and self-occlusions are much more intense. Furthermore, the strands are usually dense, containing a lot of fine details unlike surfaces and there are quite a few works~\cite{nam2019lmvs, sun2021hairinverse, paris2008hair_photobooth, paris2004capture, wei2005modeling, luo2012multi, luo2013wide, luo2013structure, hu2014robust} that look into the hair geometry acquisition problem. 
\GN{A bit wordy. We can either remove this paragraph or move it to intro.}
}

\noindent \textbf{Image-based Hair Geometry Acquisition} is challenging due to the complicated hair geometry, massive number of strands, severe self occlusion and collision and view-dependent appearance. Paris \textit{et al.}~\cite{paris2008hair_photobooth, paris2004capture} and Wei \textit{et al.}~\cite{wei2005modeling} reconstruct 3D hair geometry from 2D/3D orientation fields using multi-view images. Luo \textit{et al.}~\cite{luo2012multi, luo2013wide} further improve the 3D reconstruction by refining the point cloud from traditional MVS with structure-aware aggregation and strand-based refinement. Luo \textit{et al.}~\cite{luo2013structure} and Hu \textit{et al.}~\cite{hu2014robust} progressively fit hair specific structures like ribbons and wisps to the point cloud. Recently, Nam \textit{et al.}~\cite{nam2019lmvs} substitute the plane assumption in the conventional MVS by a line-based structure to reconstruct 3D line clouds. Sun \textit{et al.}~\cite{sun2021hairinverse} use OLAT images for more efficient reconstruction of line-based MVS and develop an inverse rendering pipeline for hair that reasons about hair specific reflectance. However, none of those methods explicitly model temporal consistency for a time series capture. 

\noindent\textbf{Dynamic Hair Capture.}
Compared to the vast body of work on hair geometry acquisition, the work on hair dynamics~\cite{hu2017simulation, zhang2012simulation, xu2014dynamic, yang2019dynamic} acquisition is much less. Zhang \textit{et al.}~\cite{zhang2012simulation} uses hair simulation to enforce better temporal consistency over a per-frame hair reconstruction result. 
Hu \textit{et al.}~\cite{hu2017simulation} solves the physics parameters of a hair dynamics model by running parallel processes under different simulation parameters and adopting the one that best matches the visual observation. 
Xu \textit{et al.}~\cite{xu2014dynamic} performs visual tracking by aligning per-frame reconstruction of hair strands with motion paths of hair strands on a horizontal slice of a video volume. 
Yang \textit{et al.}~\cite{yang2019dynamic} developed a deep learning framework for hair tracking using indirect supervision from 2D hair segmentation and a digital 3D hair dataset. 
However those methods mainly focus on geometry modeling and are not photometrically accurate or do not support drivable animation.

\IGNORE{
\ZW{We probably also need to shorten this}
\noindent\textbf{Hair spatio-temporal modeling} has long been an interesting and challenging problem due to the complicated hair geometry, massive number of strands, severe self occlusion and collision and view dependent appearance. Comparing to tracking of other parts like body~\cite{loper2015smpl, bagautdinov2021driving}, face~\cite{tran2018nonlinear3dmm, tewari2017mofa}, head~\cite{steve_meshvae, li2017flame}, hair is relatively hard to track as a result of its gigantic state space, hard to get a geometric template for tracking and the self-occlusion are much more intense. Further more, the strands are usually dense, containing a lot of fine details unlike surfaces and there are quite a few works~\cite{nam2019lmvs, sun2021hairinverse, paris2008hair_photobooth, paris2004capture, wei2005modeling, luo2012multi, luo2013wide, luo2013structure, hu2014robust} that looks into the hair geometry acquisition problem. Paris \textit{et al.}~\cite{paris2008hair_photobooth, paris2004capture} and Wei \textit{et al.}~\cite{wei2005modeling} extract 2D/3D orientation field from multi-view RGB images with either controlled illumination or visual hull. Luo \textit{et al.}~\cite{luo2012multi, luo2013wide} further improve the 3D reconstruction of hair by refine the point cloud from traditional MVS with hair structure aware aggregation and strand-based refinement. Luo \textit{et al.}~\cite{luo2013structure} and Hu \textit{et al.}~\cite{hu2014robust} then progressively fit hair specific structure like ribbons and wisps to the point cloud. Recent, Nam \textit{et al.}~\cite{nam2019lmvs} substitute the plane assumption in the conventional MVS by line-based structure to reconstruct 3D line clouds. Sun \textit{et al.}~\cite{sun2021hairinverse} uses OLAT images for more efficient reconstruction of line-based MVS and develops an inverse rendering pipeline of hair that reasons about hair specific reflectance. Comparing to the vast body of work about hair geometry acquisition, the works on hair dynamics~\cite{hu2017simulation, zhang2012simulation, xu2014dynamic, yang2019dynamic} acquisition problem are relatively fewer. Zhang \textit{et al.}~\cite{zhang2012simulation} uses hair simulation to enforce better temporal consistency over a perframe hair reconstruction result. However, the simulation parameters are empirically determined and no hair collision is considered. Hu \textit{et al.}~\cite{hu2017simulation} solves the physics parameters of a hair dynamic model by running parallel processes under different simulation parameters and take the one that best matches the visual observation. However reliable, the computation of this method is relatively heavy. \ZW{Not sure how to mention this. But whether it might be true that the efficiency and complexity is also limited by the size of the search space? Like if they are searching for too many parameters simultaneously, it take more time or even incredibly long to converge?} Xu \textit{et al.}~\cite{xu2014dynamic} performs visual tracking by aligning perframe reconstruction of hair strands by motion paths of hair strands on a horizontal slice of a video volume. However, this method only supports play back of a given video and don't have appearance modeling for hairs. Yang \textit{et al.}~\cite{yang2019dynamic} develops a deep learning framework for hair tracking using indirection supervision from 2D hair segmentation and a digital 3D hair dataset. But the results are not metrically accurate.
}

\noindent\textbf{Non-Volumetric Representations}
are widely studied in the literature of spatio-temporal modeling
Mesh-based representations~\cite{steve_meshvae, tewari2017mofa, tran2018nonlinear3dmm, li2017flame, xiang2020monoclothcap, bagautdinov2021driving} are a perfect fit for modeling surfaces and highly efficient to render.
However, they have limitations for modeling complex geometries like hair.
Multi-plane images~\cite{szeliski1998stereo, zhou2018mpi, mildenhall2019localmpi, attal2020matryodshka, broxton2020immersive} are good at modeling continuous shapes similar to volumetric representations, but are limited to a constrained set of viewing angles.
Point cloud representations~\cite{aliev2020npr, wiles2020synsin, meshry2019neural, Lassner_pulsar, ruckert2021adop} can model various geometries with high fidelity.
When used for appearance modeling, however, point-based representations might suffer from their innate sparseness which might result in holes.
Thus image-level rendering techniques~\cite{ronneberger2015unet} are often accompanied with such representations for completeness.

\noindent\textbf{Volumetric Representations}
are highly flexible and thus can model many different objects.
They are designed for geometric completeness given their dense grid-like structure.
Many previous works have demonstrated the strength of such representations in geometry modeling~\cite{choy20163dr2n2, wu20163dvolvae, kar2017lsm, tulsiani2017multi, zhu2017rethinking, NEURIPS2018_ziyan, tung2019learning}.
Some recent works~\cite{sitzmann2019deepvoxels, steve_nvs, peng2021neuralbody} have also shown their effectiveness in modeling appearance.
DeepVoxels~\cite{sitzmann2019deepvoxels} learn a 3D grid of features as the scene representation.
Neural volumes~\cite{steve_nvs} learns a grid of discrete color and density values via volumetric raymarching.
Neural body~\cite{peng2021neuralbody} incorporates SMPL~\cite{loper2015smpl} with Neural Volumes~\cite{steve_nvs} for body modeling.
Nevertheless, the rendering quality, efficiency and memory footprint of those volumetric representations is still limited by the voxel resolutions.
To conquer this major drawback of volumetric methods, MVP~\cite{steve_mvp} proposes a hybrid representation for efficient and high-fidelity rendering.
It attaches a set of local volumetric primitives to a tracked head mesh and employs a tailored volumetric raymarching algorithm that is developed for fast rendering via a BVH~\cite{karras2013fastbvh}.
The tracked mesh provides a good initialization for the positions and rotations of the primitives that are jointly learned.
Still, finding the globally optimal positions and rotations purely based on a photometric reconstruction loss is highly challenging due to many local minima in the energy formulation.

\noindent\textbf{Coordinate-based Representations}
have been the major focus of recent literature in 3D learning due to their low memory footprint and ability to dynamically assign the model capacity to the correct regions of 3D space.
Many works have demonstrated their ability to reconstruct high fidelity geometry~\cite{park2019deepsdf, saito2019pifu, chabra2020deep_local_shape, saito2020pifuhd, jiang2020local, genova2020local, mescheder2019occupancy, peng2020con} or to generate photo-realistic rendering results~\cite{sitzmann2019srn, niemeyer2020dvr, yariv2020idr, mildenhall2020nerf, liu2020nsvf, yu2021plenoctrees}. 
%
%
%
NeRF~\cite{mildenhall2020nerf} learns a volumetric radiance field of a static scene from multi-view photometric supervision using a differentiable raymarcher, but comes with a large rendering time. Several works~\cite{liu2020nsvf, yu2021plenoctrees, kilonerf, lindell2021autoint} have improved the rendering efficiency of NeRF on static scene.
%
%
Among all those approaches, the most related to ours are spatio-temporal modeling techniques ~\cite{tretschk2021nrnerf, park2021nerfies, li2021nsff, li2021neural, xian2021space, pumarola2021dnerf, Wang_2021_nvnerf, park2021hypernerf}.
Non-rigid NeRF~\cite{tretschk2021nrnerf}, D-NeRF~\cite{pumarola2021dnerf} and Nerfies~\cite{park2021nerfies} introduce a dynamic modeling framework with a canonical radiance field and per-frame warpings.
Some works~\cite{li2021neural, xian2021space, li2021nsff, Wang_2021_nvnerf, yuan2021star, wang2021ntf} model a 3D video by additionally conditioning the radiance field on temporally varying latent codes or an additional time index.
Xian \textit{et al.}~\cite{xian2021space} further leverages depth as an extra source of supervision.
STaR~\cite{yuan2021star} models scenes that consist of a background and one dynamic rigid object.
NSFF~\cite{li2021nsff} also combines a static and dynamic NeRF pipeline and uses optical flow to constrain the 3D scene flow derived from the NeRF model of adjacent time frames.
Wang \textit{et al.}~\cite{Wang_2021_nvnerf} introduce a grid of local animation codes for better generalization and improved rendering efficiency.
However, these methods are still limited by either sampling resolution or ability to model complex motions and do not generalize well to unseen motions.

%% file: method.tex
\section{Method}

\begin{figure*}[tb]
    \centering
    \includegraphics[width=1.0\textwidth]{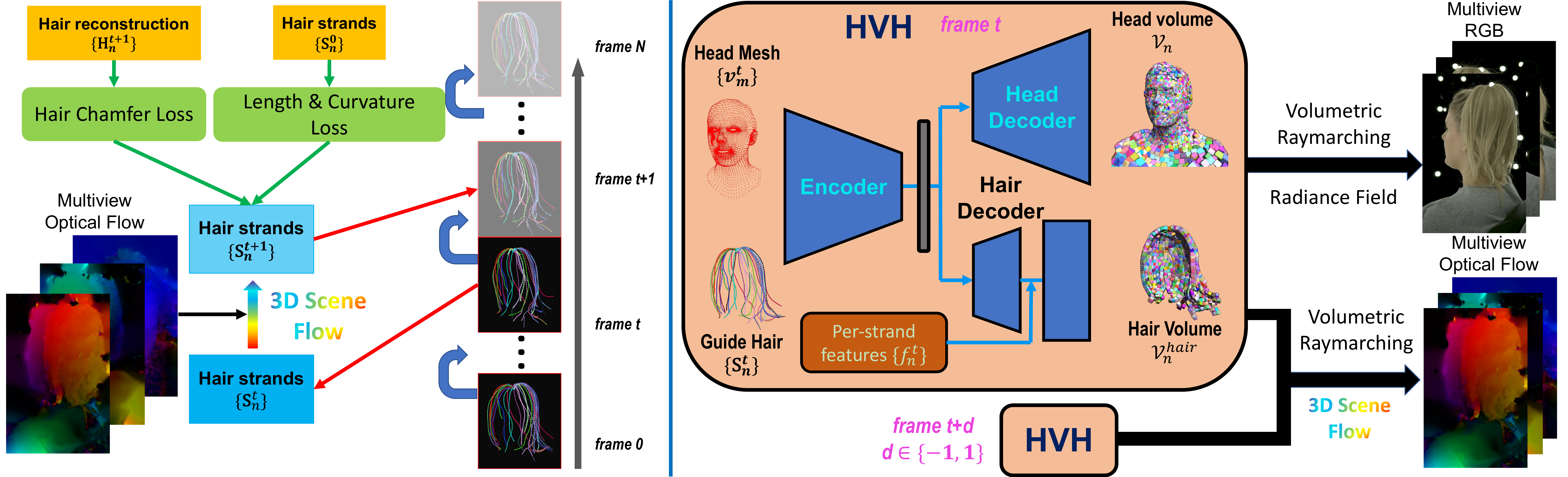}
    \caption{\label{fig:teaser}\textbf{Pipeline.} Our method consists of two stages: in the first stage, we perform guide hair tracking with multiview optical flow as well as per-frame hair reconstruction. In the second stage, we further amplify the sparse guide hair strands by attaching volumetric neural rendering primitives and optimizing them by using the multiview RGB and optical flow data. 
    }
    
\end{figure*}


In this section, we introduce our hybrid neural volumetric representation for hair performance capture. 
Our representation combines both, the drivability of guide hair strands and the completeness of volumetric primitives. Additionally, the guide hair strands serve as an efficient coarse level geometry for volumetric primitives to attach to, avoiding unnecessary computational expense on empty space. As a result of guide hair strand tracking as well as dense 3D scene flow refinement, our model is temporally consistent with better generalization over unseen motions.
As illustrated in Fig.~\ref{fig:teaser}, the whole pipeline contains two major steps which we will explain separately. 
In the first step, we perform strand-level tracking that leverages multi-view optical flow information and propagates information about a subset of tracked hair strands into future frames. To save computation time, we track only guide hairs instead of tracking all hair strands. This is a widely used technique in hair animation and simulation~\cite{iben2013artistic, petrovic2005volumetric, chai2016adaptive}, which leads to a significant boost in run time performance. 
However, getting the guide hairs tracked is not enough to model the hair motion and appearance or to animate all the hairs due to the sparseness of the guide hairs. To circumvent this, we combine it with a volumetric representation by attaching volumetric primitives to the nodes on the guide hairs. This hybrid representation has good localization of hairs in an explicit way and has full coverage of all the hairs, making use of the benefits of both representations. Another advantage is that the introduction of volumes allows optimizing hair shape and appearance by multi-view dense photometric information via differentiable volumetric ray marching.
In the second step, we use the attached volumetric primitives to model the hairs that are surrounding the guide hair strands to achieve dense hair appearance, shape and motion acquisition.
A hair specific volume decoder is designed for regressing those volumes, conditioning on both a global latent vector and hair strand feature vectors with hair structure awareness.
Additionally, we develop a volumetric raymarching algorithm for 3D scene flow that facilitates the learning from multi-view 2D optical flow. We show in the experiments that the introduction of additional optical flow supervision yields better temporal consistency and generalization of the model.

\subsection{Guide Hair Tracking}

We frame the guide hair tracking process as an optimization problem.
Given the guide hair strands and multi-view optical flow at the current frame $t$, we unproject and fuse optical flow under different camera poses into 3D flow and use that to infer the next possible position of the guide hairs at the next frame $t+1$. 

\noindent\textbf{Data Setup and Notation.} In our setting, we perform hair tracking using multi-view video data. We use a multi-camera system with around 100 synchronized color cameras that produces $2048\times1334$ resolution images at 30 Hz. The cameras are focused at the center of the capture system and distributed spherically at a distance of one meter to provide as many viewpoints as possible. Camera intrinsics and extrinsics are calibrated in an offline process. We  generate multi-view optical flow between adjacent frames for each camera, using the OpenCV~\cite{bradski2000opencv} implementation of~\cite{kroeger2016disof}. We acquire per-frame hair geometry by running~\cite{nam2019lmvs}. 
We parameterize guide hairs as connected point clouds. Given a specific hair strand $\mathbf S^t$ at time frame $t$, we denote the Euclidean coordinate of the $\mathbf{n}$th node on hair strand $\mathbf S^t$ as $\mathbf S^t_n$. Similarly, we have the future position of $\mathbf{S}^t_n$ at time frame $t+1$ as $\mathbf{S}^{t+1}_n$. Next we introduce the notations for multi-view camera related information. We denote $\Pi_i(\cdot)$ as the camera transformation matrix of camera $i$ which projects a 3D point into 2D image coordinate. We denote $\mathbf I_{of,i}$ and $\mathbf I_{d,i}$ as 2D matrix of optical flow and depth of camera $i$ respectively. We denote $\mathbf H^{t}_n$ as the reconstructed point cloud with direction from \cite{nam2019lmvs, sun2021hairinverse}. Unless otherwise stated, all bold lower case symbols denote vectors.

\noindent\textbf{Tracking Objectives.} Given camera $i$, we could project a 3D point into 2D to retrieve its 2D image index. The camera projection is defined as

\begin{align*}
    \mathbf{\hat{p}}^t_{s, i} = \begin{bmatrix} \mathbf{p}^t_{s,i} \\ \mathbf{1} \end{bmatrix} = \Pi_i(\mathbf{S}^t_n),
\end{align*}

\noindent where $\mathbf{\hat{p}}^t_{s, i}$ is the homogeneous coordinate of $\mathbf{p}^t_{s,i}$. Given the camera projection formulation, we formulate the first data-term objective based on optical flow as follows:

\begin{align*}
    \mathcal{L}_{of} &= \sum_{n,i} \mathbf{\omega}_{n,i} ||\mathbf{S}^{t+1}_n - \mathbf{Z}_i(\mathbf{S}^{t+1}_n)\Pi_i^{-1}(\mathbf{p}^t_{s,i}+\delta_{\mathbf{p}}) ||^2_2, \\
    \mathbf\omega_{n, i} &= exp(-\sigma||\mathbf{Z}_i(\mathbf{S}^{t}_n) - \mathbf{I}_{d,i}(\mathbf{p}^t_{s,i})||^2_2), \\
    \delta_{\mathbf{p}} &= \mathbf{I}_{of, i}(\mathbf{p}^t_{s,i}),
\end{align*}

\noindent where we denote $\mathbf{Z}_i(\cdot)$ as the function that represents the depth of a certain point under camera $i$ and $\omega_i$ serves as a weighting factor for view selection where a smaller value means larger mismatch of projected depth and real depth under the $i$th camera pose. We use a $\sigma=0.01$.

In parallel with the data-term objective on optical flow, we add another data-term objective to facilitate geometry preserved tracking, which compares the Chamfer distance between tracked guide hair strands and the per-frame hair reconstruction from \cite{nam2019lmvs}. This loss is designed to make sure that the guide hair geometry point cloud will not deviate too much from the true hair geometry. Unlike the conventional Chamfer loss, we also penalize the cosine distance between the directions of $\mathbf S^t_n$ and the direction of its closest $k=10$ neighbors as $\mathcal{H}(\mathbf S^{t+1}_n) \subsetneq \{\mathbf H^{t+1}_n\}$; the losses are defined as:
\begin{align*}
    \mathcal{L}_{hdir} &= \sum_{n, \mathbf h\in \mathcal{H}(\mathbf S^{t+1}_n)} \omega_{n,\mathbf h}^d (1-|\cos(\mathbf{dir}(\mathbf S^{t+1}_n), \mathbf{dir}(\mathbf h))|), \\
    \mathcal{L}_{hpos} &= \sum_{n, \mathbf h\in \mathcal{H}(\mathbf S^{t+1}_n)} \omega_{n,\mathbf h}^r ||\mathbf S^{t+1}_n-\mathbf h||^2_2, 
\end{align*}
where $\omega_{n,\mathbf h}^d=exp(-\sigma||\mathbf S^{t+1}_n-\mathbf h||^2_2)$ is a spatial weighting, $cos(\cdot, \cdot)$ is a cosine distance function between two vectors and
$\mathbf{dir}(\mathbf S^{t+1}_n)=\mathbf S^{t+1}_{n+1}-\mathbf S^{t+1}_n$ is a first order approximation of the hair direction at $\mathbf S^{t+1}_n$. $\omega_{n,\mathbf{h}}^r=cos(\mathbf{dir}(\mathbf S^{t+1}_n), \mathbf{dir}(\mathbf{h}))$ is a weighting factor that aims at describing the direction similarity between $\mathbf S^{t+1}_n$ and $\mathbf h$. With $\mathcal{L}_{hdir}$, we could groom the guide hairs $\mathbf S^{t+1}_n$ to have similar direction to its closest $k=10$ neighbors in $\mathcal{H}(\mathbf S^{t+1}_n)$, resulting in a more consistent guide hair direction distribution. Alternatively, $\mathcal{L}_{hpos}$ guarantees that the tracked guide hairs do not deviate too much from the reconstructed hair shapes.

However, with just the data-term loss, the tracked guide hairs might overfit to noise in the data terms. To prevent this, we further introduce several model-term objectives for hair shape regularization.

\begin{align*}
    \mathcal{L}_{len} =& \sum_n (||\mathbf{dir}(\mathbf S^{t+1}_n)||_2-||\mathbf{dir}(\mathbf S^{0}_n)||_2)^2, \\
       \mathcal{L}_{tang} =& \sum_n ((\mathbf S^{t+1}_{n+1} - \mathbf S^{t+1}_{n} - \mathbf S^{t}_{n+1} + \mathbf S^{t}_{n}) \cdot \mathbf{dir}(S^{t}_n))^2 + \\ 
    &((\mathbf S^{t}_{n+1} - \mathbf S^{t}_{n} - \mathbf S^{t+1}_{n+1} + \mathbf S^{t+1}_{n}) \cdot \mathbf{dir}(S^{t+1}_n))^2, \\
    \mathcal{L}_{cur} =& \sum_{n} (\mathbf{cur}(\mathbf S^{t+1}_{n}) - \mathbf{cur}(\mathbf S^{0}_{n})),
\end{align*}

\noindent where $\mathbf{cur}(\mathbf S^{t}_{n})$ is a numerical approximation of curvature at point $\mathbf S^{t}_{n}$ and is defined as:

\begin{align*}
    \sqrt{\frac{24(||\mathbf{dir}(\mathbf S^{t}_{n})||_2 + ||\mathbf{dir}(\mathbf S^{t}_{n})||_2 - ||\mathbf S^{t}_{n}-\mathbf S^{t}_{n+2}||_2)}{||\mathbf S^{t}_{n}-\mathbf S^{t}_{n+2}||_2^3}}.
\end{align*}

We optimize all loss terms together to solve $\{ \mathbf S^{t+1}_{n} \}$ given  $\{ \mathbf S^{t}_{n} \}$ with:

\begin{align*}
    \mathcal{L}_{hair} =& \mathcal{L}_{of} + \omega_{hdir}\mathcal{L}_{hdir} + \omega_{hpos}\mathcal{L}_{hpos} \\
    &+ \omega_{len}\mathcal{L}_{len} + \omega_{tang}\mathcal{L}_{tang} + \omega_{cur}\mathcal{L}_{cur}.
\end{align*}

By utilizing momentum information across the temporal axis, we can provide a better initialization of $\mathbf S^{t+1}_n$ given its trajectory and intialize $\mathbf S^{t+1}_n$ as

\begin{align*}
    \mathbf S^{t+1}_n = 3\mathbf S^{t}_n - 3\mathbf S^{t-1}_n + \mathbf S^{t-2}_n.
\end{align*}

\subsection{HVH}

\noindent\textbf{Background.} Similar to MVP, we define volumetric primitives $\mathcal{V}_{n}=\{ \mathbf t_n, \mathbf R_n, \mathbf s_n, \mathbf V_n \}$ to model a volume of local 3D space each, where $\mathbf R_n\in SO(3), \mathbf t_n\in \mathbb{R}^3$ describes the volume-to-world transformation, $\mathbf s_n\in \mathbb{R}^3$ are the per-axis scale factors and $\mathbf V_n=[\mathbf V_c, \mathbf V_\alpha]\in \mathbb{R}^{4\times M\times M\times M}$ is a volumetric grid that stores three channel color and opacity information. The volumes are placed on a UV-map that are unwrapped from a head tracked mesh and are regressed from a 2D CNN. Using an optimized BVH implementation, we can efficiently determine how the rays intersect each volume and find hit boxes. For each ray $\mathbf{r}_p(t) = \mathbf o_p + t\mathbf d_p$, we denote $(t_{min}, t_{max})$ as the start and end point for ray integration. Then, the differentiable aggregation of those volumetric primitives is defined as:

\begin{align*}
    \mathcal{I}_p &= \int_{t_{min}}^{t_{max}} \mathbf{V}_c(\mathbf{r}_p(t))\frac{dT(t)}{dt}dt, \\
    T(t) &= min(\int_{t_{min}}^{t} \mathbf{V}_\alpha(\mathbf{r}_p(t))dt , 1).
\end{align*}

\noindent We composite the rendered image as $\mathcal{\tilde{I}}_p=\mathcal{I}_p + (1-\mathcal{A}_p)I_{p,bg}$ where $\mathcal{A}_p=T(t_{max})$ and $I_{p,bg}$ is the background image. 

\noindent\textbf{Encoder.} The encoder uses the driving signal of a specific point in time and outputs a global latent code $\bm{\mathit{z}}\in \mathbb{R}^{256}$.
We use the tracked guide hairs $\{ \mathbf{S}_{n}^t \}$ and tracked head mesh vertices $\{ \bm{v}^t_m \}$ to define the driving signal. Symmetrically, we learn another decoder in parallel with the encoder in an auto-encoding way that regresses the tracked guide hairs $\{ \mathbf{S}_{n}^t \}$ and head mesh vertices $\{ \bm{v}^t_m \}$ from the global latent code $\bm{\mathit{z}}$. The architecture of the encoder is an MLP that regresses the parameter of a normal distribution $\mathcal{N}(\bm{\mu}, \bm{\sigma}), \bm{\mu},\bm{\sigma}\in\mathcal{R}^{256}$. We use the reparameterization trick from~\cite{kingma2013auto} to sample $\bm{\mathit{z}}$ from $\mathcal{N}(\bm{\mu}, \bm{\sigma})$ in a differentiable way.

\noindent\textbf{Hair Volume Decoder.}
Besides the volumes that are attached to the tracked mesh $\{ \bm{v}^t_m \}$, we define additional hair volume $\mathcal{V}_{n}^{hair}$ that are associated with guide hair nodes $\mathbf{S}^t_{n}$. The position $\mathbf t_{n}=\mathbf{\hat{t}}_n + \delta_{\mathbf t_{n}}$, orientation $\mathbf R_{n}=\delta_{\mathbf R_{n}}\cdot\mathbf{\hat{R}}_n$ and scale $\mathbf s_{n}=\mathbf{\hat{s}}_n + \delta_{\mathbf s_{n}}$ of each hair volume are determined by the base hair transformation $(\mathbf{\hat{t}}_n, \mathbf{\hat{R}}_n, \mathbf{\hat{s}}_n)$ and regressed hair relative transformation $(\delta_{\mathbf t_{n}}, \delta_{\mathbf R_{n}}, \delta_{\mathbf s_{n}})$. The base translation $\mathbf{\hat{t}}_n$ of each hair node is directly its position $\mathbf S_{n}^t$. The base rotation $\mathbf{\hat{R}}_n$ is derived from the hair tangential direction and the hair-head relative position. We denote $\tau_n$ as the hair tnagential direction at position $\mathbf S_{n}^t$ and $\nu'_n$ as the direction pointing to the tracked head center starting from $S_{n}^t$. Then, the base rotation is $\mathbf{\hat{R}}_n=[\tau_n^T; \rho_n^T; \nu_n^T]$, where $\rho_n=\tau_n\times\nu'_n, \nu_n=\rho_n\times\tau_n$.

The geometry of hair can not be simply described by a surface. Therefore, we design a 2D CNN that convolves along the hair growing direction and the rough hair spatial position separately. Specifically, in the each layer of the 2D CNN, we seperate a $k\times k$ filter into two $k\times 1$ and $1\times k$ filters and apply convolution along two orthogonal directions respectively, similar to~\cite{NEURIPS2019_bbf94b34}. To learn a more consistent hair shape and appearance model, we optimize per-strand hair features $\{ f_n^t \}$ that are shared across all time frames besides the temporally varying global latent code $\bm{\mathit{z}}$. For each node $\mathbf{S_n^t}$ on a hair strand $\mathbf{S}^t$, we assign an unique feature vector $f^t_n$. The shared per-strand hair features and the temporal varying latent code $\bm{\mathit{z}}$ are fused to serve as the input to the hair volume decoder, which is shown in Fig.~\ref{fig:hair_dec}.

\begin{figure}[htb]
    \centering
    \adjincludegraphics[width=0.45\textwidth, trim={0 0 0 {0.02\height}}, clip]{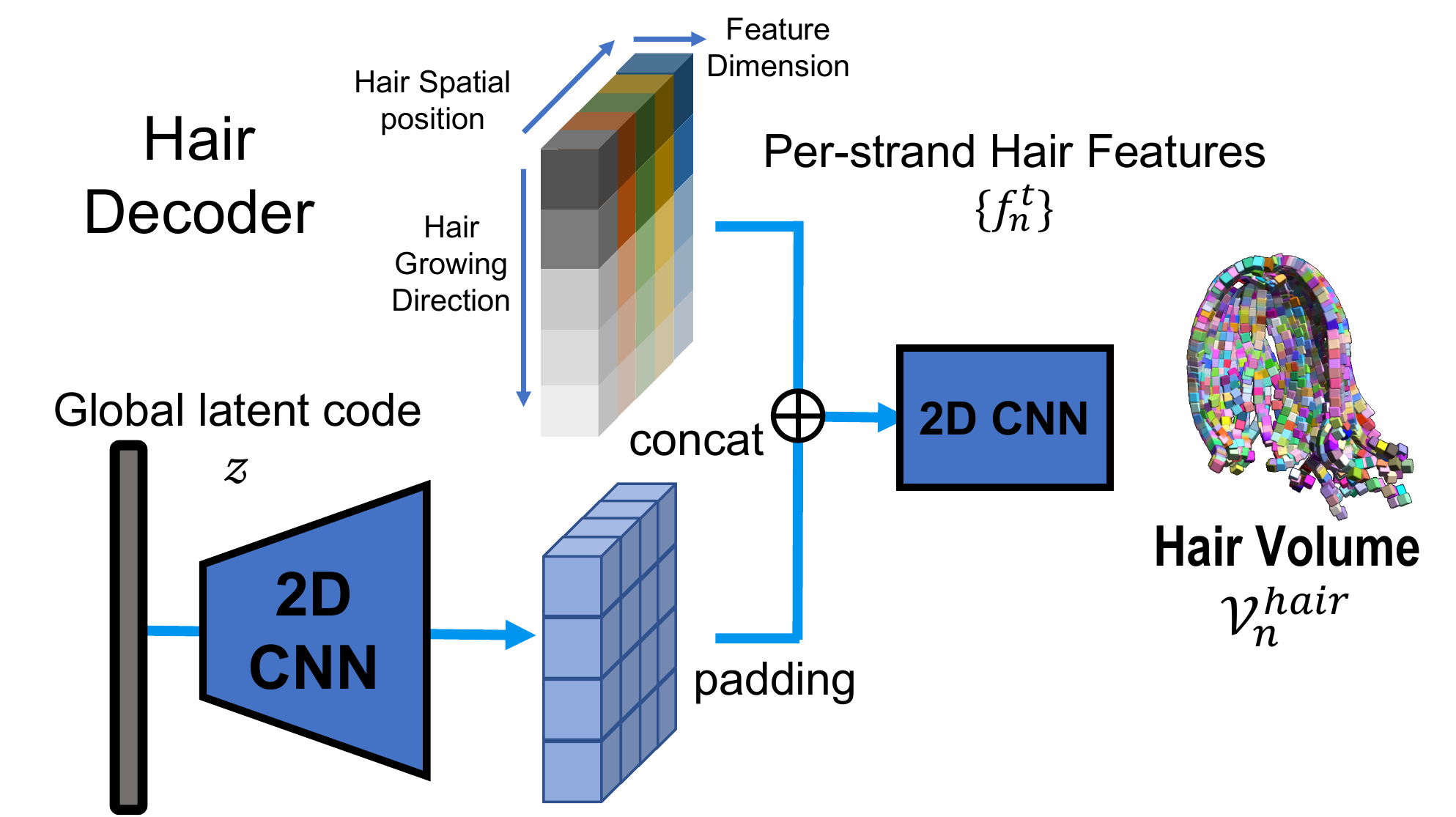}
    \caption{\label{fig:hair_dec}\textbf{Architecture of the hair decoder.} The hair decoder takes both the global latent code $z$ and the per-strand hair features $\{ f_n^t\}$ as inputs. $z$ is first deconvolved into a 2D feature tensor. It is then padded and concatenated with $\{ f_n^t\}$. In the following operation, the 2D convolution layers are applied along the hair growing direction and the hair spatial position seperately.}
    \label{fig:hair_dec}
\end{figure}

\noindent\textbf{Differentiable Volumetric Raymarching of 3D Scene Flow.} 
Learning a volumetric scene representation by multi-view photometric information is sufficient for high fidelity rendering and novel view synthesis. However, it is challenging for the model to reason about motion given the limited supervision and the results have poor temporal consistency, especially on unseen sequences. To better enforce temporal consistency, we develop a differentiable volumetric ray marching algorithm of 3D scene flow which enables training via multi-view 2D optical flow.

Given the transformations of each primitive as $(\mathbf t_n, \mathbf R_n, \mathbf s_n)$, we express the coordinate of each node on a volumetric grid at frame $u$ as $\mathbf{V}_{xyz}^{u}=\mathbf{s}_t\mathbf{R}_t\mathbf{V}_{tpl}+\mathbf{t}_n$, where $\mathbf{V}_{tpl}$ are the coordinates of a 3D mesh grid ranging between $[-1, 1]$. Given that the 3D scene flow from frame $u$ to $u+\delta$ can be expressed by each volumetric primitives as $\{ \mathbf{\delta V}_{xyz}^{u,u+\epsilon}=\mathbf{V}_{xyz}^{u+\epsilon} - \mathbf{V}_{xyz}^u \}$ and rendered into 2D flow as:

\begin{align*}
    \mathcal{I}_{p, flow}^{u, u+\delta} &= \int_{t_{min}}^{t_{max}} (\mathbf{\delta V}_{xyz}^{u,u+\epsilon}(\mathbf{r}_p(t)))\frac{dT(t)}{dt}dt, \\
    T(t) &= min(\int_{t_{min}}^{t_{max}} \mathbf{V}_\alpha^{u}(\mathbf{r}_p(t))dt , 1).
\end{align*}

\noindent\textbf{Training Objectives.} We train our model in an end-to-end manner with the following loss:

\begin{align*}
    \mathcal{L} =& \mathcal{L}_{pho} + \lambda_{flow}\mathcal{L}_{flow} + \lambda_{geo}\mathcal{L}_{geo} \\
    &+\lambda_{vol}\mathcal{L}_{vol} + \lambda_{cub}\mathcal{L}_{cub} + \lambda_{KL}\mathcal{L}_{KL}.
\end{align*}

\noindent The first term $\mathcal{L}_{pho}$ is the photometric loss that compares the difference between the rendered image $\mathcal{\tilde{I}}_p$ and ground truth image $I_{p}$ on all sampled pixels $p\in\mathcal{P}$,

\begin{align*}
    \mathcal{L}_{pho} = \sum_{p\in\mathcal{P}} ||I_{p,gt} - \mathcal{\tilde{I}}_p||^2_2.
\end{align*}

\noindent The second term $\mathcal{L}_{flow}$ aims to enforce temporal consistency of volumetric primitives from frame $u$ and its adjacent frame $u+\epsilon$ by minimizing the projected 2D flow and ground truth optical flow $I_{p,flow}^{u,u+\epsilon}$,

\begin{align*}
    \mathcal{L}_{flow} = \sum_{p\in\mathcal{P}} \mathcal{A}_p||I_{p,of}^{u, u+\epsilon} - \mathcal{I}_{p, flow}^{u, u+\epsilon}||^2_2,
\end{align*}

\noindent where $\epsilon\in\{-1, 1\}$. It is important to note that we use $\mathcal{A}_p$ to mask out the background part and we do not back propagate the errors from $\mathcal{L}_{flow}$ to $\mathcal{A}_p$ in order to get rid of the background noise in optical flows. To better enforce hair and head primitives moving with the tracked head mesh and guide hair strands, $\mathcal{L}_{geo}$ is designed to measure the difference between the mesh/strand vertices and their corresponding regressed value.

\begin{align*}
    \mathcal{L}_{geo} = \sum_{n}||\bm{S}^t_n - \bm{S}^t_{n,gt}||^2_2 + \sum_m||\bm{v}^t_m - \bm{v}^t_{m, gt}||^2_2,
\end{align*}

\noindent where $\bm{S}_n^t$ and $\bm{v}_m^t$ are the coordinate of the $n$th node of the tracked guide hair and tracked head mesh at frame $t$ and the $X_{gt}$ denotes the corresponding ground truth value.

We also add several regularization terms to inform the layout of the volumetric primitives:

\begin{align*}
    \mathcal{L}_{vol} &= \sum_{i=1,\cdots,N_p} \prod_{j\in\{x,y,z\}} s_i^j, \\
    \mathcal{L}_{cub} &= \sum_{i=1,\cdots,N_p} ||max(s_i^x, s_i^y, s_i^z) - min(s_i^x, s_i^y, s_i^z)||,
\end{align*}

\noindent where $N_p$ stands for the total number of volumetric primitives and $s_i^x, s_i^y, s_i^z$ are the three entries of each volumetric primitive's scale $\bm{s}_j$. The two regularization terms aim to prevent each primitive from growing too big while preserving the aspect ratio so that they remain approximately cubic. The last term is the Kullback-Leibler divergence loss $\mathcal{L}_{KL}$ which makes the learnt distribution of latent code $\bm{z}$ smooth and enforces similarity with a normal distribution $\mathcal{N}(0, 1)$.

%% file: exp.tex
\section{Experiments}

\subsection{Dataset}

%

For each video recorded with our multi camera system, we hold out approximately a quarter of the time frames as test sequence and the rest as training sequence. The test sequence will be used for conducting test experiments of drivable animation. This results in roughly 300 frames for training sequence and 100 frames for testing sequence. Additionally, on the training sequence, we hold out 7 cameras that are distributed around the rear and side view of the head. The captured images are downsampled to $1024\times667$ resolution for training and testing. We train our model exclusively on the training portion of each sequence with $m=93$ training views.

\subsection{Novel View Synthesis}
\label{sec:nvs}

\begin{table*}[h!tb]
\centering
\begin{tabular}{cc}
\resizebox{\columnwidth}{!}{%
\begin{tabular}{r|ccc|ccc|ccc|}
\multicolumn{1}{l|}{\multirow{2}{*}{}} & \multicolumn{3}{c|}{Seq01}                                                       & \multicolumn{3}{c|}{Seq02}                                                       & \multicolumn{3}{c|}{Seq03}                                                       \\ \cline{2-10} 
\multicolumn{1}{l|}{}                  & \multicolumn{1}{c|}{MSE} & \multicolumn{1}{c|}{SSIM} & \multicolumn{1}{c|}{PSNR} & \multicolumn{1}{c|}{MSE} & \multicolumn{1}{c|}{SSIM} & \multicolumn{1}{c|}{PSNR} & \multicolumn{1}{c|}{MSE} & \multicolumn{1}{c|}{SSIM} & \multicolumn{1}{c|}{PSNR} \\ \hline
PFNeRF & 51.25 & 0.9269 & 31.16 & 103.41 & 0.8659 & 28.15 & 76.59 & 0.9000 & 29.50 \\
NSFF & 50.13 & 0.9346 & 31.21 & 90.06 & 0.8885 & 28.75 & 83.18 & 0.8936 & 29.1 \\
NRNeRF & 56.78 & 0.9231 & 30.78 & 132.16 & 0.8549 & 27.13 & 79.83 & 0.8987 & 29.33 \\ \hline
MVP & 47.54 & 0.9476 & 31.6 & 77.23 & 0.9088 & 29.62 & 73.78 & 0.9224 & 29.66 \\
Ours & \textbf{41.89} & \textbf{0.9543} & \textbf{32.17} & \textbf{59.84} & \textbf{0.9275} & \textbf{30.69} & \textbf{71.58} & \textbf{0.9314} & \textbf{29.81}                       
\end{tabular}
} & 
\resizebox{\columnwidth}{!}{%
\begin{tabular}{r|ccc|ccc|ccc|}
\multicolumn{1}{l|}{\multirow{2}{*}{}} & \multicolumn{3}{c|}{Seq01}                                                                             & \multicolumn{3}{c|}{Seq02}                                                                             & \multicolumn{3}{c|}{Seq03}                                                   \\ \cline{2-10} 
\multicolumn{1}{l|}{}                  & \multicolumn{1}{c|}{MSE}        & \multicolumn{1}{c|}{SSIM}         & PSNR                             & \multicolumn{1}{c|}{MSE}        & \multicolumn{1}{c|}{SSIM}         & PSNR                             & \multicolumn{1}{c|}{MSE} & \multicolumn{1}{c|}{SSIM} & PSNR                  \\ \hline
MVP & 47.54 & 0.9476 & 31.6 & 77.23 & 0.9088 & 29.62 & 73.78 & 0.9224 & 29.66 \\
MVP w/ $\mathcal{L}_{flow}$ & 46.49 & 0.9473 & 31.69 & 71.07 & 0.9107 & 29.93 & 75.13 & 0.9240 & 29.58 \\
Ours w/o $\mathcal{L}_{flow}$ & 43.82 & 0.9508 & 31.99 & 65.98 & 0.9186 & 30.27 & 69.97  & 0.9359  & 29.93  \\
Ours & \textbf{41.89} & \textbf{0.9543} & \textbf{32.17} & \textbf{59.84} & \textbf{0.9275} & \textbf{30.69} & \textbf{71.58} & \textbf{0.9314} & \textbf{29.81} \\ \hline
MVP & 75.68 & 0.9200 & 29.49 & 85.10 & 0.9039 & 29.62 & 83.76 & 0.9086 & 29.16 \\
MVP w/ $\mathcal{L}_{flow}$ & 67.86 & 0.9276 & 30.00 & 83.11 & 0.9037 & 29.93 & 80.96 & 0.9086 & 29.16 \\
Ours w/o $\mathcal{L}_{flow}$ & 71.90 & 0.9223 & 29.74 & 72.74 & 0.9137 & 30.27 & 78.34 & 0.9198  & 29.44  \\
Ours & \textbf{65.96} & \textbf{0.9280} & \textbf{30.09} & \textbf{67.75} & \textbf{0.9208} & \textbf{30.69} & \textbf{75.66} & \textbf{0.9222} & \textbf{29.57}     
\end{tabular}
}
\end{tabular}
\caption{\label{tab:nvs_train}\textbf{Novel view synthesis}. On the left, we compare our method with both NeRF stemmed methods like NSFF\cite{li2021nsff}, NRNeRF~\cite{tretschk2021nrnerf} and a per-frame NeRF(PFNeRF) baseline, and a volumetric method like MVP~\cite{steve_mvp}. On the right, we further compare our method and different variants of our methods with MVP on novel views of both seen (top) and unseen (bottom) sequences. 
}
\end{table*}

\begin{figure}[tb]
\setlength\tabcolsep{0pt}
\renewcommand{\arraystretch}{0}
\centering
\begin{tabular}{ccccc}
 \textbf{\scriptsize NSFF~\cite{li2021nsff}} &
 \textbf{\scriptsize NRNeRF~\cite{tretschk2021nrnerf}} &
 \textbf{\scriptsize MVP~\cite{steve_mvp}} & 
 \textbf{\scriptsize Ours} &
 \textbf{\scriptsize Ground Truth} \\
 \adjincludegraphics[width=0.1\textwidth, trim={0 {0.12\height} 0 0}, clip]{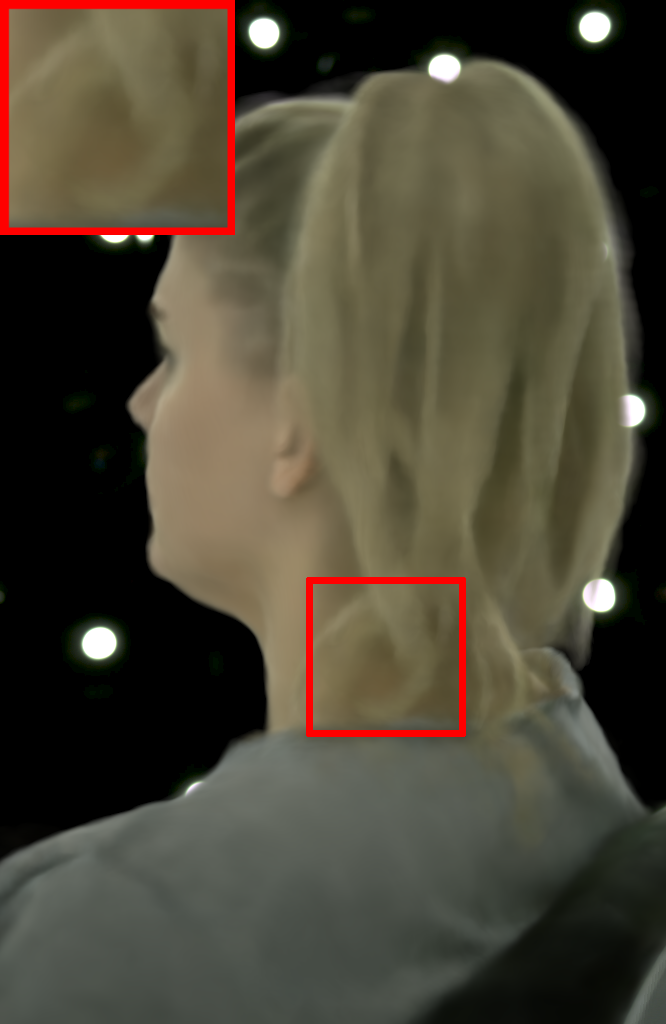} &
 \adjincludegraphics[width=0.1\textwidth, trim={0 {0.12\height} 0 0}, clip]{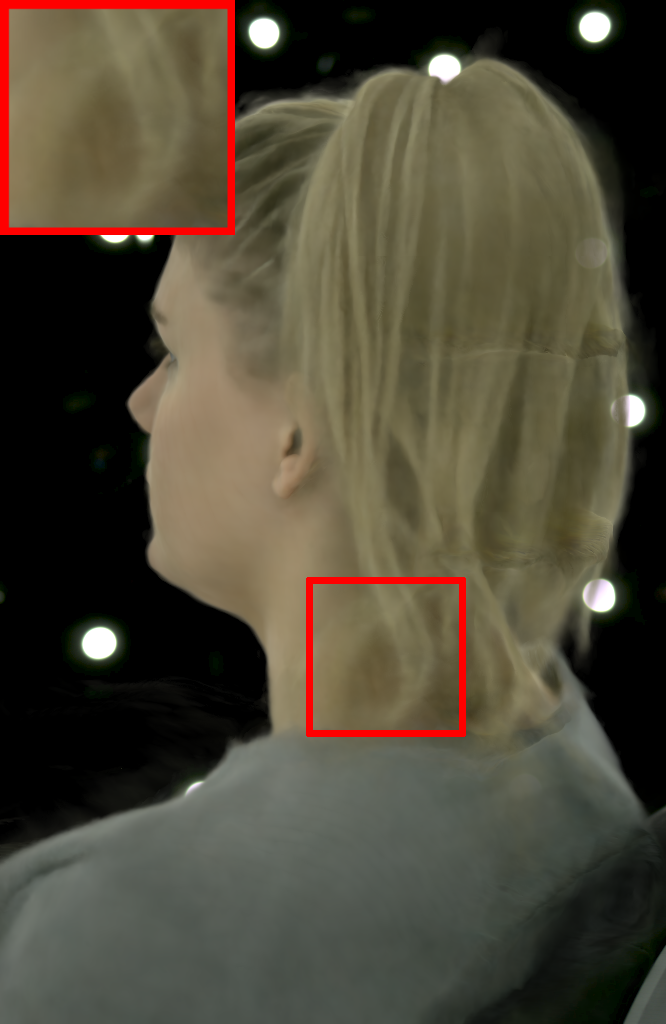} &
 \adjincludegraphics[width=0.1\textwidth, trim={0 {0.12\height} 0 0}, clip]{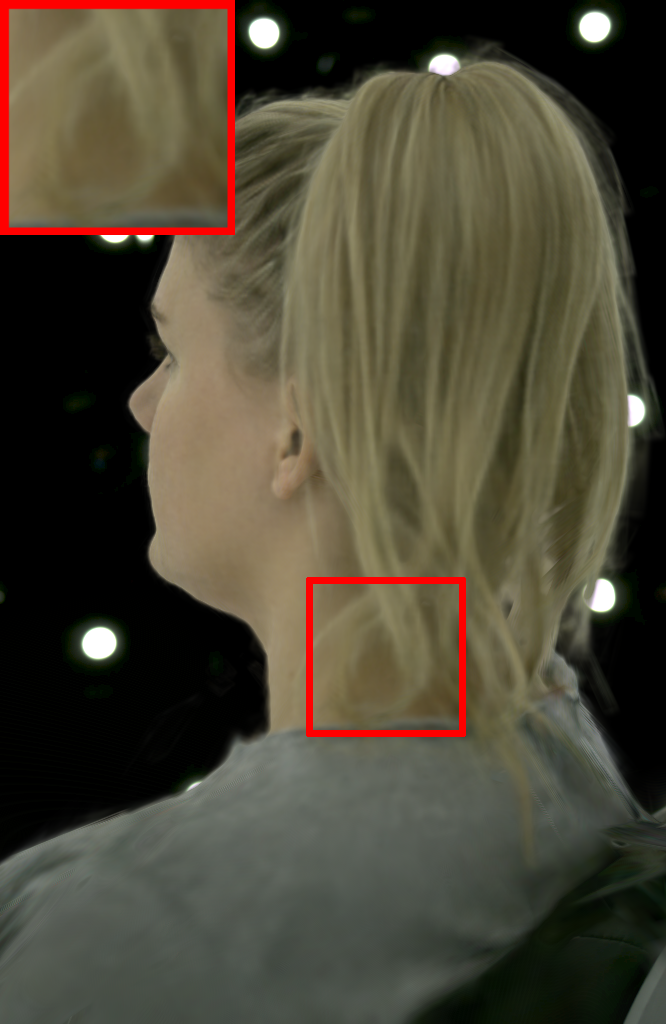} &
 \adjincludegraphics[width=0.1\textwidth, trim={0 {0.12\height} 0 0}, clip]{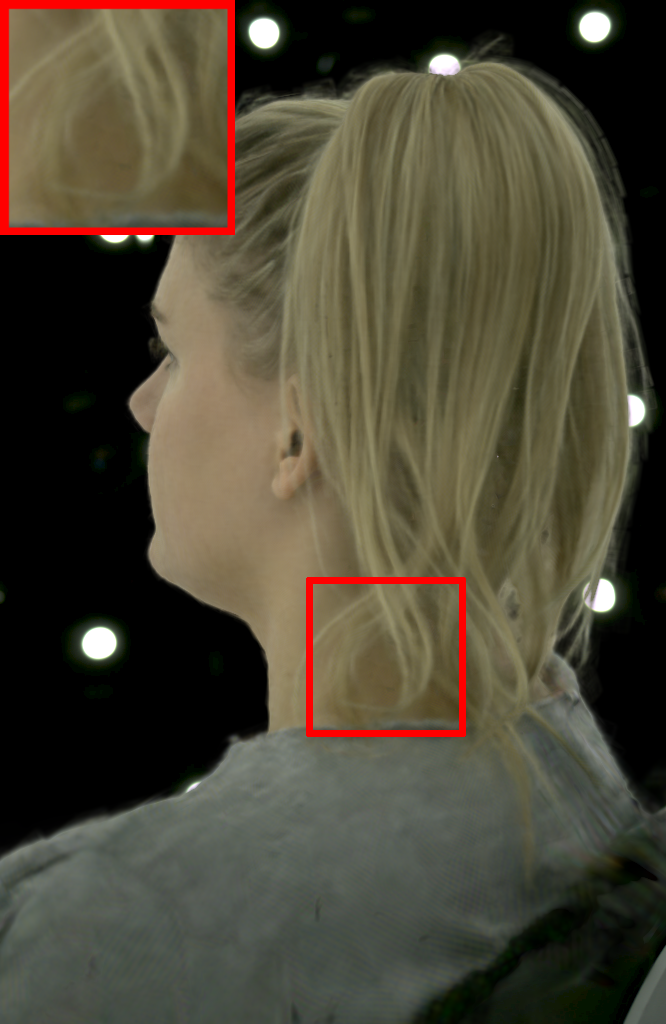} &
 \adjincludegraphics[width=0.1\textwidth, trim={0 {0.12\height} 0 0}, clip]{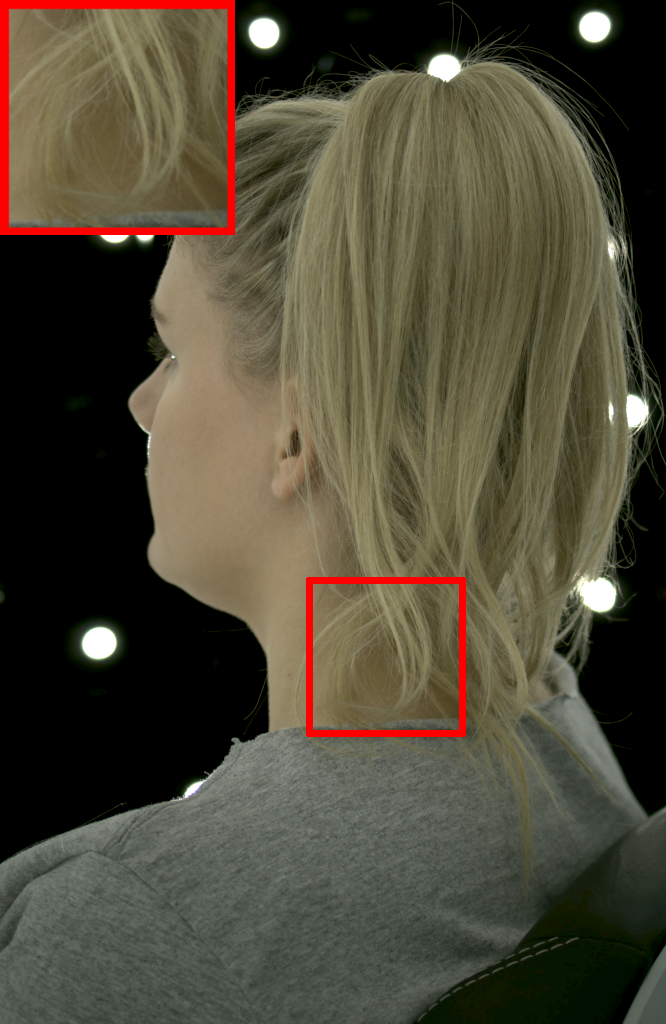} \\
 \adjincludegraphics[width=0.1\textwidth, trim={0 {0.12\height} 0 0}, clip]{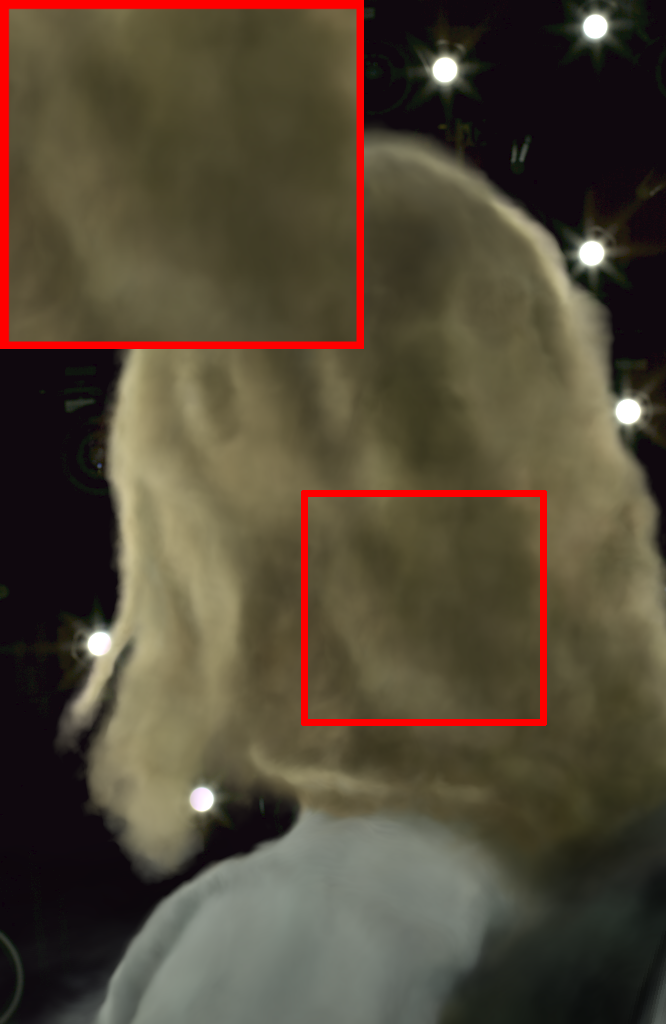} &
 \adjincludegraphics[width=0.1\textwidth, trim={0 {0.12\height} 0 0}, clip]{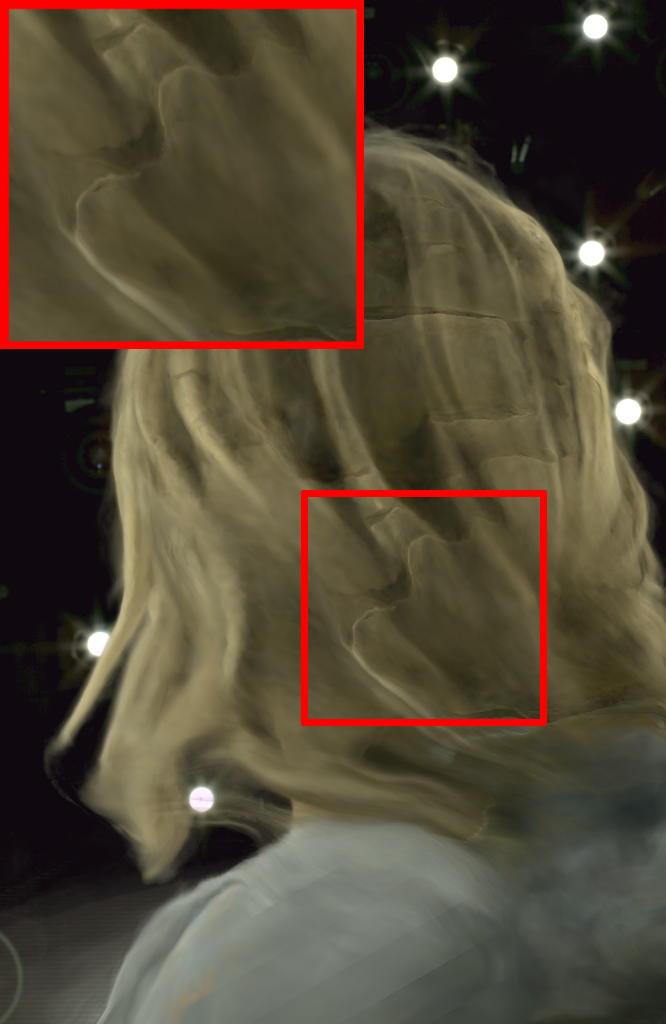} &
 \adjincludegraphics[width=0.1\textwidth, trim={0 {0.12\height} 0 0}, clip]{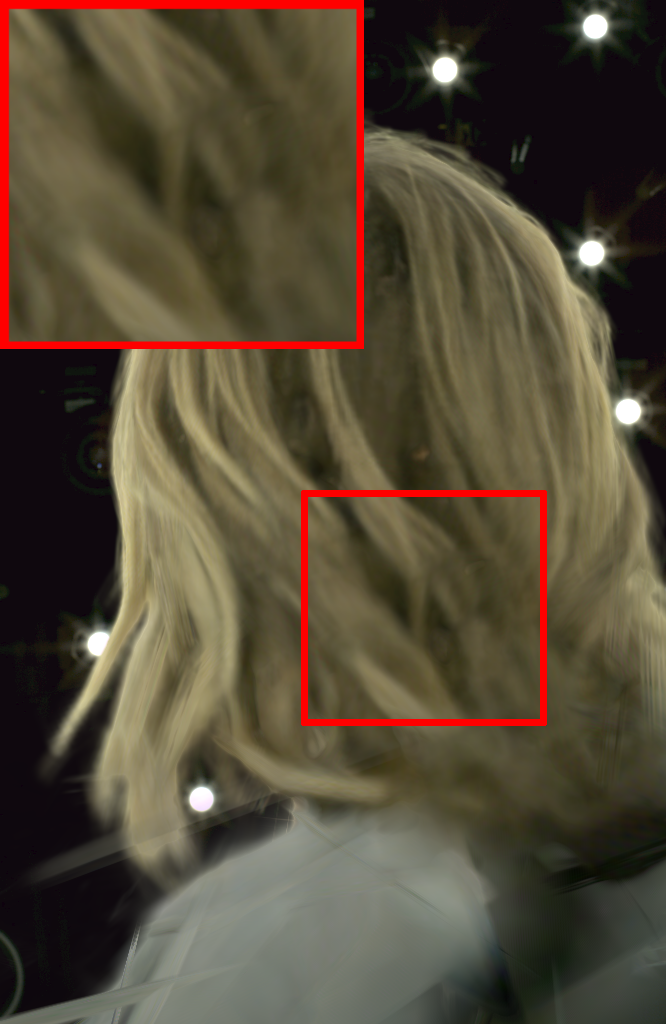} &
 \adjincludegraphics[width=0.1\textwidth, trim={0 {0.12\height} 0 0}, clip]{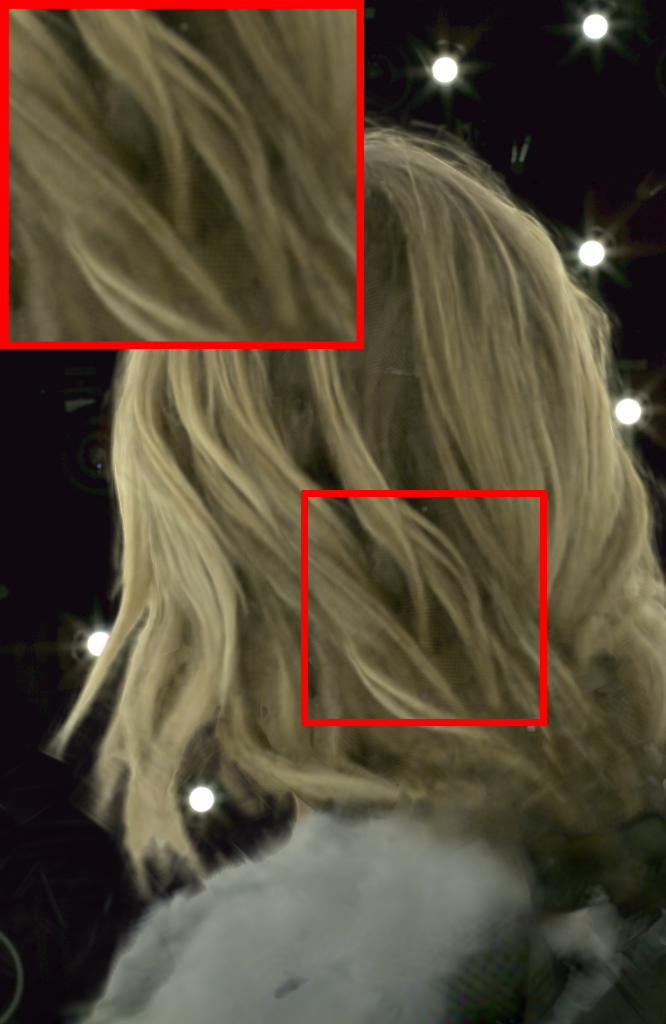} &
 \adjincludegraphics[width=0.1\textwidth, trim={0 {0.12\height} 0 0}, clip]{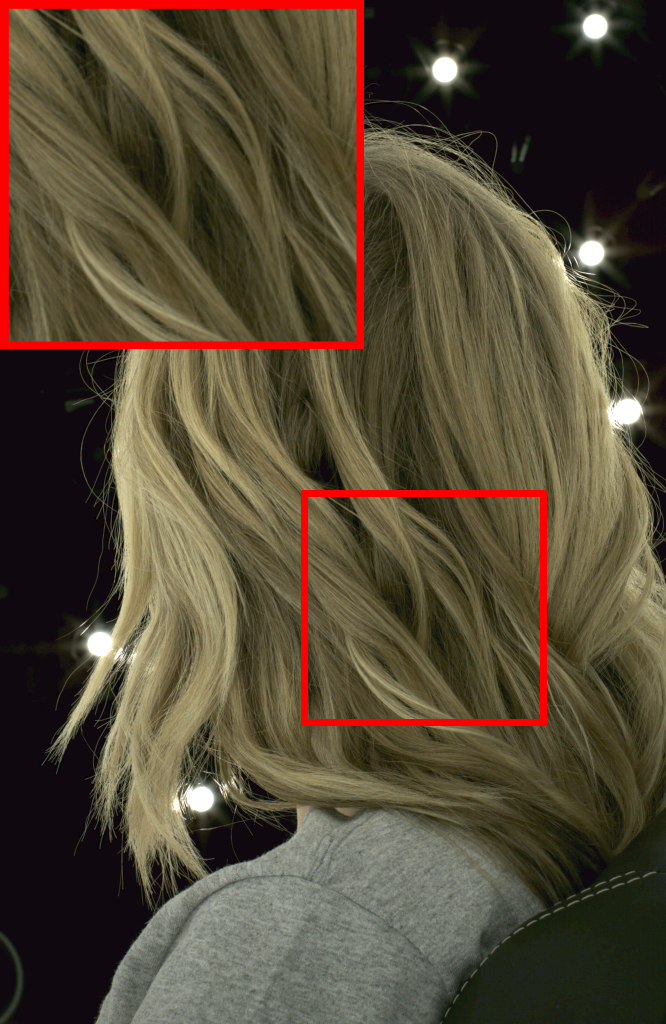} \\
 \adjincludegraphics[width=0.1\textwidth, trim={0 {0.12\height} 0 0}, clip]{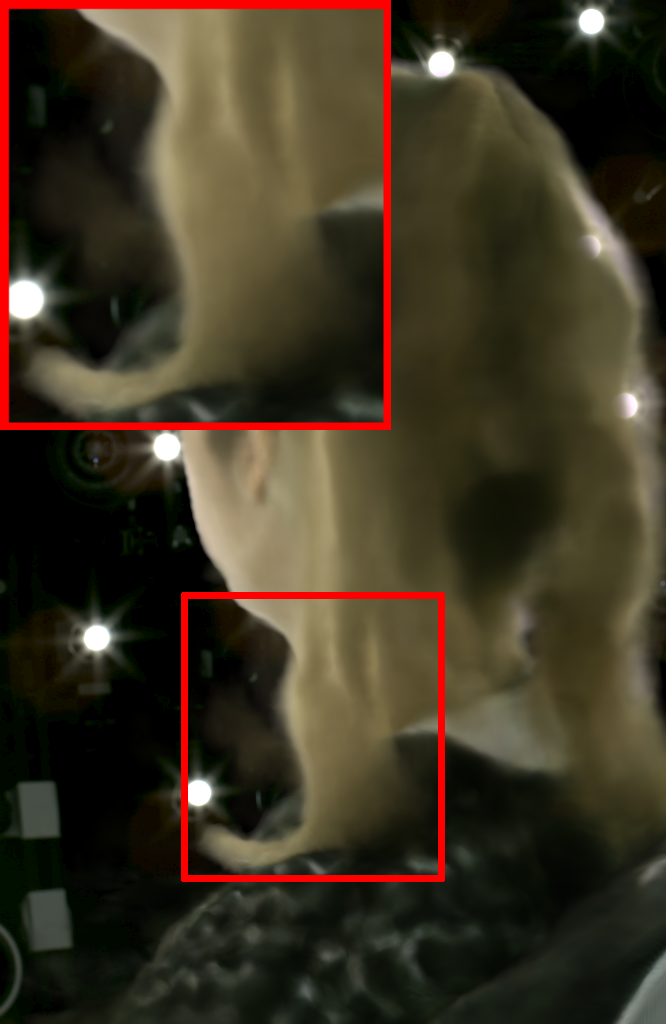} &
 \adjincludegraphics[width=0.1\textwidth, trim={0 {0.12\height} 0 0}, clip]{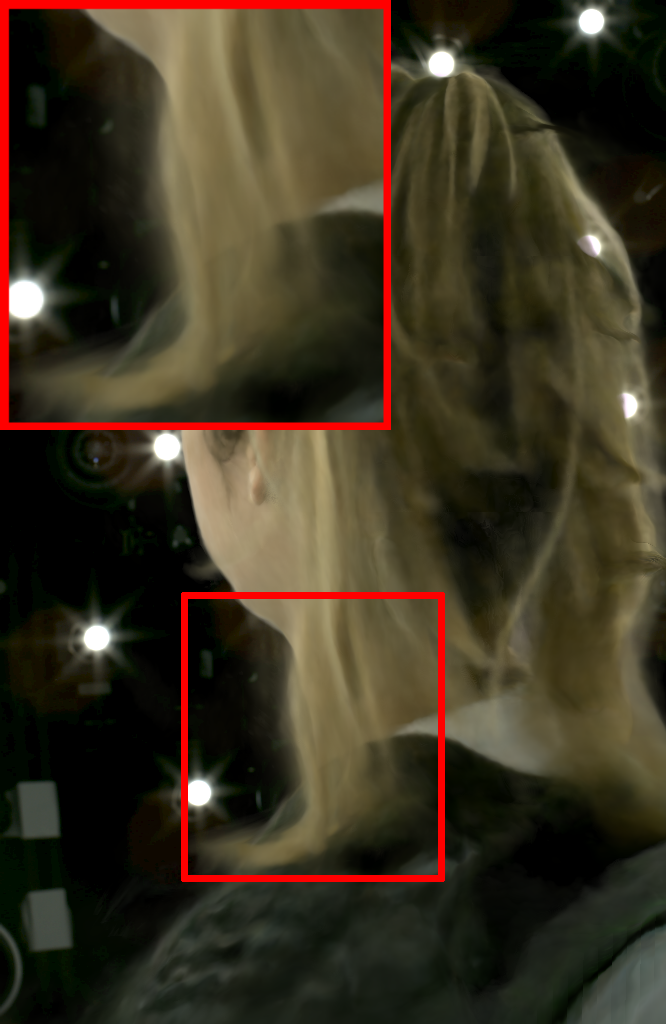} &
 \adjincludegraphics[width=0.1\textwidth, trim={0 {0.12\height} 0 0}, clip]{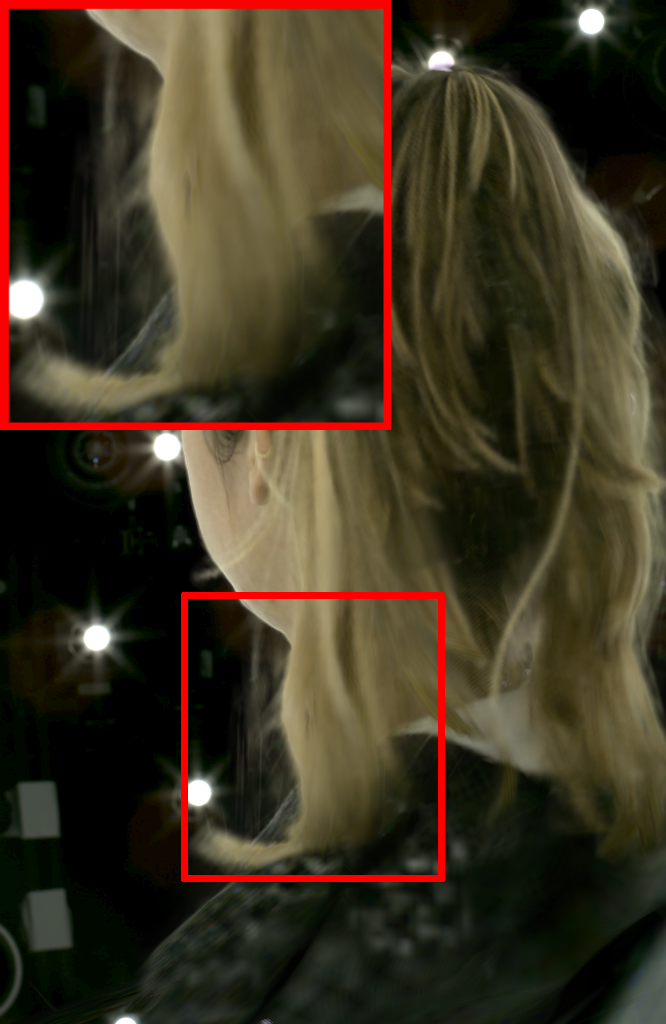} &
 \adjincludegraphics[width=0.1\textwidth, trim={0 {0.12\height} 0 0}, clip]{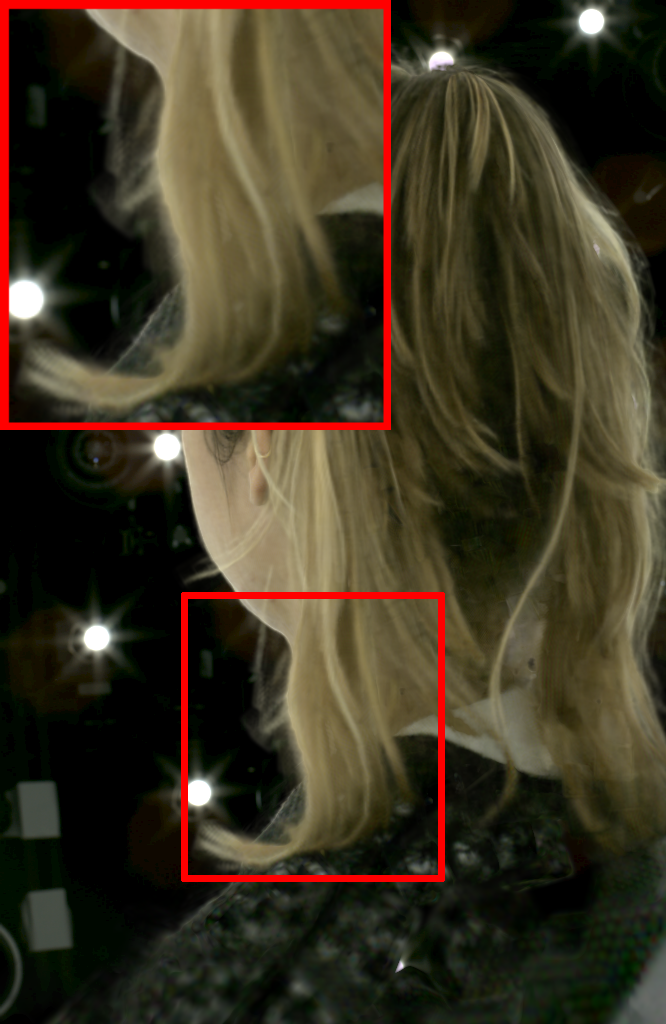} &
 \adjincludegraphics[width=0.1\textwidth, trim={0 {0.12\height} 0 0}, clip]{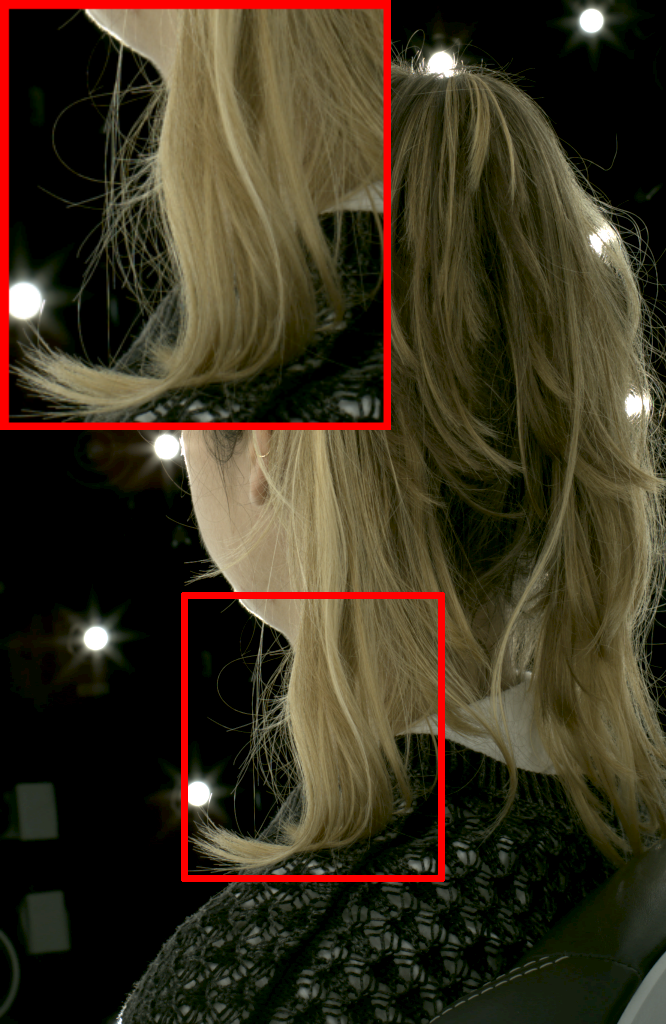}
\end{tabular}
\caption{\label{fig:nvs_compare}\textbf{Comparison on novel view synthesis between different methods.} Please see supplementary material for a bigger version of this figure.}
\end{figure}

We show both qualitative and quantitative comparisons with other methods~\cite{steve_mvp, tretschk2021nrnerf, li2021nsff} on the novel view synthesis task. In the left of Tab.~\ref{tab:nvs_train}, we show the mean squared error(MSE), SSIM and PSNR between predicted images and ground truth images from the novel views of the training sequences. Qualitative results are shown in Fig.~\ref{fig:nvs_compare}. Our method has smaller image prediction errors and is able to generate sharper results, especially on the hair regions.

\subsection{Ablation Studies}

\begin{figure}[tb]
\setlength\tabcolsep{0pt}
\renewcommand{\arraystretch}{0}
\centering
\begin{tabular}{ccccc}
 \textbf{\scriptsize MVP} &
 \textbf{\scriptsize MVP w/ flow} &
 \textbf{\scriptsize Ours w/o flow} &
 \textbf{\scriptsize Ours} &
 \textbf{\scriptsize Ground Truth} \\
 \adjincludegraphics[width=0.095\textwidth]{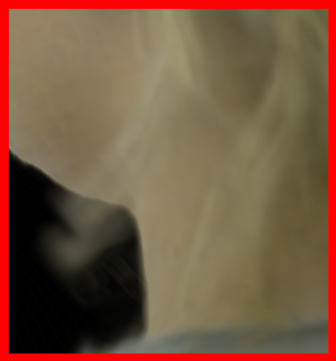} &
 \adjincludegraphics[width=0.095\textwidth]{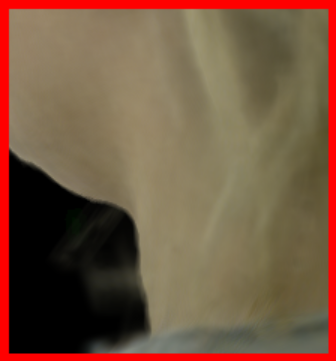} &
 \adjincludegraphics[width=0.095\textwidth]{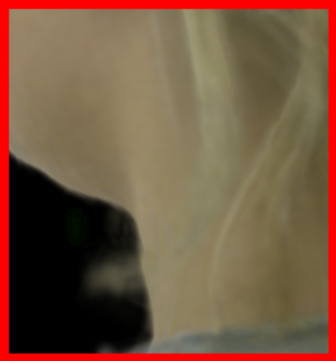} &
 \adjincludegraphics[width=0.095\textwidth]{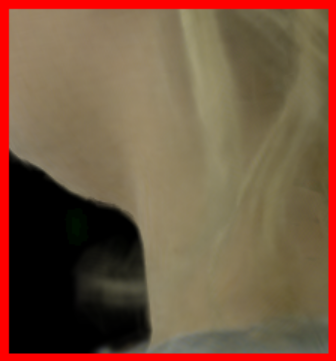} &
 \adjincludegraphics[width=0.095\textwidth]{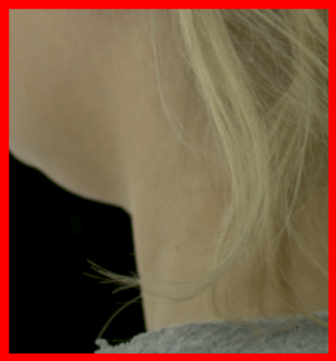} \\
 \adjincludegraphics[width=0.095\textwidth]{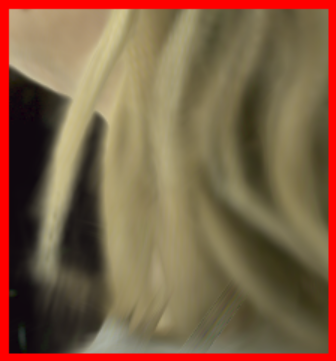} &
 \adjincludegraphics[width=0.095\textwidth]{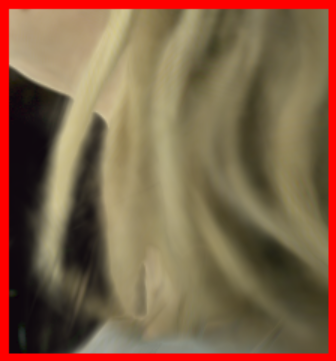} &
 \adjincludegraphics[width=0.095\textwidth]{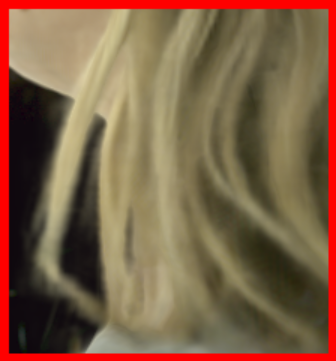} &
 \adjincludegraphics[width=0.095\textwidth]{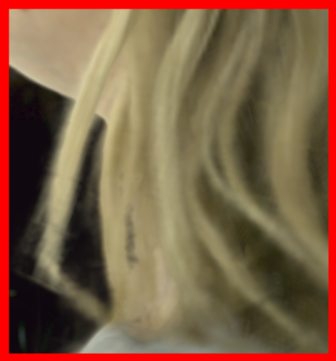} &
 \adjincludegraphics[width=0.095\textwidth]{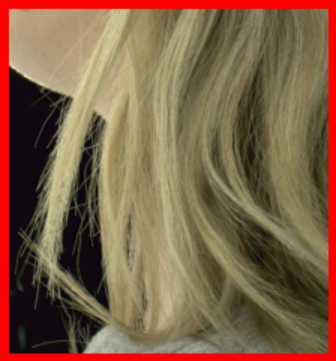} \\
 \adjincludegraphics[width=0.095\textwidth]{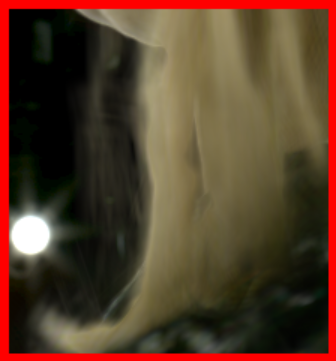} &
 \adjincludegraphics[width=0.095\textwidth]{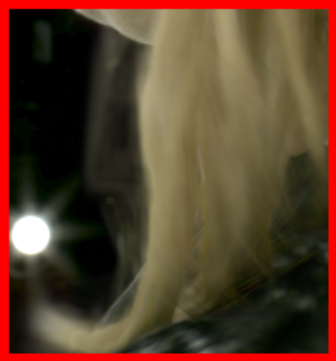} &
 \adjincludegraphics[width=0.095\textwidth]{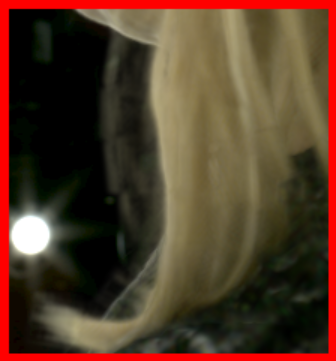} &
 \adjincludegraphics[width=0.095\textwidth]{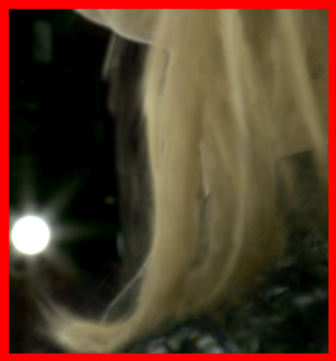} &
 \adjincludegraphics[width=0.095\textwidth]{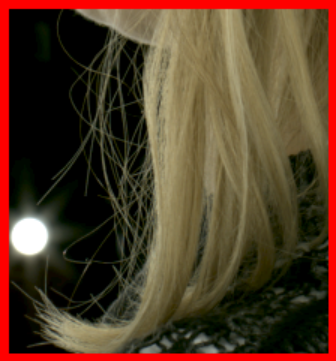}
\end{tabular}
\vspace*{0.2cm}
\caption{\label{fig:nvs_abl_mini}\textbf{Ablation of temporal consistency.} We compare our method and MVP w/ and w/o flow supervision. With flow supervision, better temporal consistency and generalization for unseen sequence can be observed. Please see supplementary for a bigger version of this figure.}
\end{figure}

\begin{figure}[tb]
\setlength\tabcolsep{0pt}
\renewcommand{\arraystretch}{0}
\centering
\begin{tabular}{ccc}
 \textbf{\small w/o flow sup.} &
 \textbf{\small w/ flow sup.} &
 \textbf{\small Ground truth} \\
  \adjincludegraphics[width=0.15\textwidth, trim={{0.15\width} {0.12\height} 0 {0.1\height}}, clip]{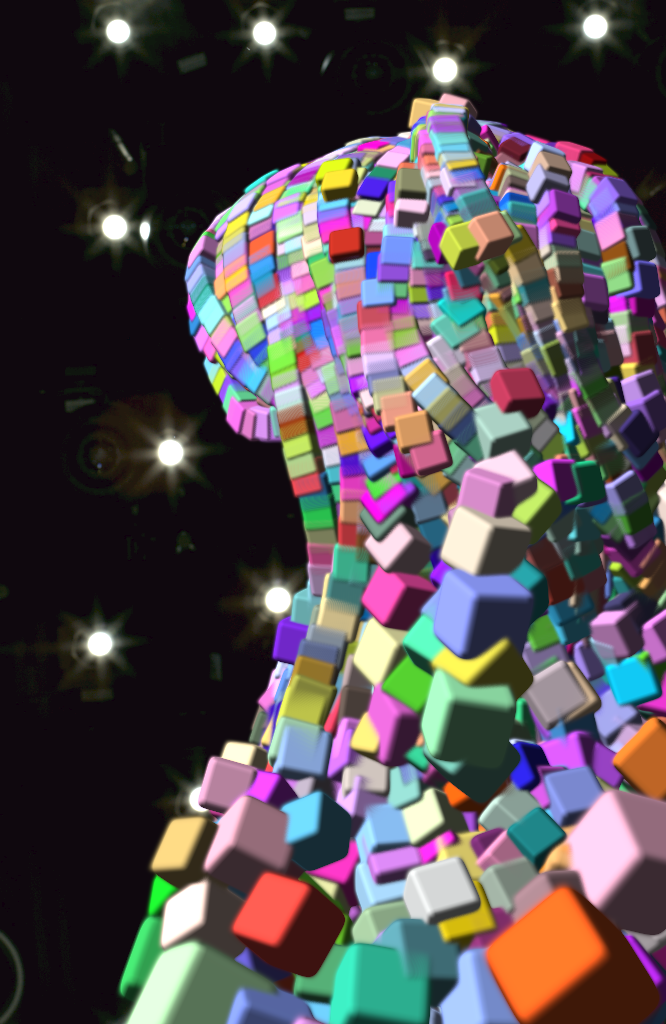} &
 \adjincludegraphics[width=0.15\textwidth, trim={{0.15\width} {0.12\height} 0 {0.1\height}}, clip]{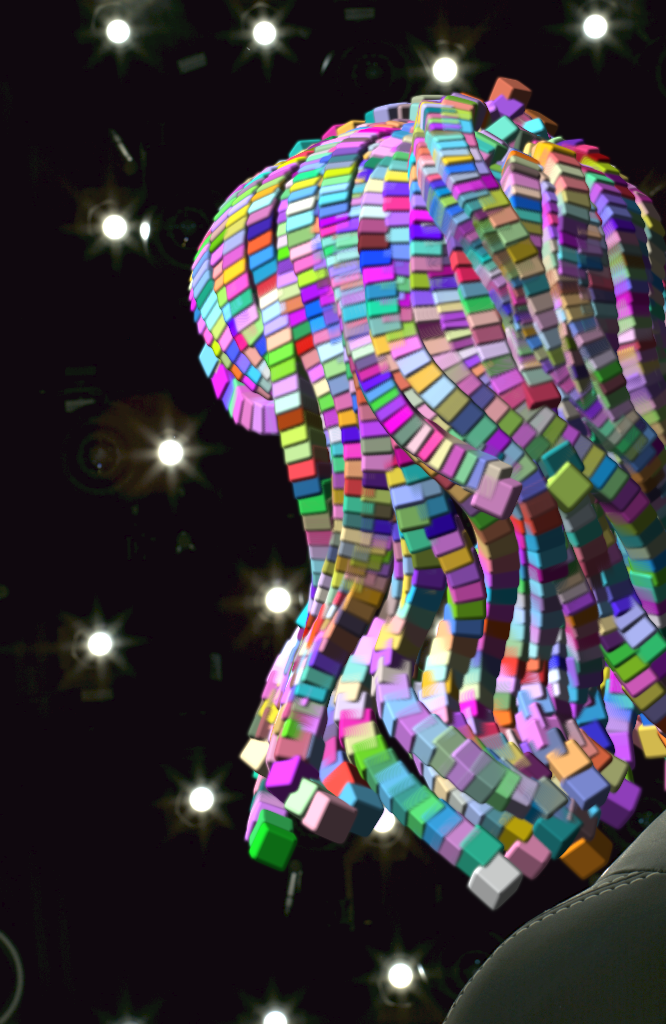} &
 \adjincludegraphics[width=0.15\textwidth, trim={{0.15\width} {0.12\height} 0 {0.1\height}}, clip]{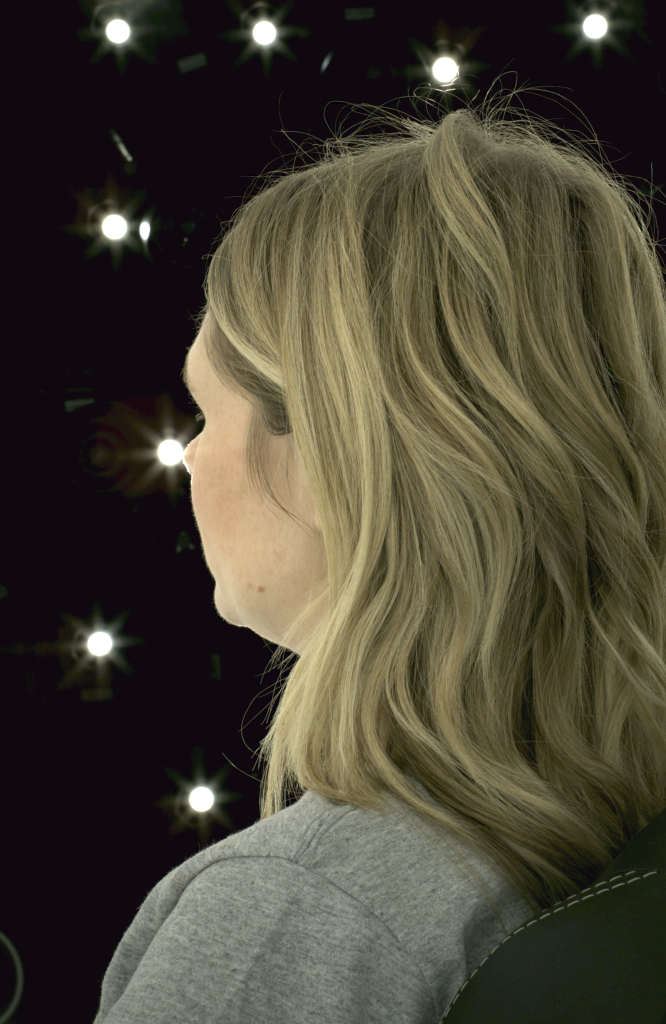}
\end{tabular}
\caption{\label{fig:flow_vol_abl}\textbf{Ablation on flow supervision.} We further compare the volumetric primitives of the models w/ and w/o flow supervision. We see that model with additional flow supervision yields a consistent and reasonable shape for hair and yields better hair shoulder disentanglement.}
\end{figure}

\noindent\textbf{Temporal consistency.} To test the effects of the temporal consistency and the tracked guide hair, we also conduct a novel view synthesis task on the test portion of our captured sequence. Note that our model is not trained using any part of the test sequence data. On the right of Tab.~\ref{tab:nvs_train}, we report MSE, SSIM, PSNR on novel views of both seen and unseen sequences. As we can see, having the coarse level guide hair strands tracked and without flow supervision gives us better rendering quality. With flow supervision, the results are improved further. This improvement is because the tracking information helps the volumetric primitives to better localize the hair region with higher consistency. While the improvement for seen motions is relatively small, both our model and MVP are notably improved for unseen sequences with novel hair motion when flow supervision is added. Rendering results on unseen sequences are shown in Fig.~\ref{fig:nvs_abl_mini}. In Fig.~\ref{fig:flow_vol_abl}, we visualize the volumetric primitives of the hairs of our model with and without flow supervision. Including flow supervision produces notably better disentanglement between the hair and shoulder.

\begin{figure}[tb]
\setlength\tabcolsep{0pt}
\renewcommand{\arraystretch}{0}
\centering
\begin{tabular}{cccc}
\textbf{\scriptsize \begin{tabular}[c]{@{}c@{}}w/o $\mathcal{L}_{cur}$\\ $\mathcal{L}_{len}+\mathcal{L}_{tang}$\end{tabular}} &
 \textbf{\scriptsize w/o $\mathcal{L}_{len}+\mathcal{L}_{tang}$} &
 \textbf{\scriptsize w/o $\mathcal{L}_{cur}$} &
 \textbf{\scriptsize \begin{tabular}[c]{@{}c@{}}w/ $\mathcal{L}_{cur}$\\ $\mathcal{L}_{len}+\mathcal{L}_{tang}$\end{tabular}} \\
\adjincludegraphics[width=0.12\textwidth, valign=m, trim={{0.1\width} {0.2\height} {0.1\width} {0.05\height}}, clip]{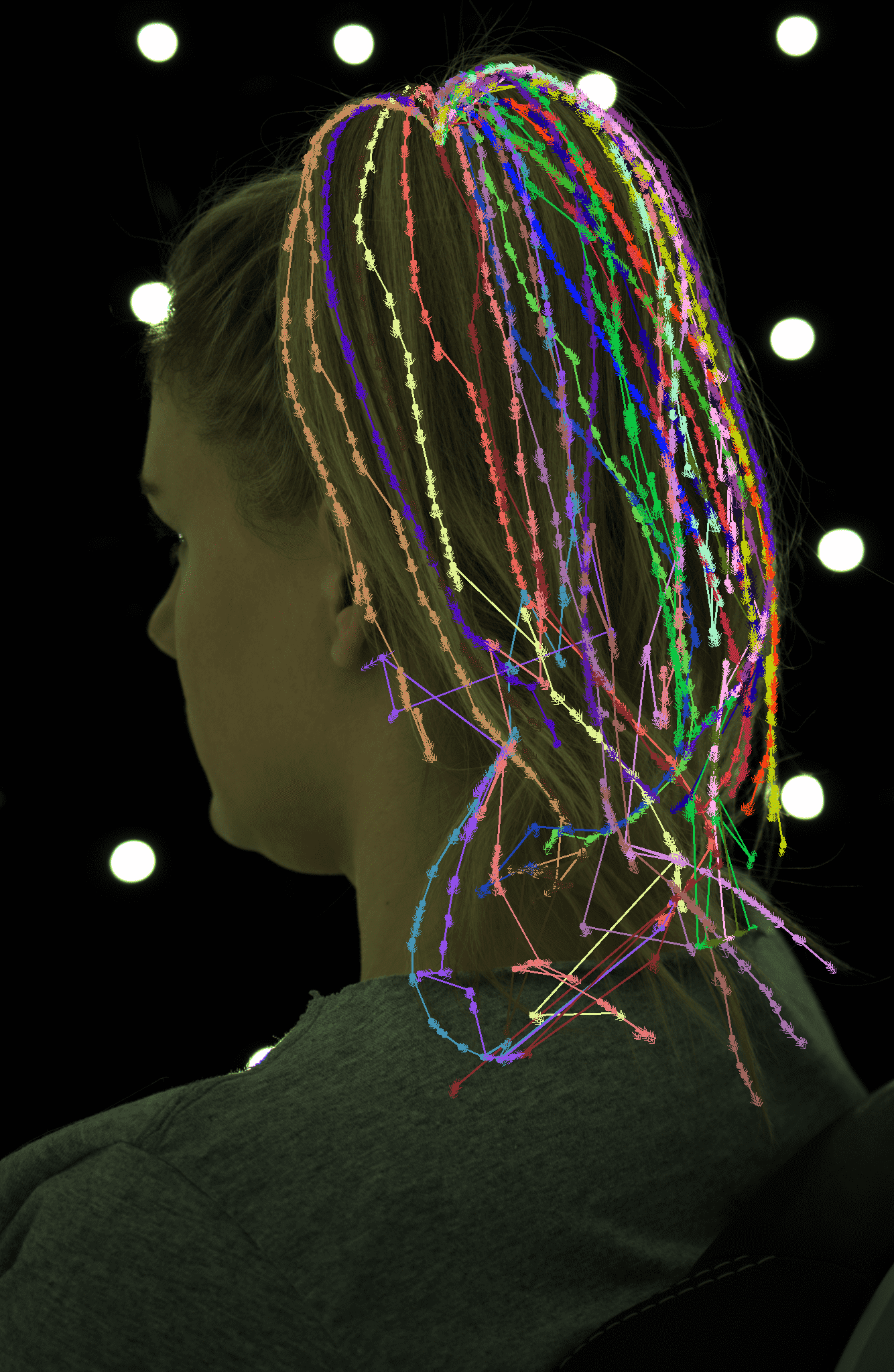} &
 \adjincludegraphics[width=0.12\textwidth, valign=m, trim={{0.1\width} {0.2\height} {0.1\width} {0.05\height}}, clip]{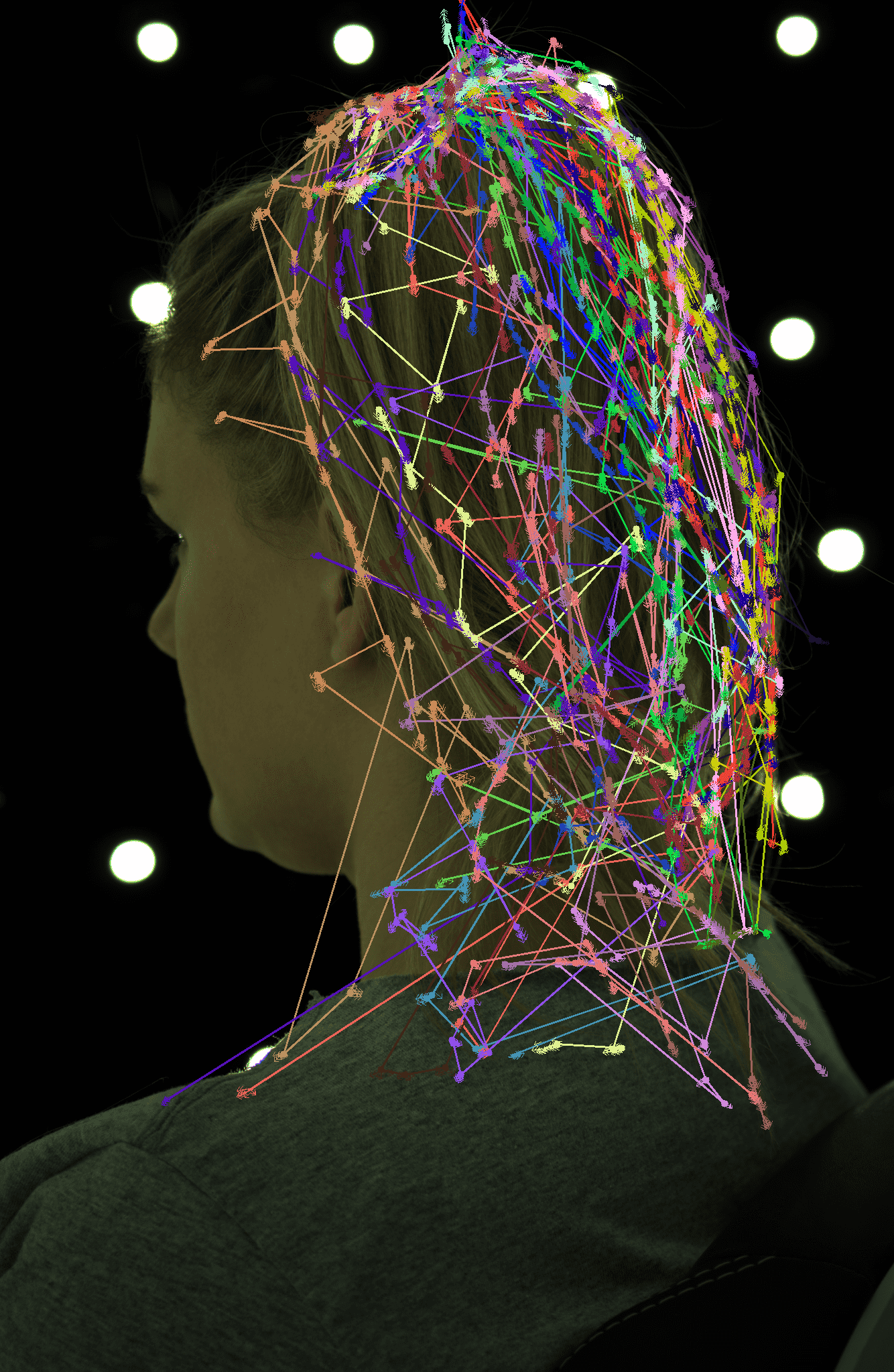} &
 \adjincludegraphics[width=0.12\textwidth, valign=m, trim={{0.1\width} {0.2\height} {0.1\width} {0.05\height}}, clip]{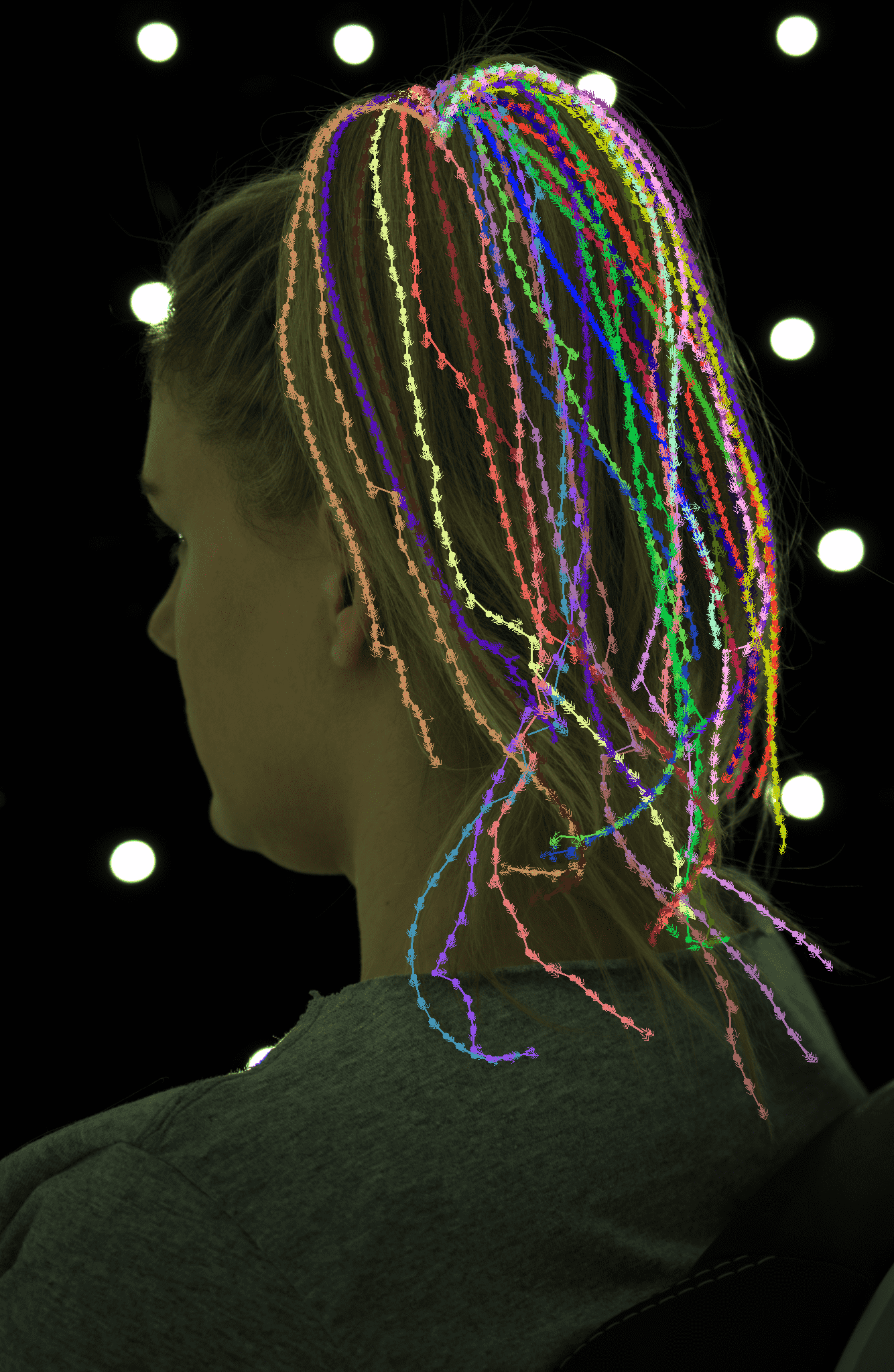} &
 \adjincludegraphics[width=0.12\textwidth, valign=m, trim={{0.1\width} {0.2\height} {0.1\width} {0.05\height}}, clip]{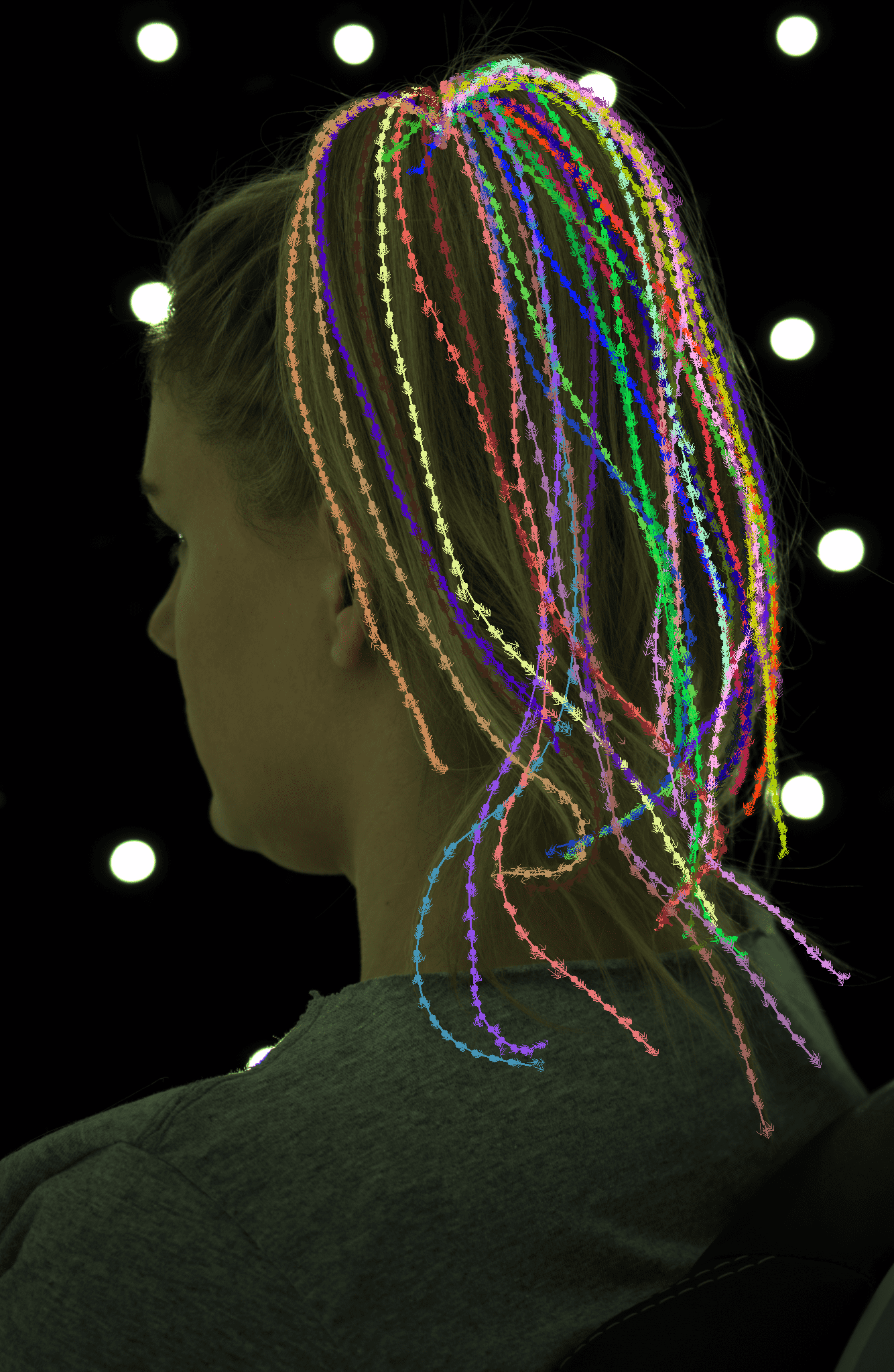}
\end{tabular}
\caption{\label{fig:track_loss_abl}\textbf{Effects of $\mathcal{L}_{len}+\mathcal{L}_{tang}$ and $\mathcal{L}_{cur}$.}  We show how the shape and curvature of tracked hair strands are preserved with both $\mathcal{L}_{len}+\mathcal{L}_{tang}$ and $\mathcal{L}_{cur}$.
}
\end{figure}

\noindent\textbf{Hair tracking analysis.} We first study the impact of different objectives $\mathcal{L}_{len}+\mathcal{L}_{tang}$ and $\mathcal{L}_{cur}$ in hair tracking.
As in Fig.~\ref{fig:track_loss_abl}, when both $\mathcal{L}_{cur}$ and $\mathcal{L}_{len}+\mathcal{L}_{tang}$ are applied, the tracking results are more smooth and without kinks. 
%
%
We observe that, when using the loss $\mathcal{L}_{len}+\mathcal{L}_{tang}$ as the only regularization term, the length of each hair strand segments are already preserved but could cause some kinks without awareness of the correct hair strand curvatures.
$\mathcal{L}_{cur}$ itself does not help and exaggerates the error when the hair strand length is not correct, but yields smooth results when combined with $\mathcal{L}_{len}+\mathcal{L}_{tang}$.  This is because curvature computation is agnostic to absolute length of the hair and only controls the relative length ratio. 

\begin{figure}[tb]
\setlength\tabcolsep{0pt}
\renewcommand{\arraystretch}{0}
\centering
\begin{tabular}{cccccc}
 &
 \textbf{\scriptsize frame 118} &
 \textbf{\scriptsize frame 119} &
 \textbf{\scriptsize frame 120} &
 \textbf{\scriptsize frame 121} &
 \textbf{\scriptsize frame 122} \\
 \parbox{0.04\textwidth}{no \\ mm.} & 
 \adjincludegraphics[width=0.088\textwidth, valign=m]{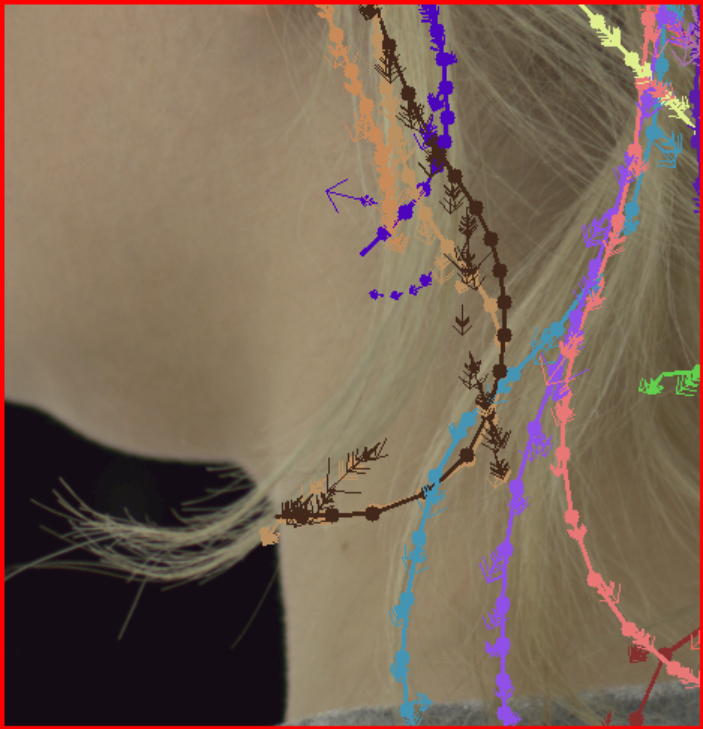} &
 \adjincludegraphics[width=0.088\textwidth, valign=m]{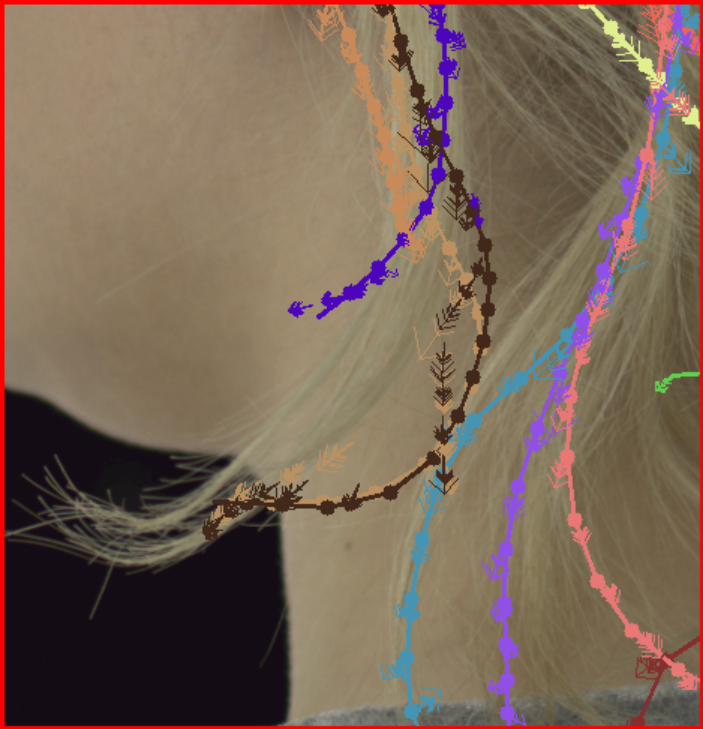} &
 \adjincludegraphics[width=0.088\textwidth, valign=m]{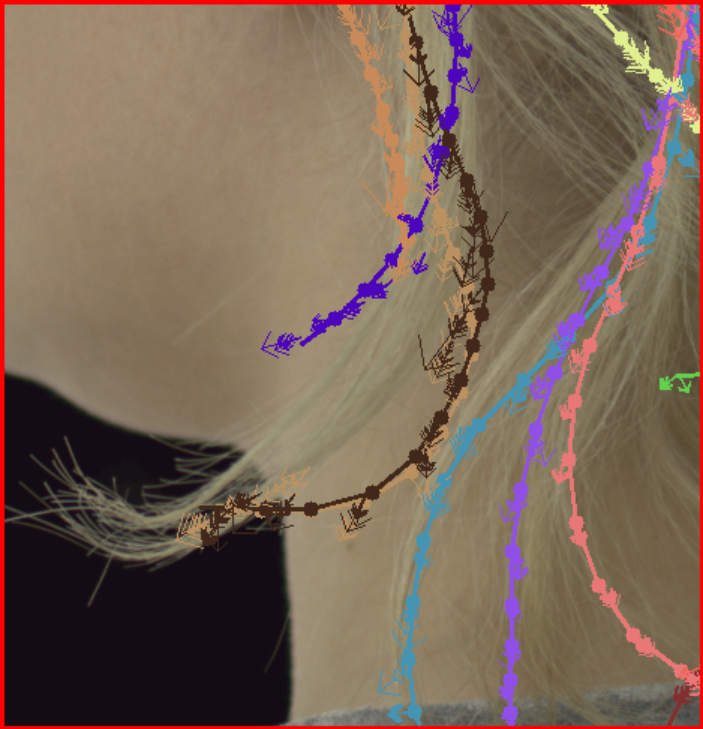} &
 \adjincludegraphics[width=0.088\textwidth, valign=m]{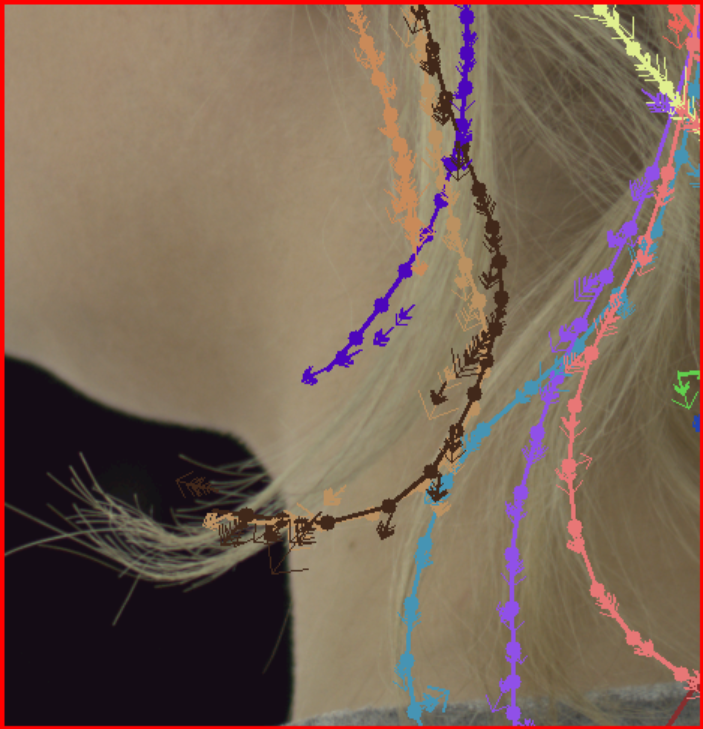} &
 \adjincludegraphics[width=0.088\textwidth, valign=m]{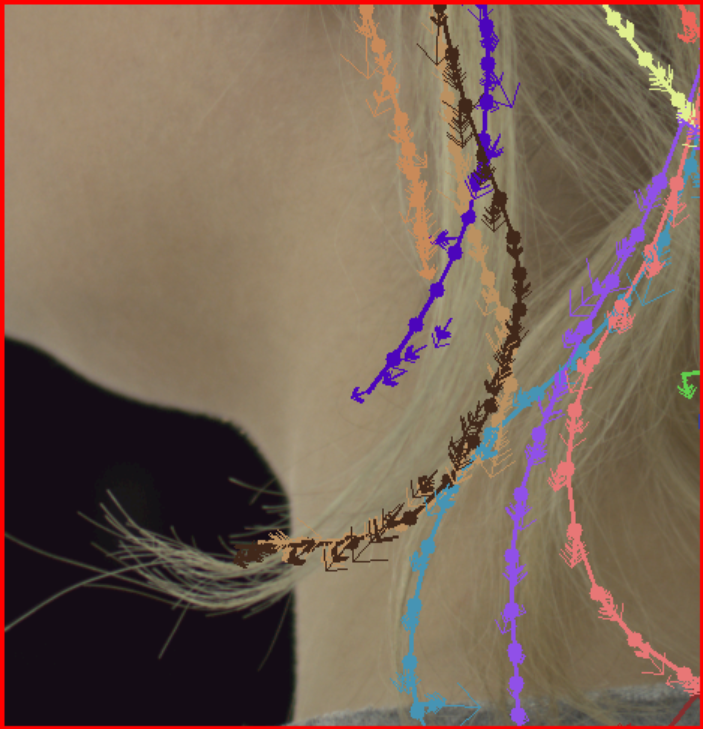} \\
 \parbox{0.04\textwidth}{1st ord. \\ mm.} & 
 \adjincludegraphics[width=0.088\textwidth, valign=m]{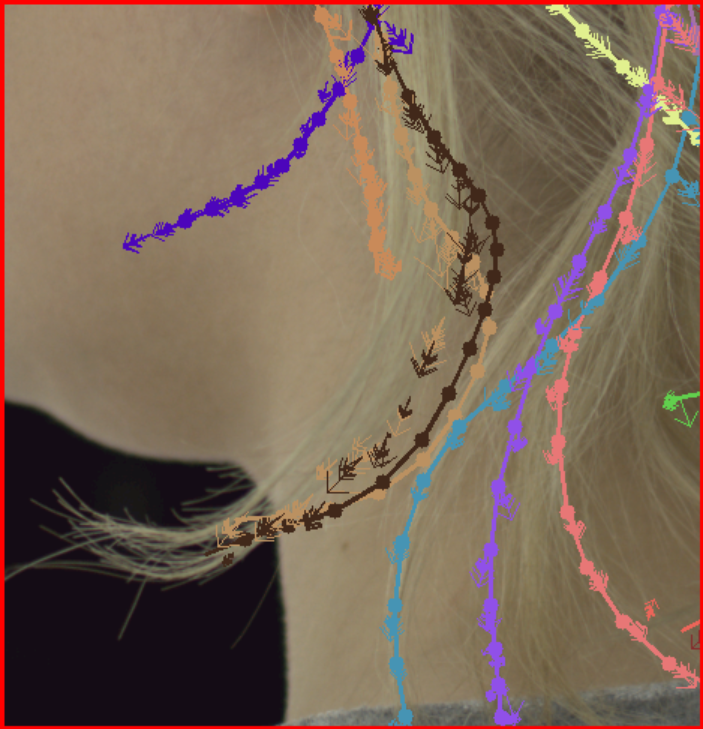} &
 \adjincludegraphics[width=0.088\textwidth, valign=m]{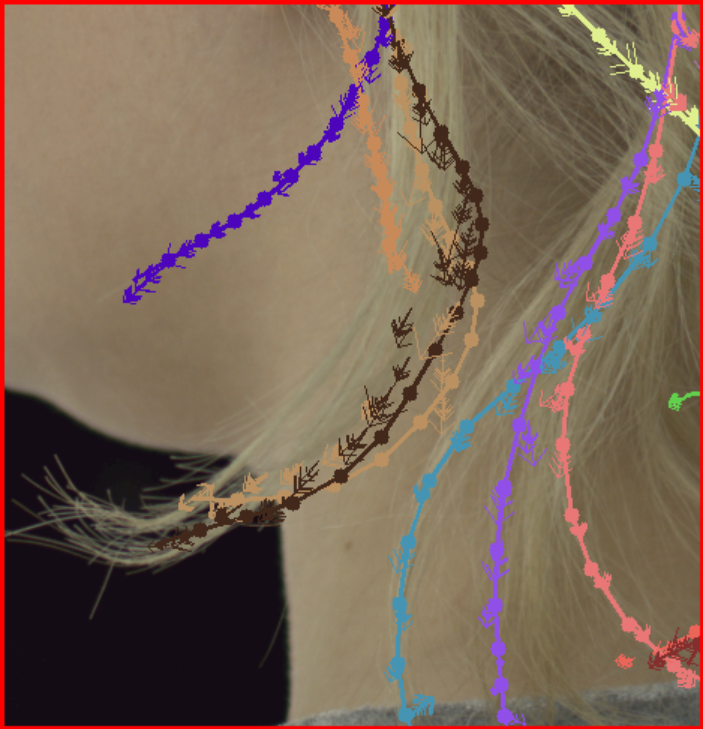} &
 \adjincludegraphics[width=0.088\textwidth, valign=m]{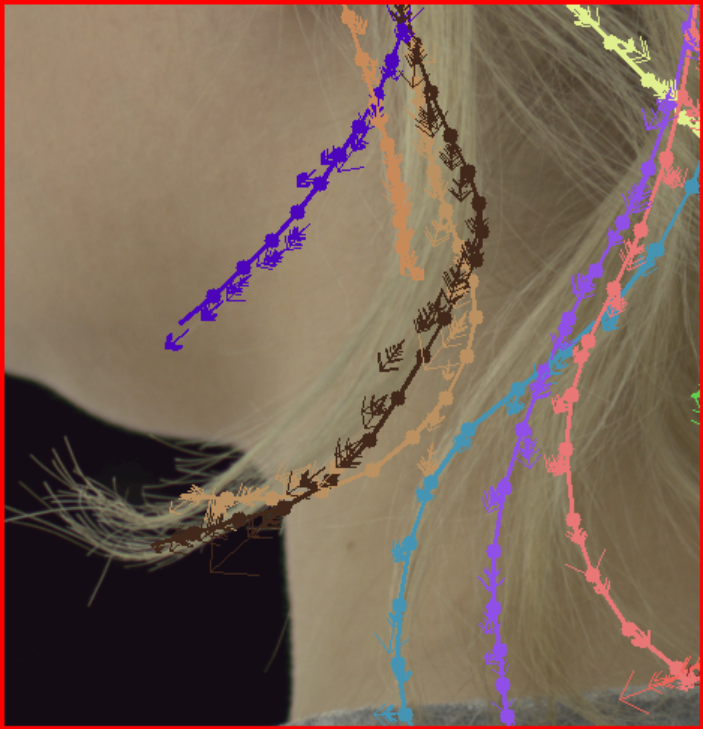} &
 \adjincludegraphics[width=0.088\textwidth, valign=m]{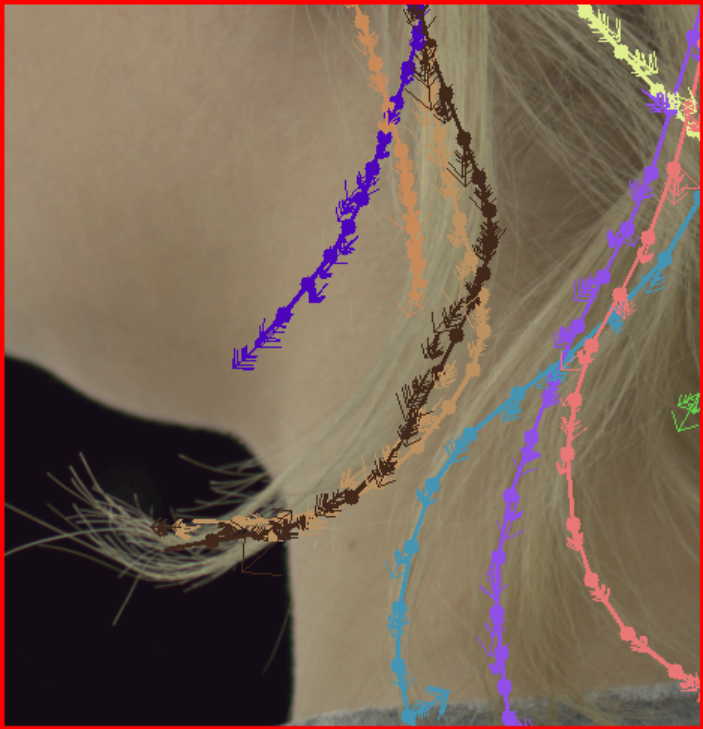} &
 \adjincludegraphics[width=0.088\textwidth, valign=m]{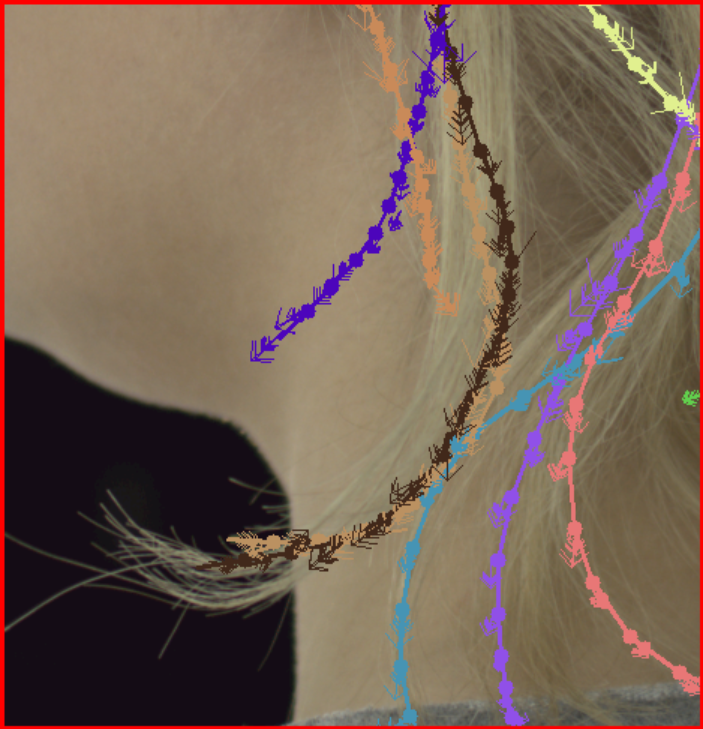} \\
 \parbox{0.04\textwidth}{2nd ord. \\ mm.} & 
 \adjincludegraphics[width=0.088\textwidth, valign=m]{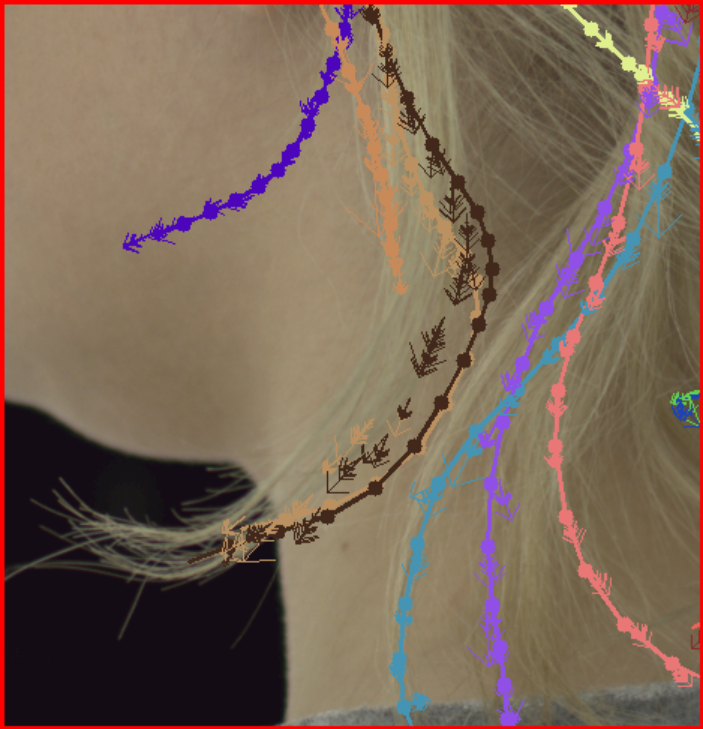} &
 \adjincludegraphics[width=0.088\textwidth, valign=m]{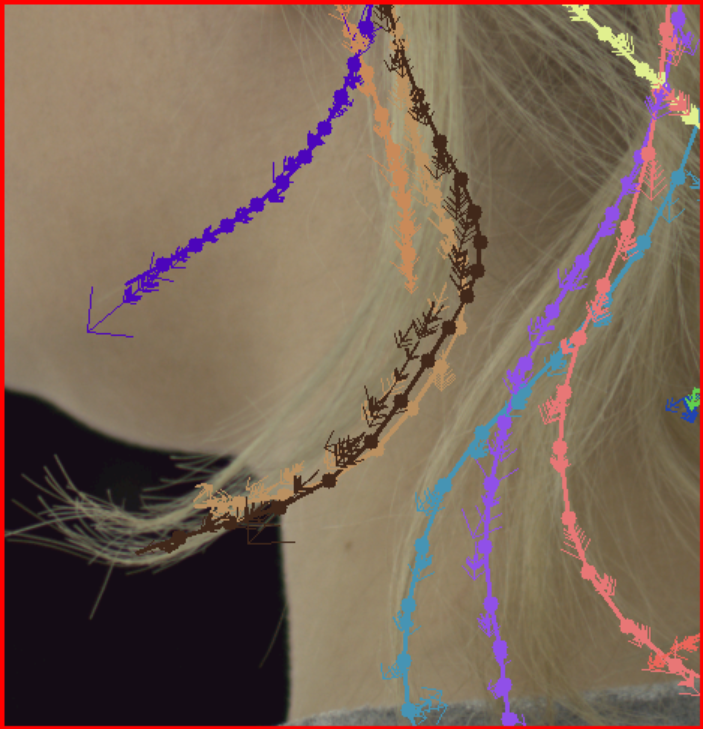} &
 \adjincludegraphics[width=0.088\textwidth, valign=m]{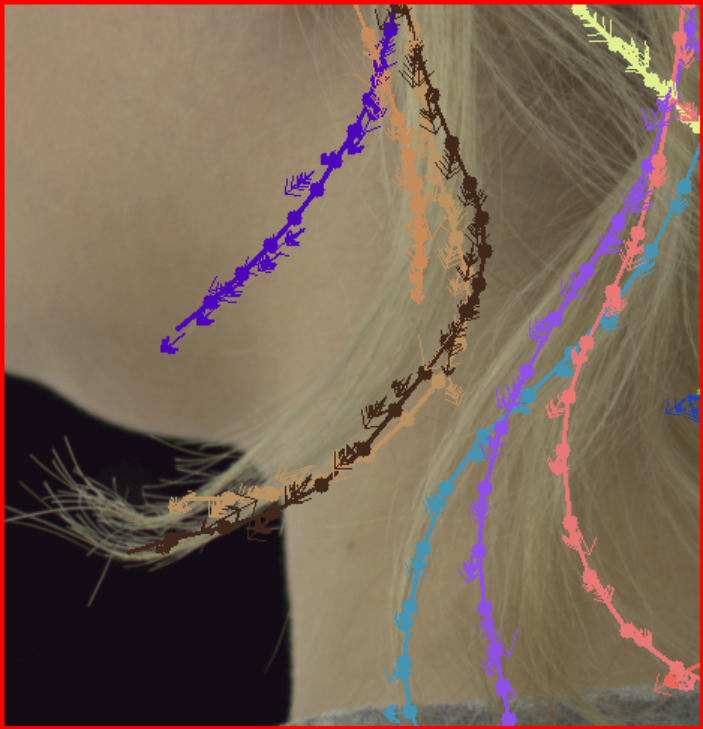} &
 \adjincludegraphics[width=0.088\textwidth, valign=m]{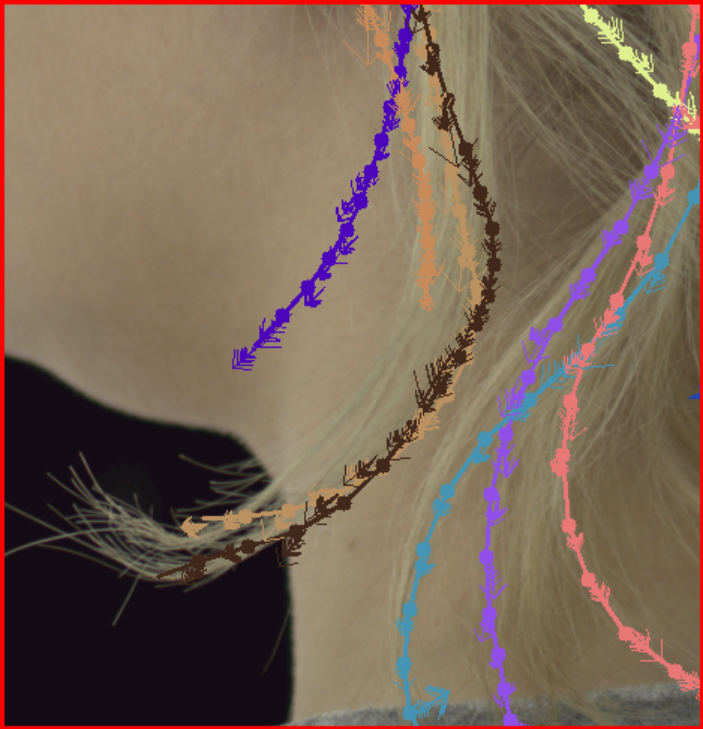} &
 \adjincludegraphics[width=0.088\textwidth, valign=m]{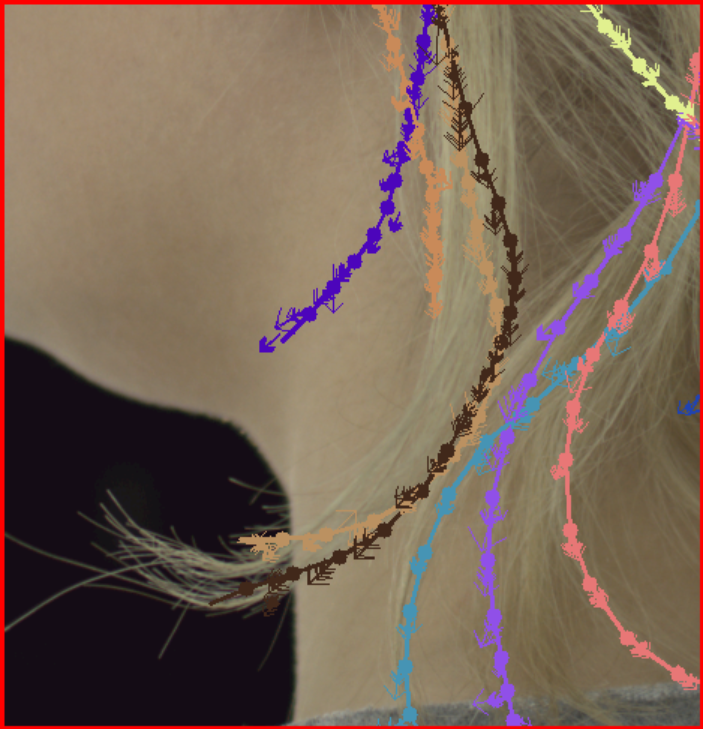} \\
\end{tabular}
\caption{\label{fig:track_abl_ims}\textbf{Ablation of different initialization in hair tracking.} We show tracking results of our methods with different initializations. From top to bottom, we use no momentum information, first and second order momentum information for tracking initialization. Please note the brown and orange strands. As we can see, the hairs are better tracked when we utilize the dynamic information from previous frames. Better view in color version.}
\end{figure}

We show the impact of different initialization for hair tracking in Fig.~\ref{fig:track_abl_ims}. When no momentum information from previous frames is used, there is more obvious drifting on some of the strands happening, while the drifting is less severe when we take advantage of the motion information from previous frames.

\section{Applications and Limitations} 

\begin{figure}[tb]
\setlength\tabcolsep{0pt}
\renewcommand{\arraystretch}{0}
\centering
\begin{tabular}{cccc}
\adjincludegraphics[width=0.12\textwidth, valign=m, trim={{0.4\width} {0.2\height} {0.2\width} 0}, clip]{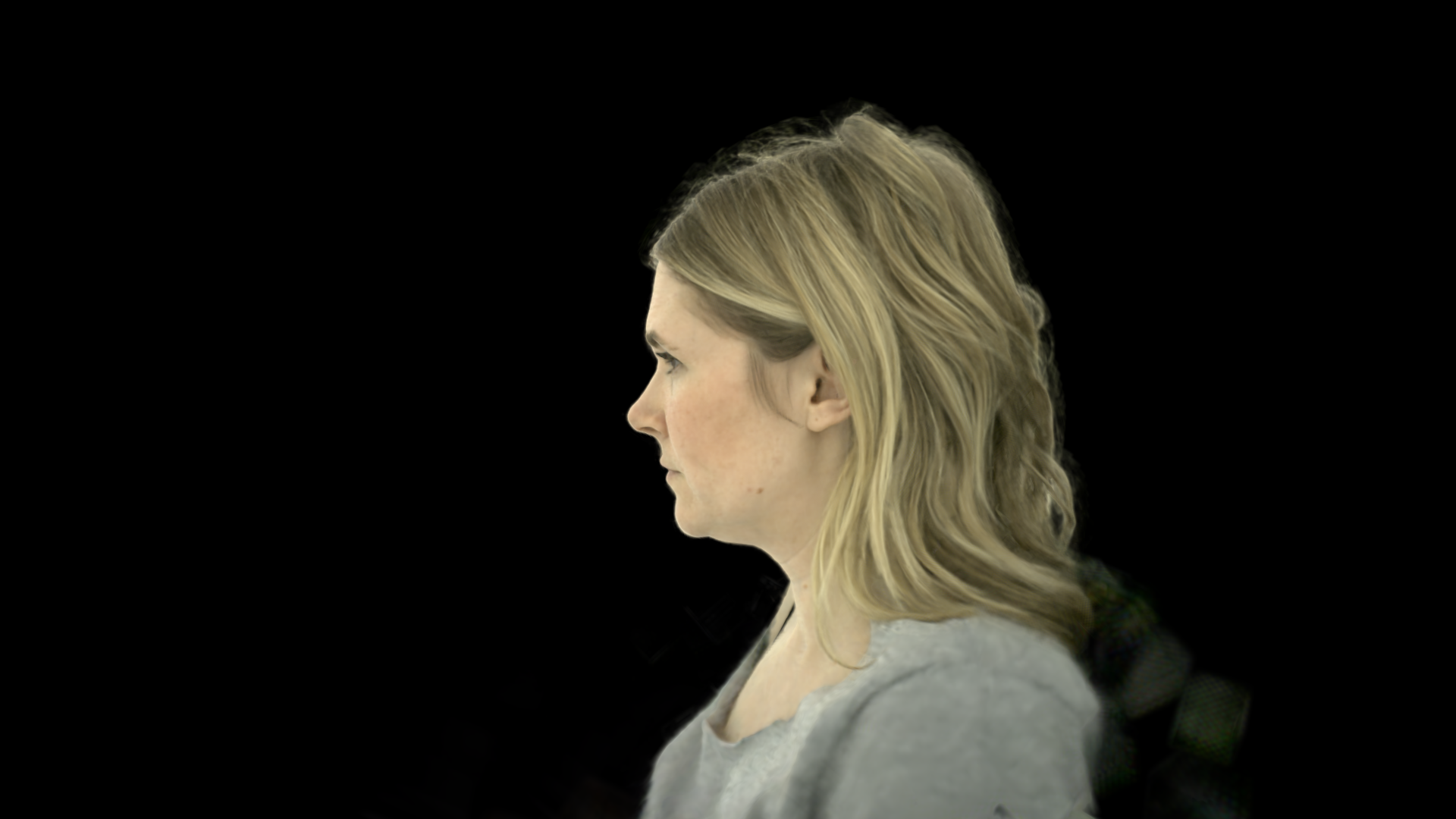} &
 \adjincludegraphics[width=0.12\textwidth, valign=m, trim={{0.4\width} {0.2\height} {0.2\width} 0}, clip]{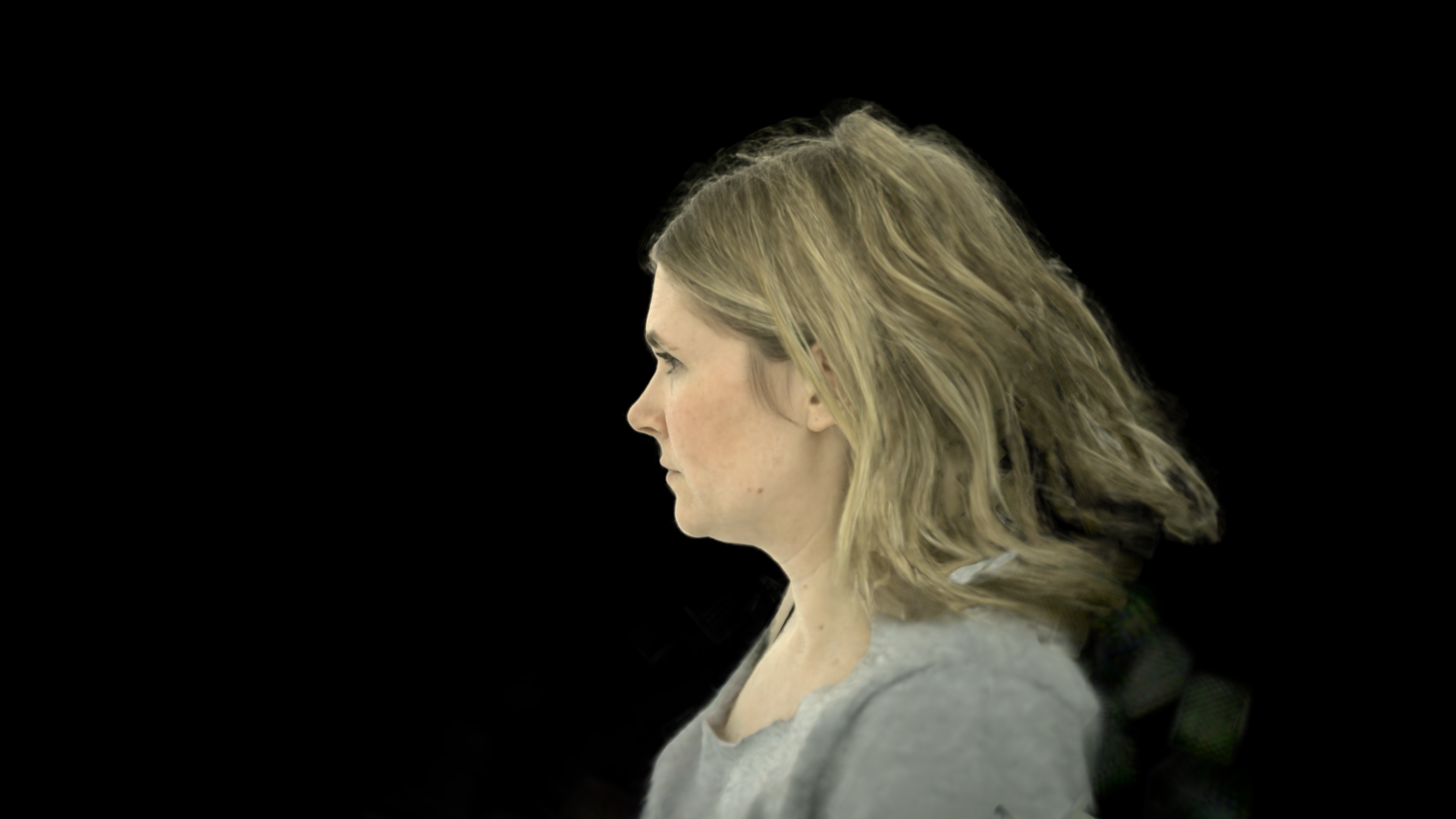} &
 \adjincludegraphics[width=0.12\textwidth, valign=m, trim={{0.3\width} {0.2\height} {0.3\width} 0}, clip]{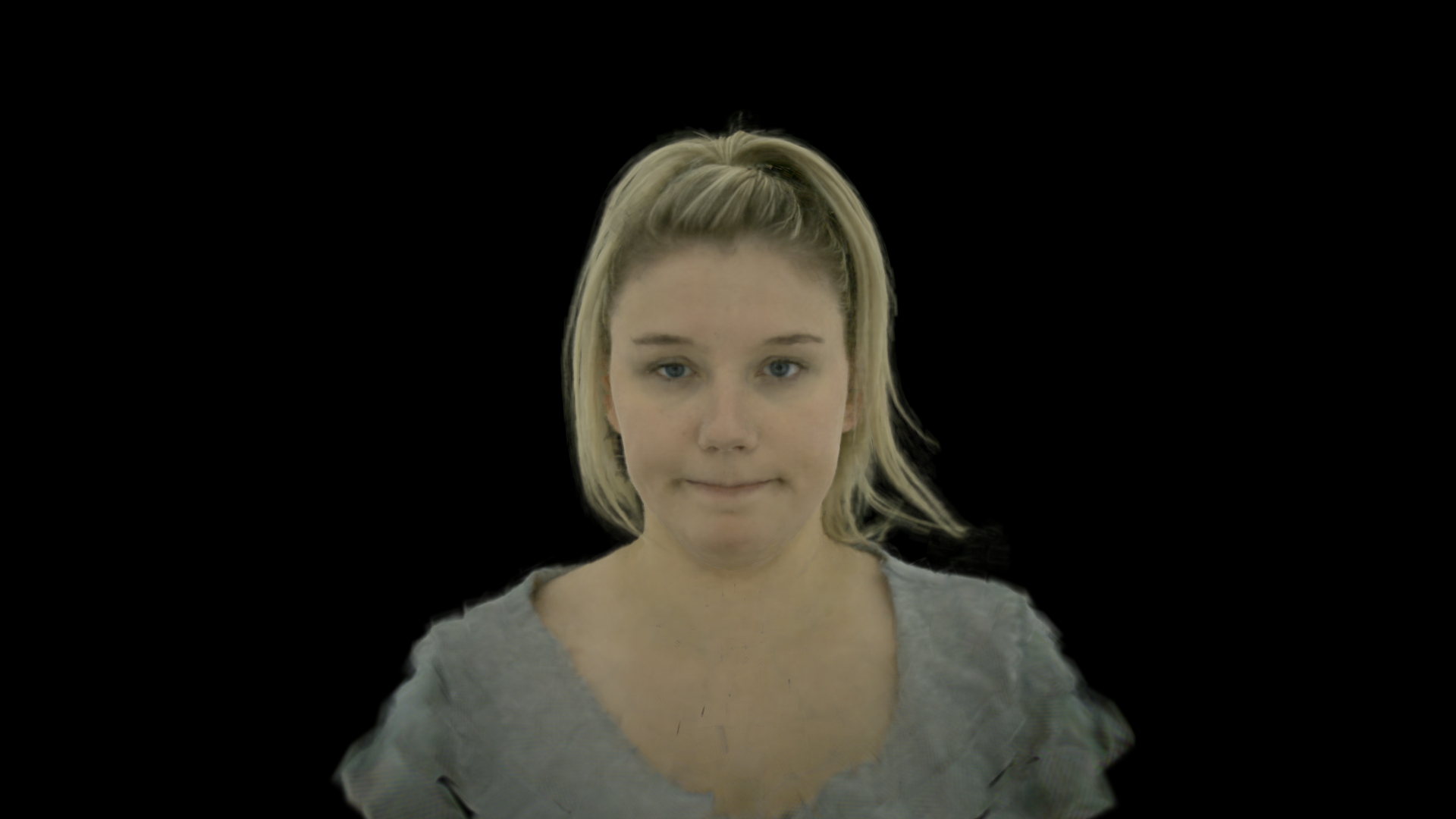} &
  \adjincludegraphics[width=0.12\textwidth, valign=m, trim={{0.3\width} {0.2\height} {0.3\width} 0}, clip]{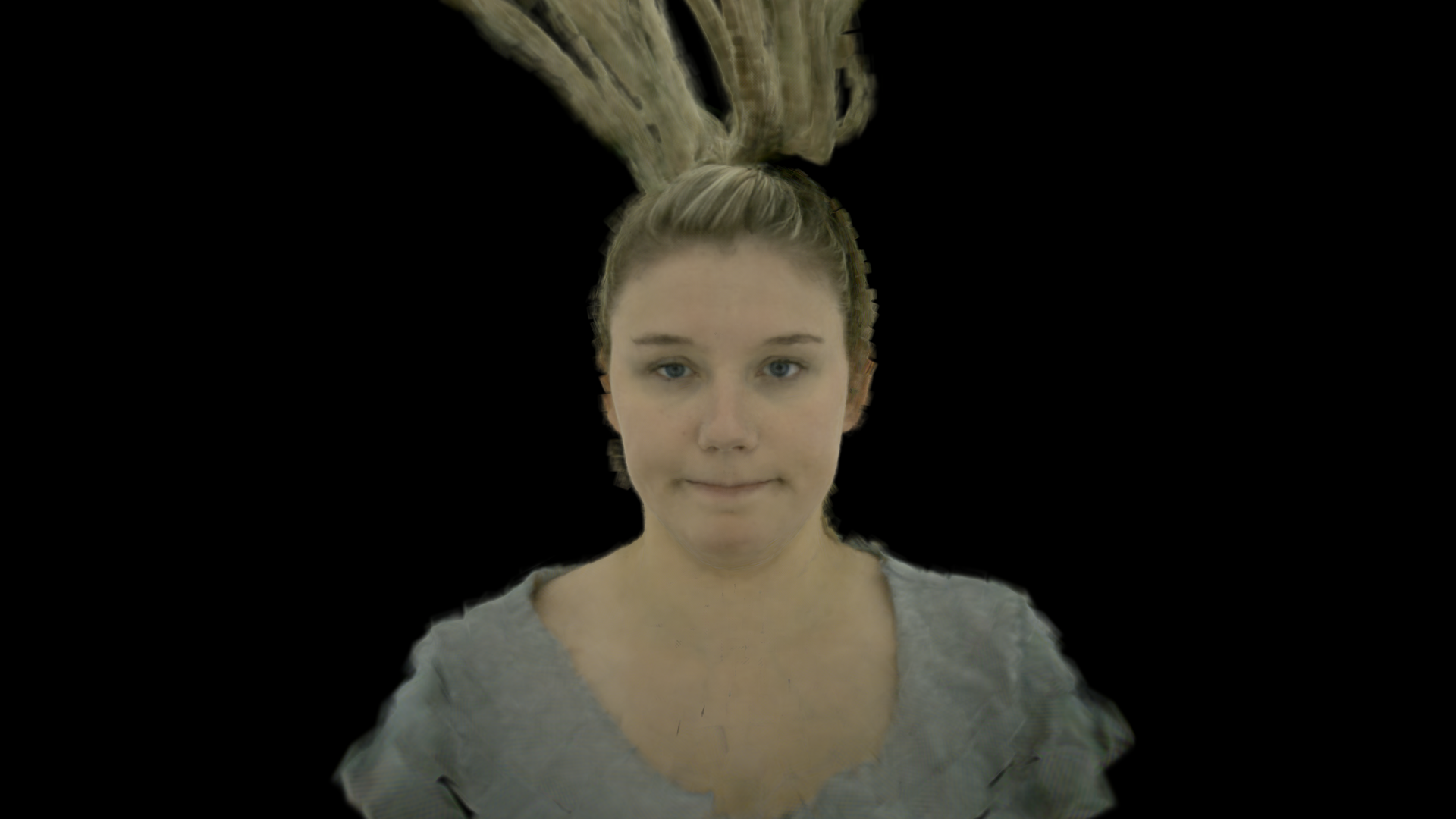}
\end{tabular}
\caption{\label{fig:edit_hair}\textbf{Hair position editing.} We create a new animation by direct editing on the guide hair strands. As we can see the volumes of hair are driven by the lifted guide hair to create a new hair motion. Please see supplementary material for video results.}
\end{figure}

One major application that is enabled by our neural volumetric scene representation is novel view synthesis as we have shown in Sec.~\ref{sec:nvs}. Our neural volumetric representation is also animatable with a sparse driving signals like guide hair strands. Given that we have explicitly modeled hair in the form of guide strands, our method 
%
allows modifying the guide hairs directly.
In Fig.~\ref{fig:edit_hair}, we show four snapshots of different configurations of hair positions. Please see more results and details in the supplementary material.

There are several limitations of our work which we plan to address in the future: 
%
%
1) Our method requires the help from artist to prepare guide hair at the first frame and some flyaway hair might be excluded.
2) We currently do not consider physics based interactions between hair and other objects like the shoulder or the chair. 
3) Although we achieved certain level of disentanglement between hair and other objects without any human labeling, it is still not perfect. 
We only showed results on blonde hair which could be better distinguished from a dark background. Our method might be limited by other hairstyles.
%
Future directions like incorporating a physics aware module or leveraging additional supervision from semantic information for disentanglement could be interesting.

%% file: discuss.tex
\section{Discussion}


In this paper, we present a hybrid neural volumetric representation for hair dynamic performance capture. Our representation leverages the efficiency of guide hair representation in hair simulation by attaching volumetric primitives to them as well as the high DoF of volumetric representation. 
With both hair tracking and 3D scene flow refinement, our model enjoys better temporal consistency.
%
%
We empirically show that our method generates sharper and higher quality results on hair and our method achieves better generalization. 
Our model also supports multiple applications like drivable animation and hair editing.

%% file: appendix.tex
\section{Appendix}

\subsection{Dataset and Capture system}

We use a multi-camera system with around 100 synchronized color cameras that produces $2048\times1334$ resolution images at 30 Hz. The cameras are focused at the center of the capture system and distributed spherically at a distance of one meter to provide as many viewpoints as possible. Camera intrinsics and extrinsics are calibrated in an offline process. We captured three sequences of different hair styles and hair motions. In the first sequence, we have one actor with a short high pony tail performing nodding and rotating. In the second sequence, we have one actor with a curly long releasing style hair and leaning her head towards four directions(left, right, up and down) and rotating. In the third sequence, we have one actor with a long high pony tail performing nodding and rotating.

\subsection{Baselines}

We compare against several volume-based or implicit function based baseline methods~\cite{steve_mvp, tretschk2021nrnerf, li2021nsff} for spatio-temporal modeling. 


\noindent\textbf{MVP\cite{steve_mvp}} presents an efficient 4D representation for dynamic scenes with humans which is capable of doing animation and novel view synthesis. It combines explicitly tracked head mesh with volumetric primitives to model the human appearance and geometry with better completeness. The volumetric primitives can be aligned onto an unwrapped 2D UV-map from a tracked head mesh and can be regressed from a 2D convolutional neural network that leverages shared spatially computation. Similar to Neural Volumes~\cite{steve_nvs}, a differentiable volumetric ray marching algorithm is designed to render 2D rgb images on MVP in real time. We use $N_{p}=4096$ volumetric primitives with a voxel resolution $8\times8\times8$ on each sequence with a ray marching step size around $dt=1mm$. We use a global latent size of $256$.

\noindent\textbf{Non-rigid NeRF\cite{tretschk2021nrnerf}} presents an implicit function based representation for dynamic scene reconstruction and novel view synthesis based on NeRF~\cite{mildenhall2020nerf}. It utilizes a hierarchical model by disentangling a dynamic scene into a canonical frame NeRF and its corresponding deformation field which is parameterized by another MLP. In our experiments, we use 128 sampling points for both coarse and fine level sampling. We use the original implementation from the authors ~\href{https://github.com/facebookresearch/nonrigid_nerf}{here}. We train different models for each sequences and each model is trained for at least 300k iterations until convergence.

\noindent\textbf{NSFF\cite{li2021nsff}} is another implicit function based representation for dynamic scenes that is also based on  NeRF~\cite{mildenhall2020nerf}. It learns a per-frame NeRF that is additionally conditioned on the time index. It brings optical flow as additional supervision and learns a 3D scene flow in parallel with the per-frame NeRF for enforcing temporal consistency. NSFF is able to perform both spatial and temporal interpolation on a given video sequence. We use a setting of 256 sampling points in our experiments, using~\cite{kroeger2016disof} as a substitute for generating optical flow. We use the original implementation from the authors ~\href{https://github.com/zl548/Neural-Scene-Flow-Fields}{here}. We train different models for each sequences and each model is trained for at least 300k iterations until convergence.

\subsection{Training Details}

For both tracking optimization and HVH training, We deploy Adam~\cite{kingma2014adam} for optimization. For hair tracking, we use a learning rate of $1$. We set the weighting coefficients of each losses as $\omega_{hdir}=3$, $\omega_{hpos}=1$, $\omega_{len}=3$, $\omega_{tang}=3$ and $\omega_{cur}=1e4$. For each time step, 100 iterations are taken for optimization to solve the possible hair strands at next frame out. For HVH, we set weighting parameters for each objective as $\lambda_{flow}=1$. $\lambda_{geo}=0.1$, $\lambda_{vol}=0.01$, $\lambda_{cub}=0.01$ and $\lambda_{KL}=0.001$. All models are trained with approximately 100-150k iterations. We use a latent code size of 256 and per-strand hair code size of 256, raymarching step size around $dt=1mm$ and around $N_{p}=5500$ volumetric primitives with a voxel resolution $8\times8\times8$ for each sequence depending on the number of guide hairs. For each sequence, we have roughly 30 strands for guide hair and we sample 50 points on each strands.

\subsection{Novel View Synthesis}

\begin{figure*}[htb]
\setlength\tabcolsep{0pt}
\renewcommand{\arraystretch}{0}
\centering
\begin{tabular}{cccccc}
 \textbf{\small Perframe NeRF} &
 \textbf{\small NSFF~\cite{li2021nsff}} &
 \textbf{\small Non-rigid NeRF~\cite{tretschk2021nrnerf}} &
 \textbf{\small MVP~\cite{steve_mvp}} & 
 \textbf{\small Ours} &
 \textbf{\small Ground Truth} \\
 \adjincludegraphics[width=0.162\textwidth, trim={0 0 0 0}, clip]{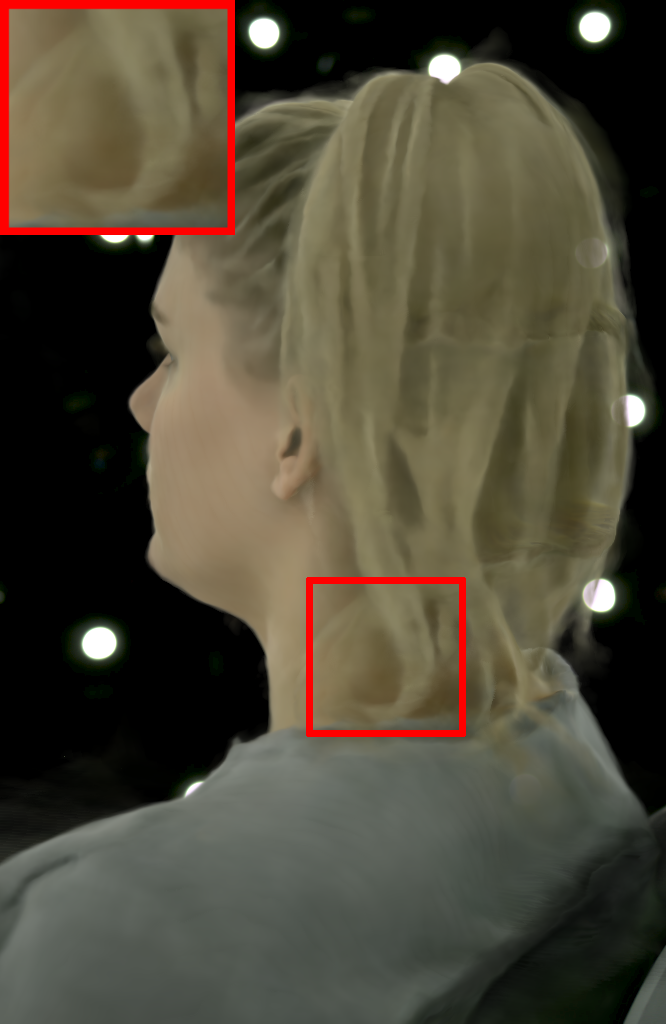} &
 \adjincludegraphics[width=0.162\textwidth, trim={0 0 0 0}, clip]{figs/nvs_compare/seq01_57/000057_nsff_comp_307_578_464_735_nw.png} &
 \adjincludegraphics[width=0.162\textwidth, trim={0 0 0 0}, clip]{figs/nvs_compare/seq01_57/000057_nrnerf_comp_307_578_464_735_nw.png} &
 \adjincludegraphics[width=0.162\textwidth, trim={0 0 0 0}, clip]{figs/nvs_compare/seq01_57/000057_mvp_comp_307_578_464_735_nw.png} &
 \adjincludegraphics[width=0.162\textwidth, trim={0 0 0 0}, clip]{figs/nvs_compare/seq01_57/000057_ours_comp_307_578_464_735_nw.png} &
 \adjincludegraphics[width=0.162\textwidth, trim={0 0 0 0}, clip]{figs/nvs_compare/seq01_57/000057_gt_comp_307_578_464_735_nw.png} \\
 \adjincludegraphics[width=0.162\textwidth, trim={0 0 0 0}, clip]{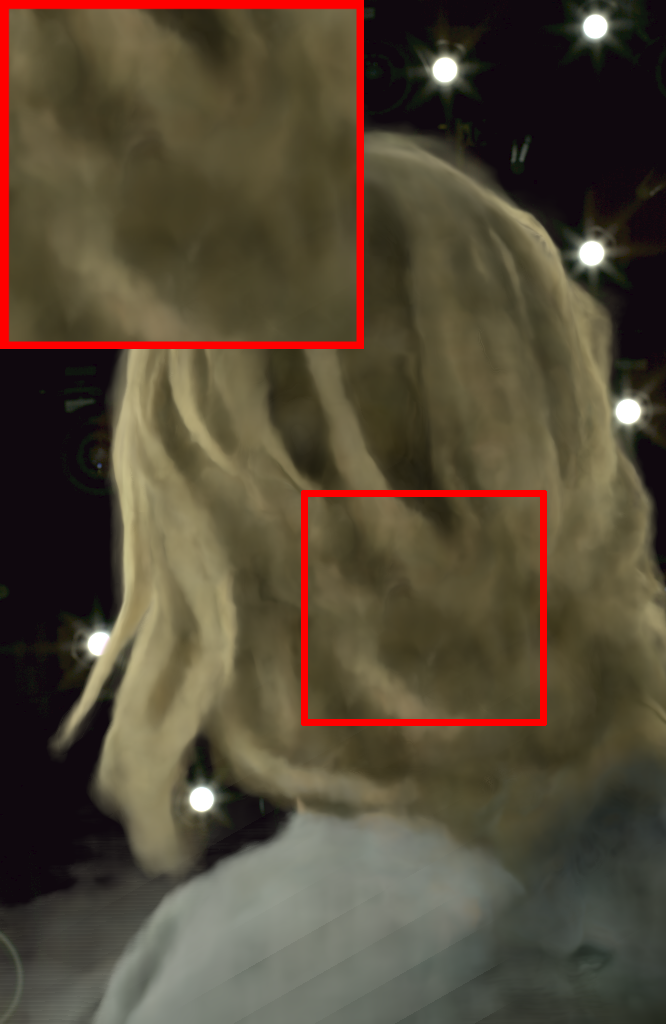} &
 \adjincludegraphics[width=0.162\textwidth, trim={0 0 0 0}, clip]{figs/nvs_compare/seq02_182/000182_nsff_comp_302_491_545_724_nw.png} &
 \adjincludegraphics[width=0.162\textwidth, trim={0 0 0 0}, clip]{figs/nvs_compare/seq02_182/000182_nrnerf2_comp_302_491_545_724_nw.png} &
 \adjincludegraphics[width=0.162\textwidth, trim={0 0 0 0}, clip]{figs/nvs_compare/seq02_182/000182_mvp_comp_302_491_545_724_nw.png} &
 \adjincludegraphics[width=0.162\textwidth, trim={0 0 0 0}, clip]{figs/nvs_compare/seq02_182/000182_ours_comp_302_491_545_724_nw.png} &
 \adjincludegraphics[width=0.162\textwidth, trim={0 0 0 0}, clip]{figs/nvs_compare/seq02_182/000182_gt_comp_302_491_545_724_nw.png} \\
 \adjincludegraphics[width=0.162\textwidth, trim={0 0 0 0}, clip]{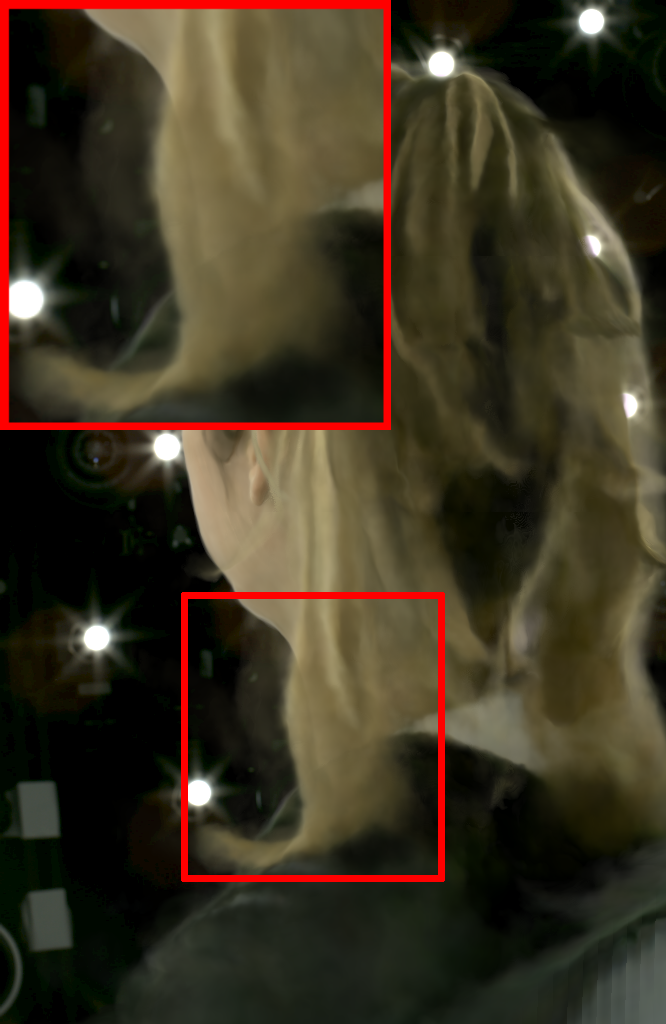} &
 \adjincludegraphics[width=0.162\textwidth, trim={0 0 0 0}, clip]{figs/nvs_compare/seq03_220/000220_nsff_comp_182_593_443_880_nw.png} &
 \adjincludegraphics[width=0.162\textwidth, trim={0 0 0 0}, clip]{figs/nvs_compare/seq03_220/000220_nrnerf_comp_182_593_443_880_nw.png} &
 \adjincludegraphics[width=0.162\textwidth, trim={0 0 0 0}, clip]{figs/nvs_compare/seq03_220/000220_mvp_comp_182_593_443_880_nw.png} &
 \adjincludegraphics[width=0.162\textwidth, trim={0 0 0 0}, clip]{figs/nvs_compare/seq03_220/000220_ours_comp_182_593_443_880_nw.png} &
 \adjincludegraphics[width=0.162\textwidth, trim={0 0 0 0}, clip]{figs/nvs_compare/seq03_220/000220_gt_comp_182_593_443_880_nw.png}
\end{tabular}
\caption{\label{fig:nvs_compare_apdx}\textbf{Comparison on novel view synthesis between different methods.}}
\end{figure*}

We show a larger version of comparison figure between different methods in Figure~\ref{fig:nvs_compare_apdx}. For completeness, we also include visualizations from a perframe NeRF model which takes a perframe temporal code as input liker non-rigid NeRF~\cite{tretschk2021nrnerf}.

\subsection{Ablation Studies}

\begin{figure*}[htb]
\setlength\tabcolsep{0pt}
\renewcommand{\arraystretch}{0}
\centering
\begin{tabular}{ccccc}
 \textbf{\small MVP} &
 \textbf{\small MVP w/ flow} &
 \textbf{\small Ours w/o flow} &
 \textbf{\small Ours} &
 \textbf{\small Ground Truth} \\
 \adjincludegraphics[width=0.2\textwidth, trim={0 0 0 0}, clip]{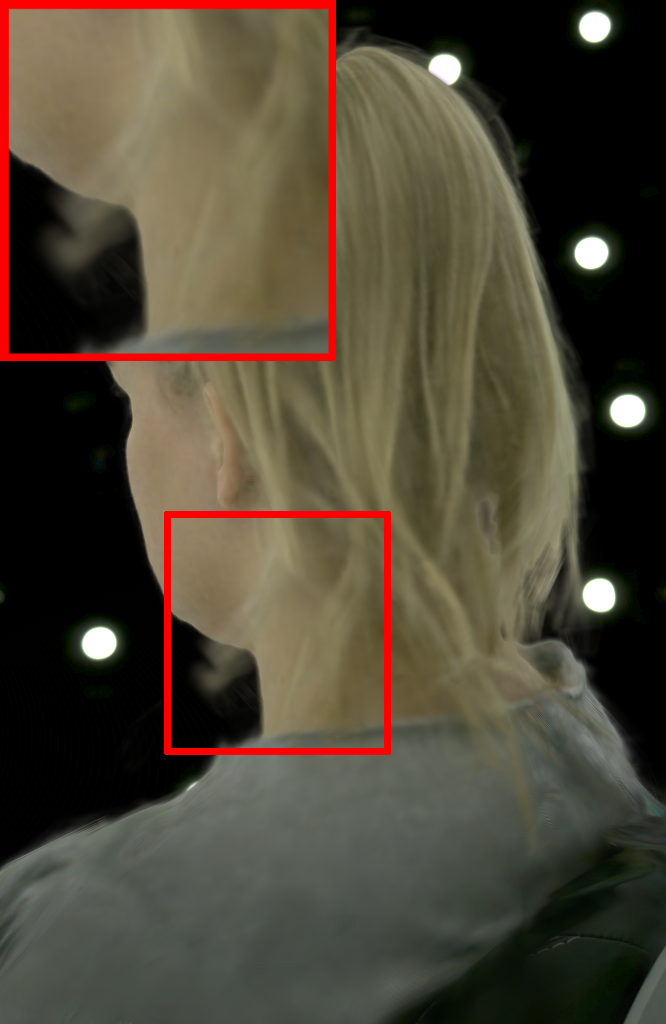} &
 \adjincludegraphics[width=0.2\textwidth, trim={0 0 0 0}, clip]{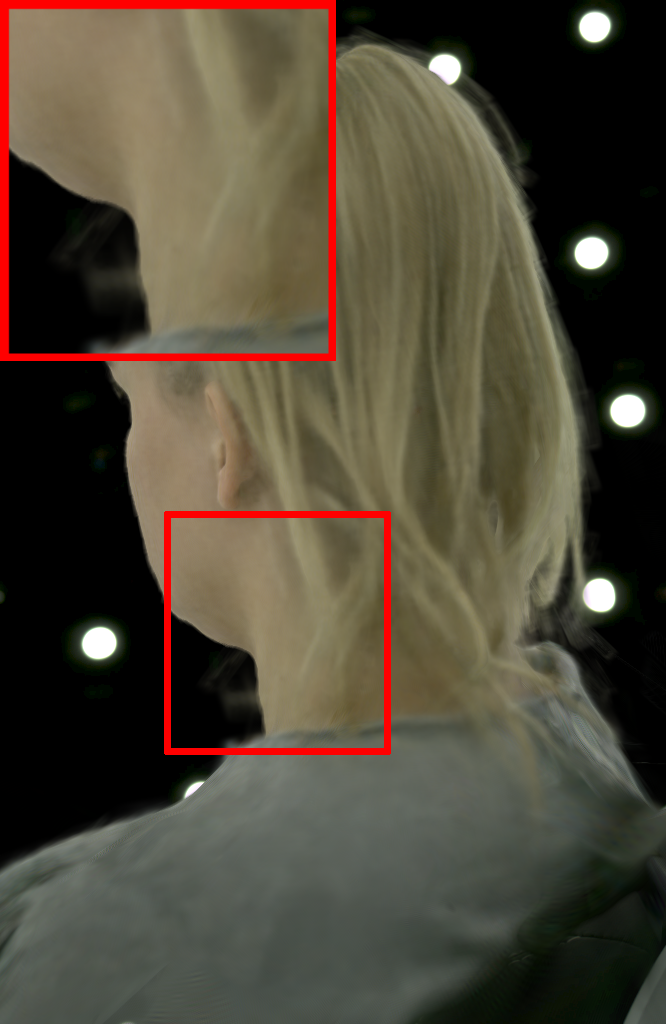} &
 \adjincludegraphics[width=0.2\textwidth, trim={0 0 0 0}, clip]{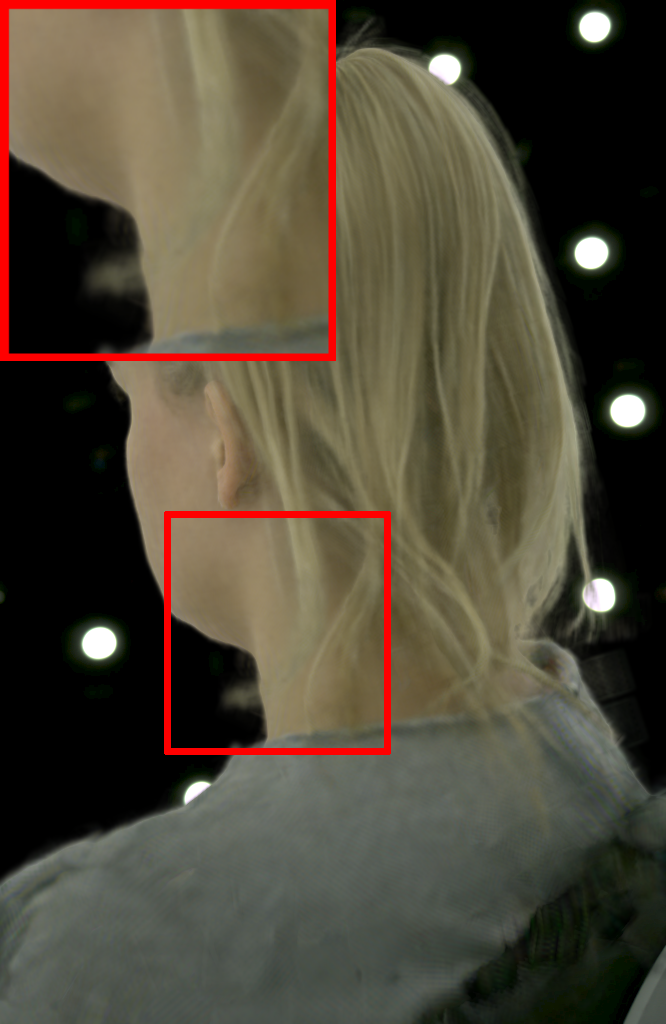} &
 \adjincludegraphics[width=0.2\textwidth, trim={0 0 0 0}, clip]{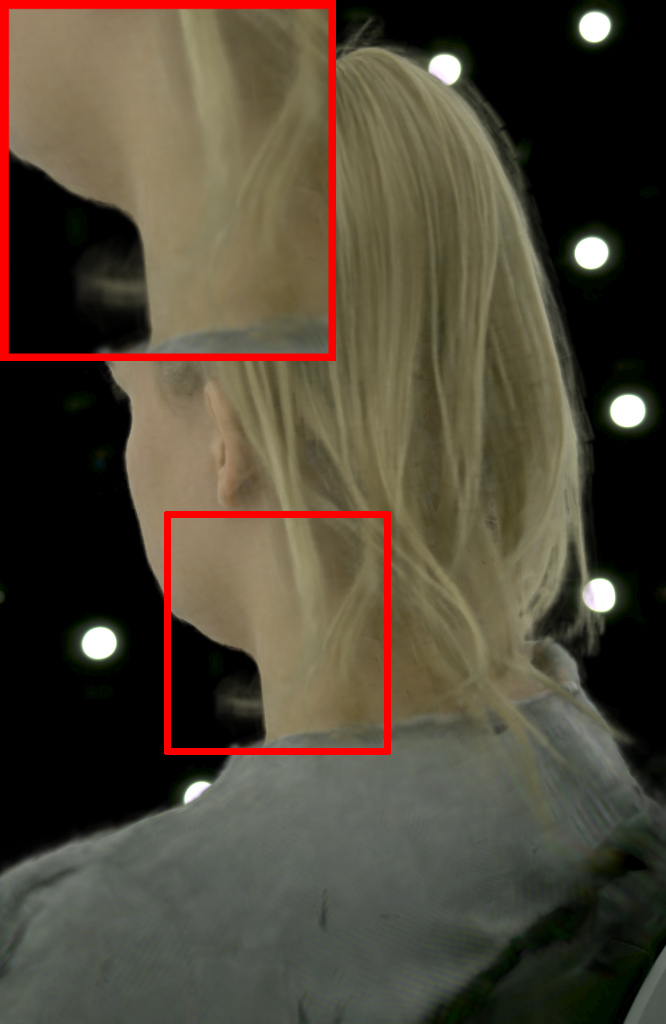} &
 \adjincludegraphics[width=0.2\textwidth, trim={0 0 0 0}, clip]{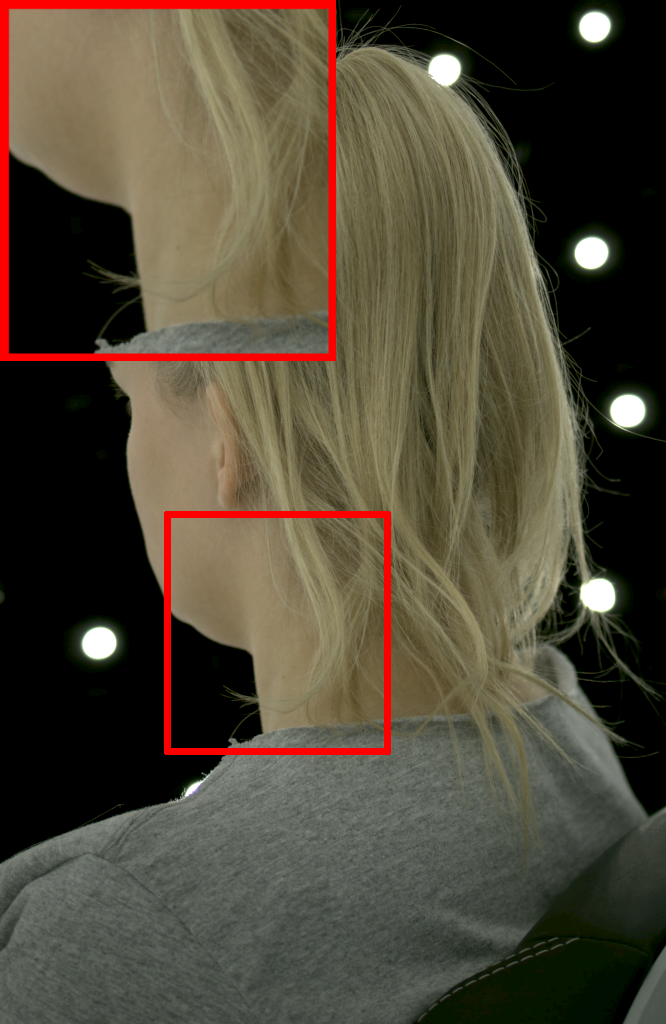} \\
 \adjincludegraphics[width=0.2\textwidth, trim={0 0 0 0}, clip]{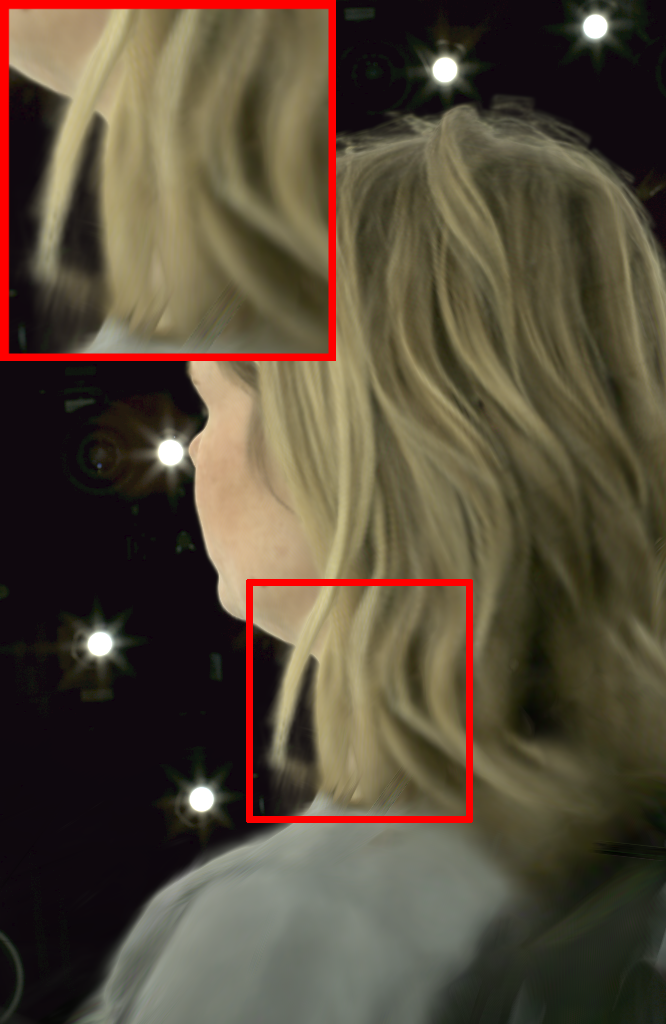} &
 \adjincludegraphics[width=0.2\textwidth, trim={0 0 0 0}, clip]{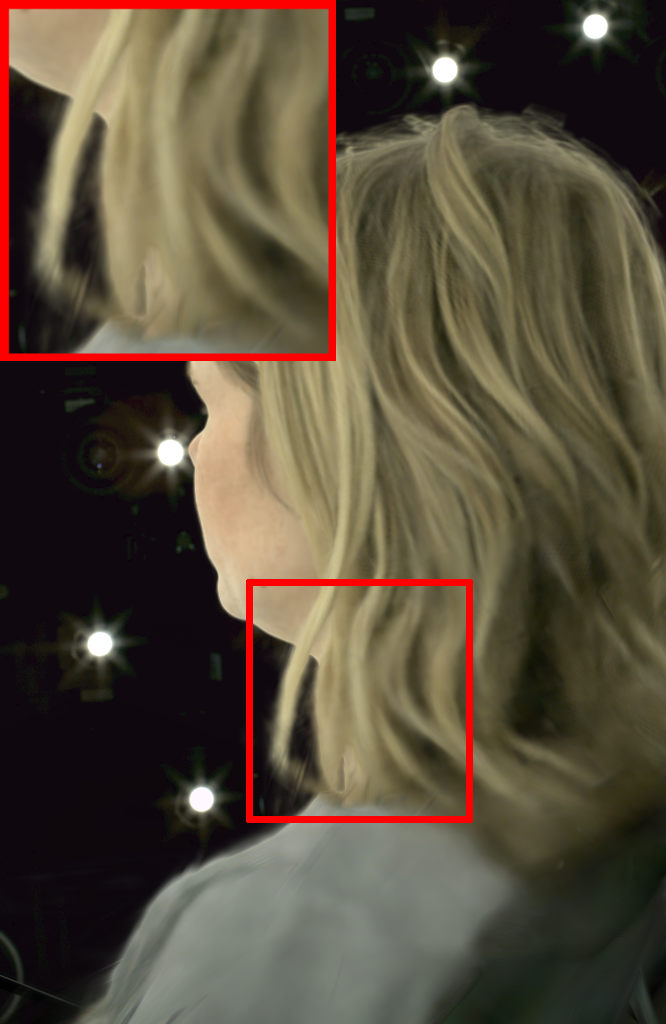} &
 \adjincludegraphics[width=0.2\textwidth, trim={0 0 0 0}, clip]{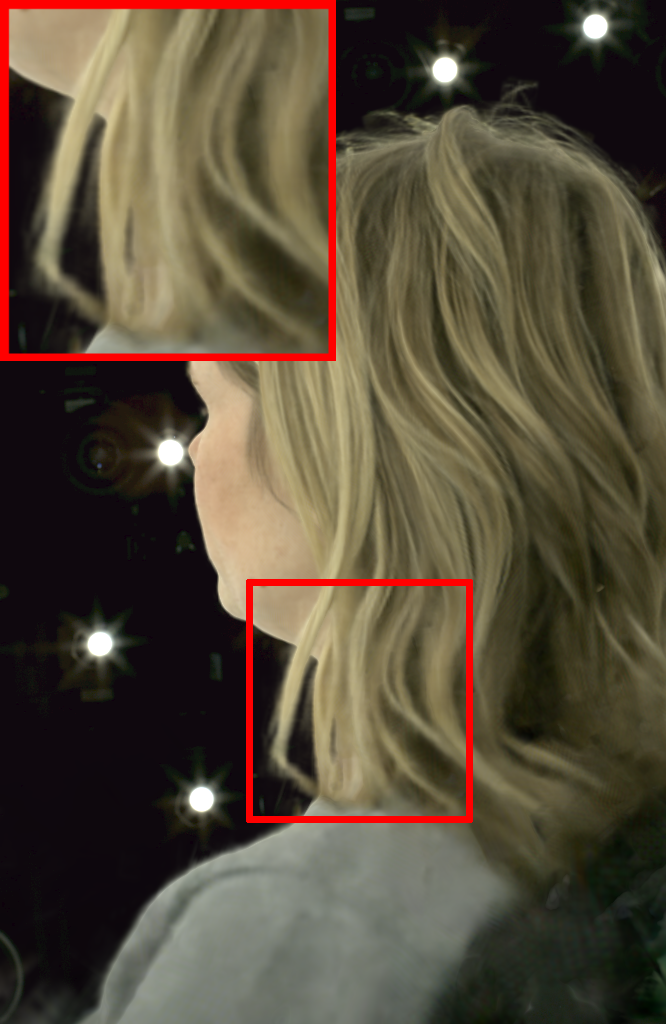} &
 \adjincludegraphics[width=0.2\textwidth, trim={0 0 0 0}, clip]{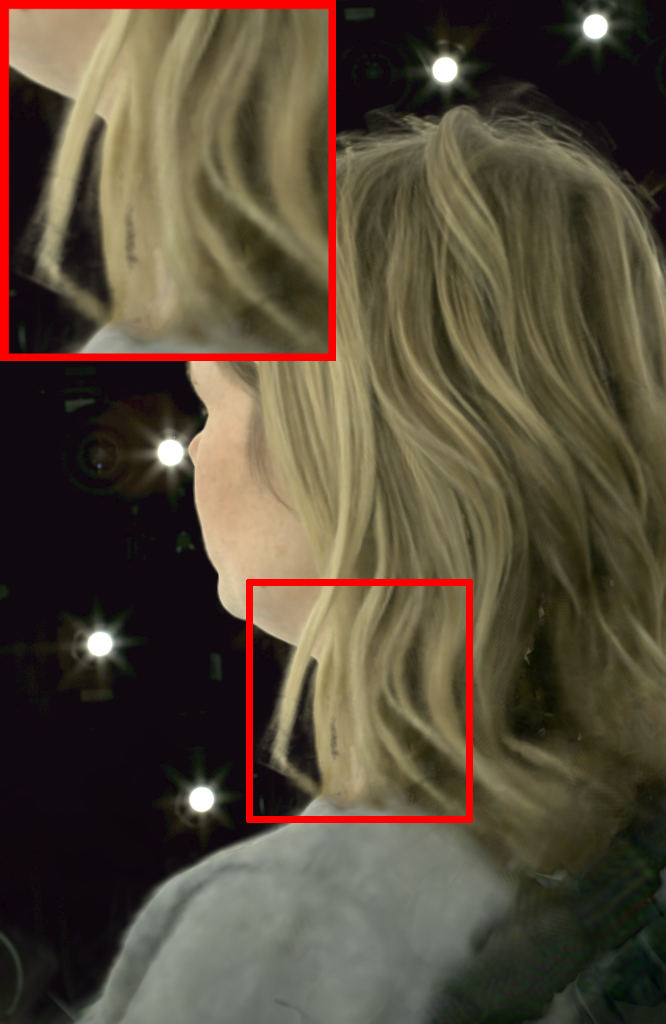} &
 \adjincludegraphics[width=0.2\textwidth, trim={0 0 0 0}, clip]{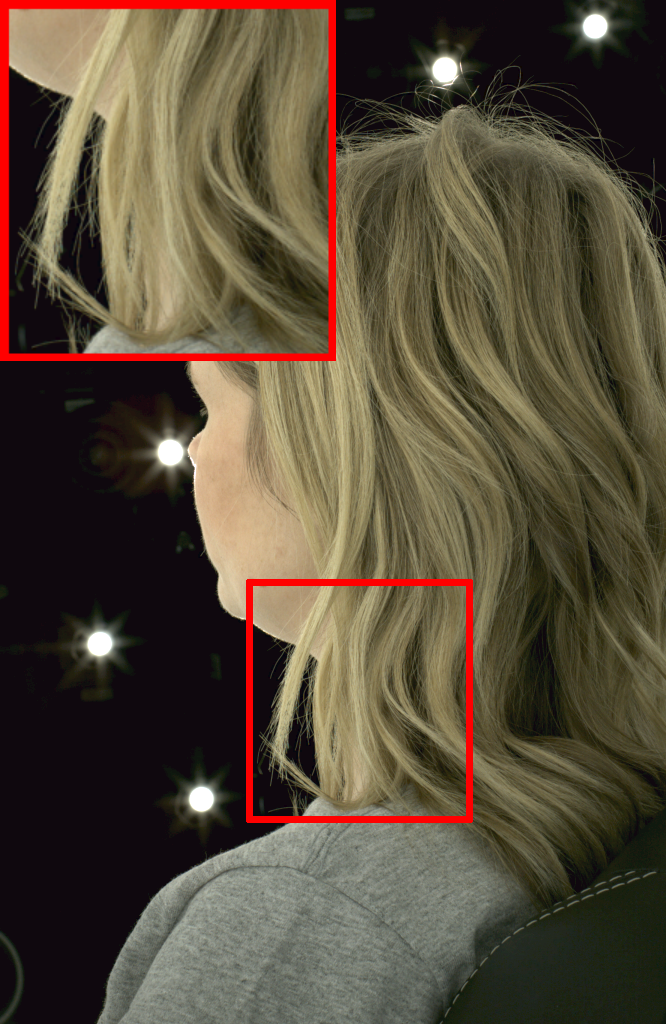} \\
 \adjincludegraphics[width=0.2\textwidth, trim={0 0 0 0}, clip]{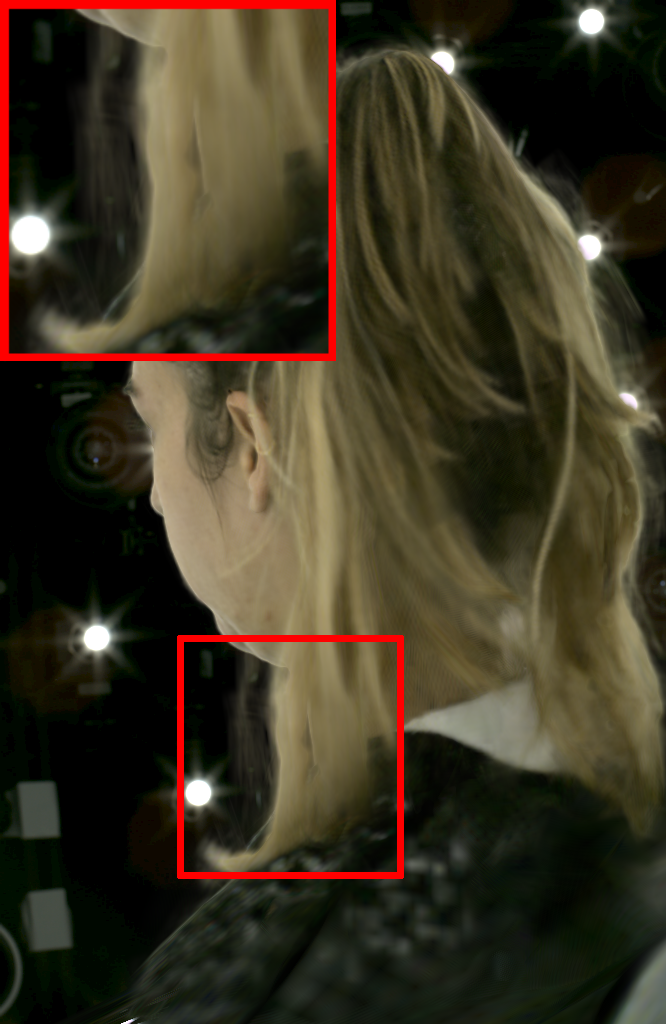} &
 \adjincludegraphics[width=0.2\textwidth, trim={0 0 0 0}, clip]{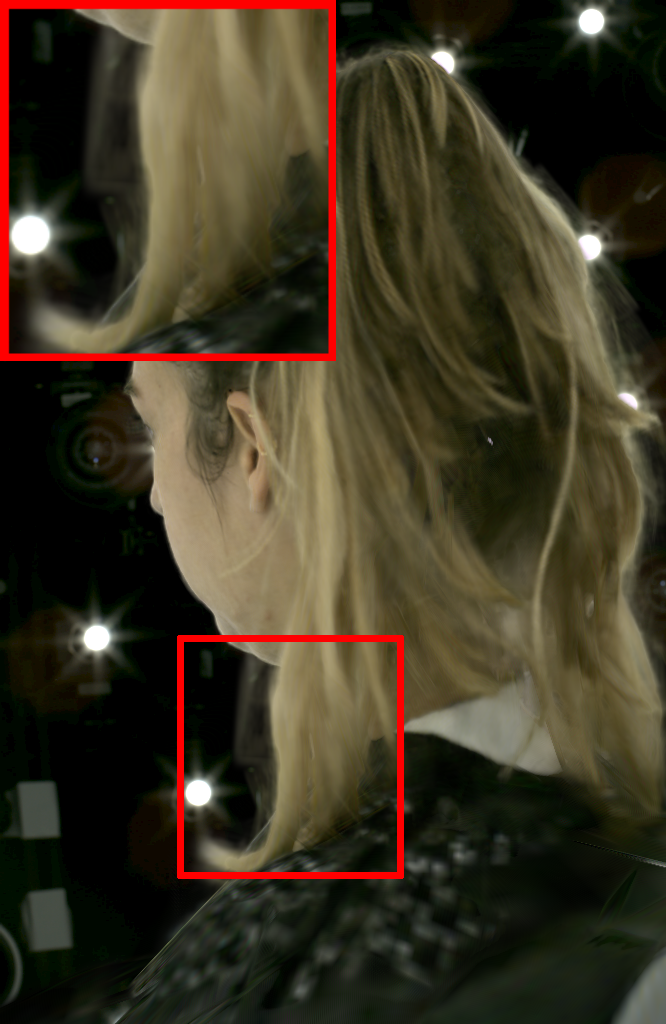} &
 \adjincludegraphics[width=0.2\textwidth, trim={0 0 0 0}, clip]{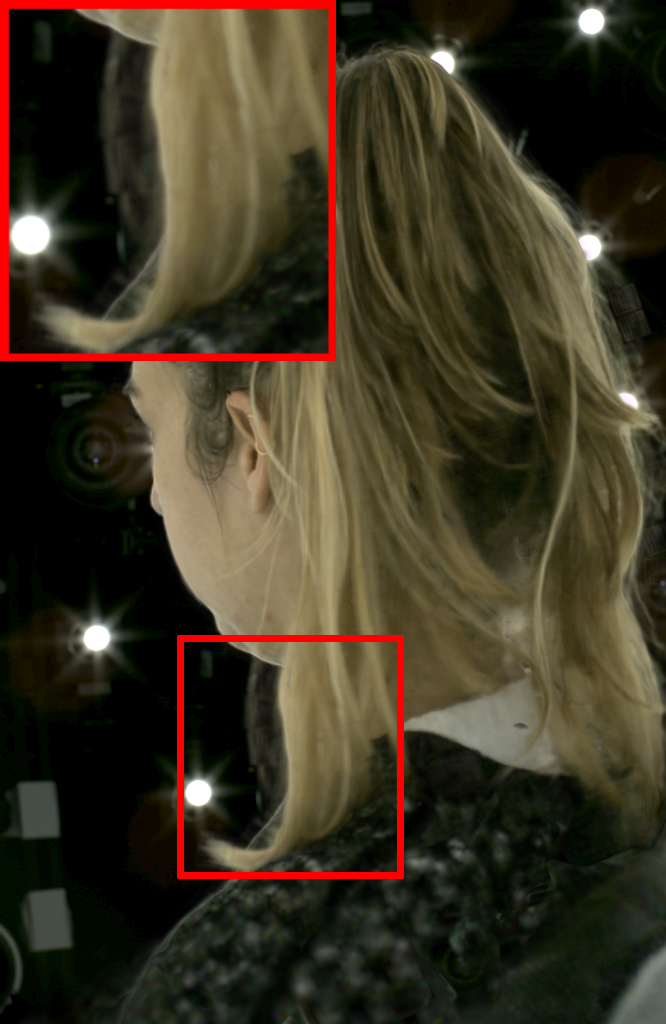} &
 \adjincludegraphics[width=0.2\textwidth, trim={0 0 0 0}, clip]{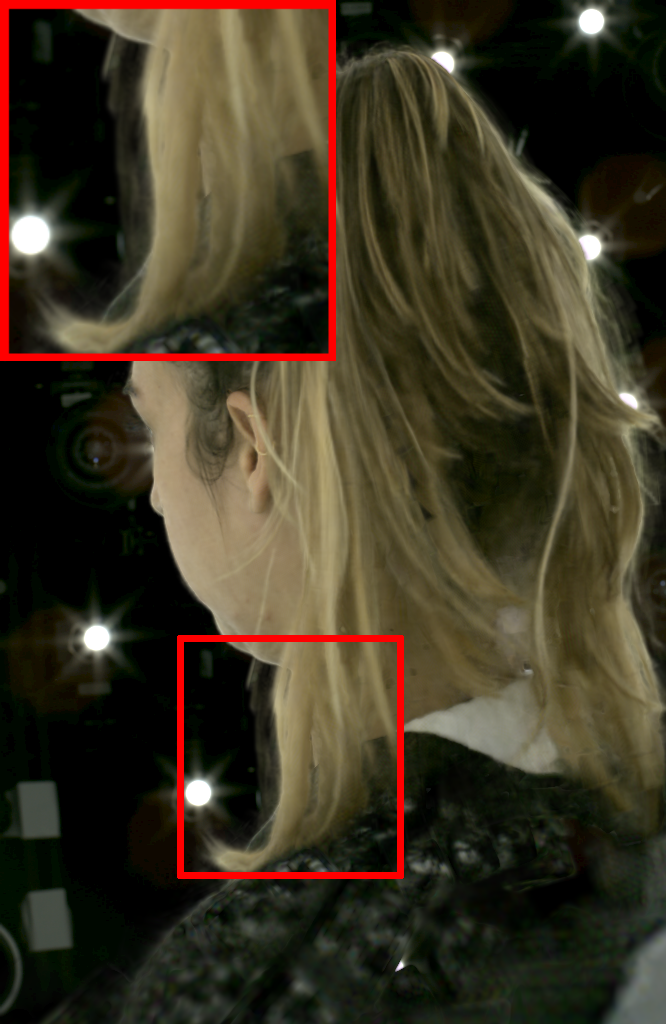} &
 \adjincludegraphics[width=0.2\textwidth, trim={0 0 0 0}, clip]{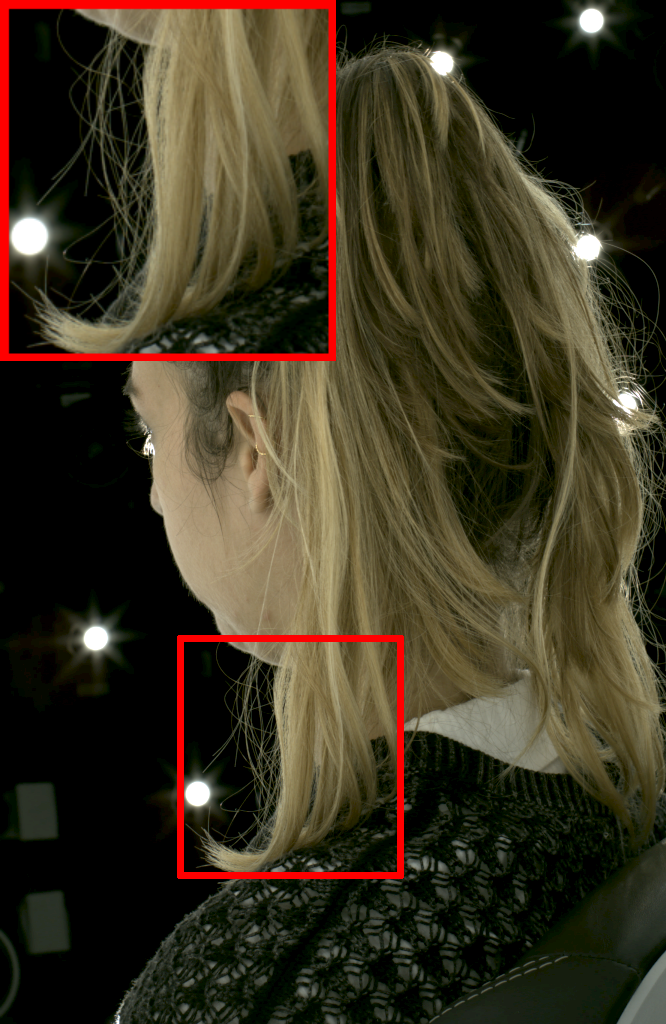}
\end{tabular}
\caption{\label{fig:nvs_abl_apx}\textbf{Ablation of temporal consistency.} We compare MVP~\cite{steve_mvp} and ours with different variations.}
\end{figure*}

\noindent\textbf{Temporal Consistency.} We show a bigger version of rendering results on unseen sequence in Figure~\ref{fig:nvs_abl_apx}.

\begin{figure}[h!]
\centering
\begin{tabular}{cc}
\textbf{\small Naive Decoder} &
\textbf{\small Our Decoder} \\
\adjincludegraphics[width=0.24\textwidth, trim={0 0 0 0}, clip]{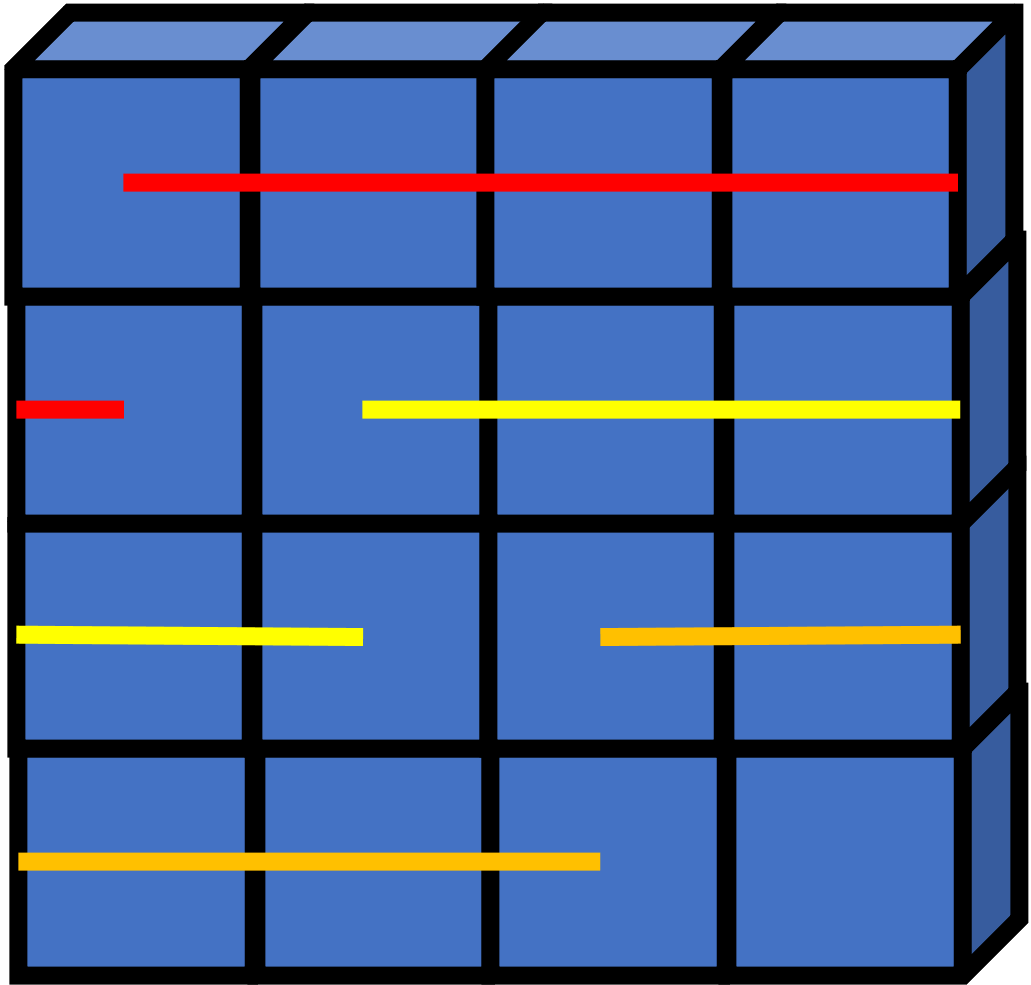} &
\adjincludegraphics[width=0.24\textwidth, trim={0 0 0 0}, clip]{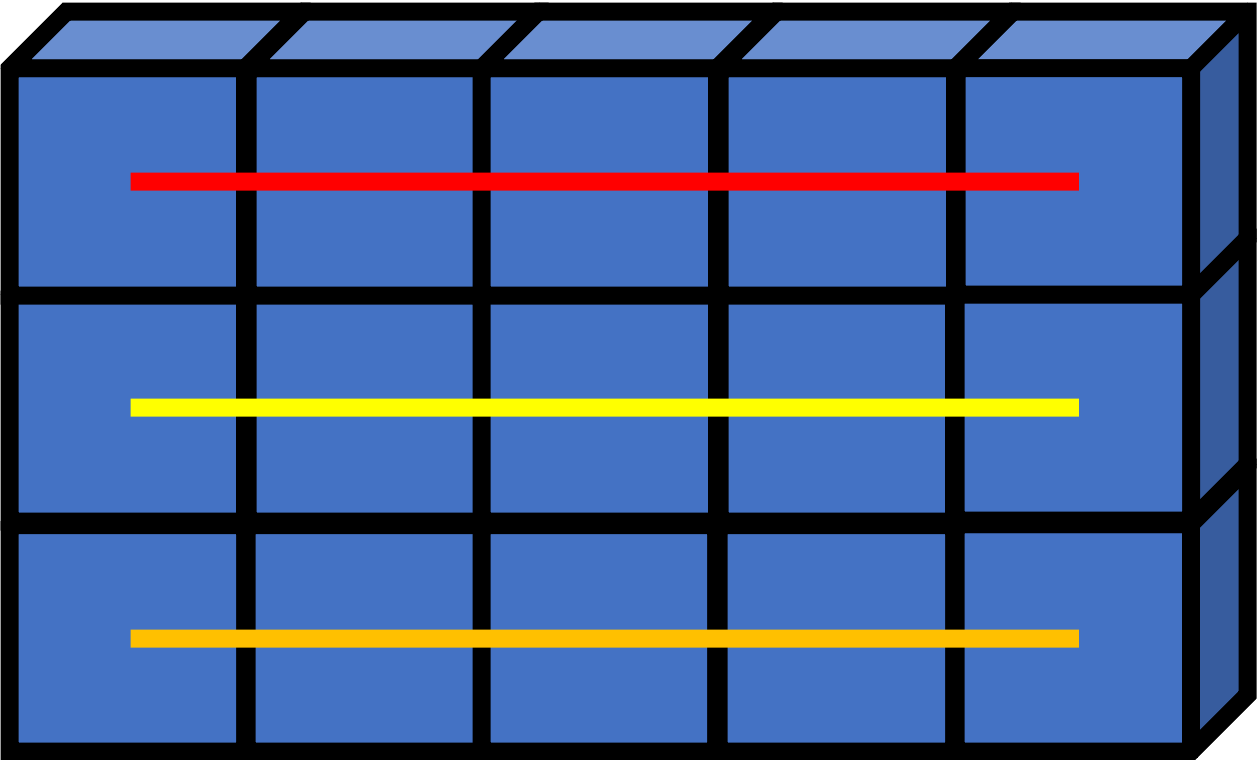}
\end{tabular}
\caption{\label{fig:apx_dec_layout}\textbf{Hair volumes layout.} We show the hair volume layout of both naive decoder and ours.}
\end{figure}

\noindent\textbf{Hair Decoder structure.} As part of the hair decoder ablation, we compare our method with a naive decoder that uses the same volume decoder as MVP~\cite{steve_mvp} for hair volumes. There are two major differences: 1) the naive decoder does not take the per-strand hair feature as input; 2) The design of the naive decoder does not take into account the hair specific structure where it regresses the same slab as for head tracked mesh and we take the first $N_{hair}$ volumes as the output. In this way, the naive decoder discards all intrinsic geometric structural information while doing convolutions in each layers. We show the hair volumes layout in Figure~\ref{fig:apx_dec_layout}. In the naive design, the hair strands are randomly squeezed into a square UV-map which could break the inner connections of each hair. In our design, we groom the hair strands into the their directions which could preserve the hair specific geometric structure. We compare different designs of decoder on Seq01. As in Table~\ref{tab:abl_dec}, our hair structure aware decoder produces a smaller image reconstruction error and better SSIM, a result of inductive bias of the designed hair decoder.

\begin{table}[h!]
\centering
\begin{tabular}{r|ccc|}
\multicolumn{1}{r|}{decoder} & \multicolumn{1}{c|}{MSE} & \multicolumn{1}{c|}{SSIM} & PSNR        \\ \hline
naive         & 45.68/75.15              & 0.9549/0.9220             & 31.83/29.54 \\
early fus.  & 43.75/71.08              & 0.9533/0.9259             & 31.97/29.82 \\
late fus.   & 41.89/65.96              & 0.9543/0.9280             & 32.17/30.09
\end{tabular}
\caption{\label{tab:abl_dec}\textbf{Decoder structure.} We compare different designs of the hair decoder. We report all metrics on both training and testing and we use a \/ to separate them where on the left are the results of novel synthesis on training sequence.}
\end{table}

\begin{figure}[htb]
    \centering
    \adjincludegraphics[width=0.45\textwidth, trim={0 0 0 {0.02\height}}, clip]{figs/hair_decoder.pdf} \\
    \adjincludegraphics[width=0.45\textwidth, trim={0 0 0 {0.02\height}}, clip]{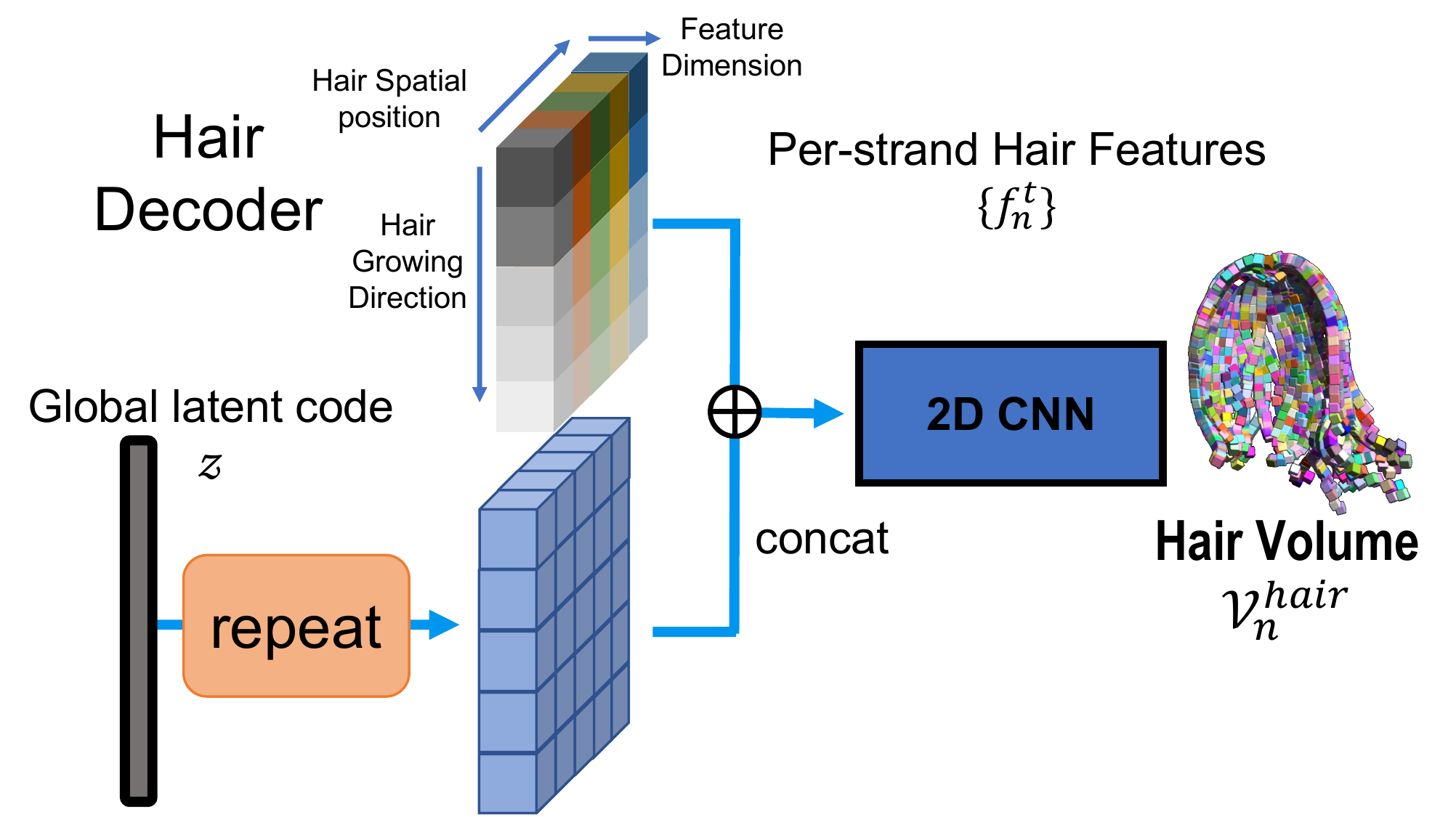}
    \caption{\label{fig:hair_dec_apd}\textbf{Architecture of the hair decoder.} We show late fusion on the top and early fusion on the bottom. The late fusion model first deconvolves the 1D global latent code into a 2D feature map and then concatenate it with the per-strand hair features. A 2D CNN is used afterwards to generate the hair volumes. The early fusion model first repeat the 1D global latent vector spatially and then concatenate the repeated feature map with per-strand hair features. The concatenated features are than fed into a deeper 2D CNN to generate the hair volumes.}
    \label{fig:hair_dec}
\end{figure}

We additionally compare two different designs of the hair decoder where we do late and early fusion of the per-strand hair feature and the global latent feature. We show two different designs in Figure~\ref{fig:hair_dec_apd}. Table~\ref{tab:abl_dec} shows that the late fusion model performs better than early fusion model. This could be because the late fusion model transfers the 1d global latent code into a spatially varying feature tensor which is a more expressive form of feature representation.

\subsection{Visualization of Flow}

\begin{figure}[h!]
\setlength\tabcolsep{0pt}
\renewcommand{\arraystretch}{0}
\centering
\begin{tabular}{ccc}
 \textbf{\small Flow from ours} &
 \textbf{\small Flow from ~\cite{kroeger2016disof}} &
 \textbf{\small Ground truth} \\
 \adjincludegraphics[width=0.16\textwidth]{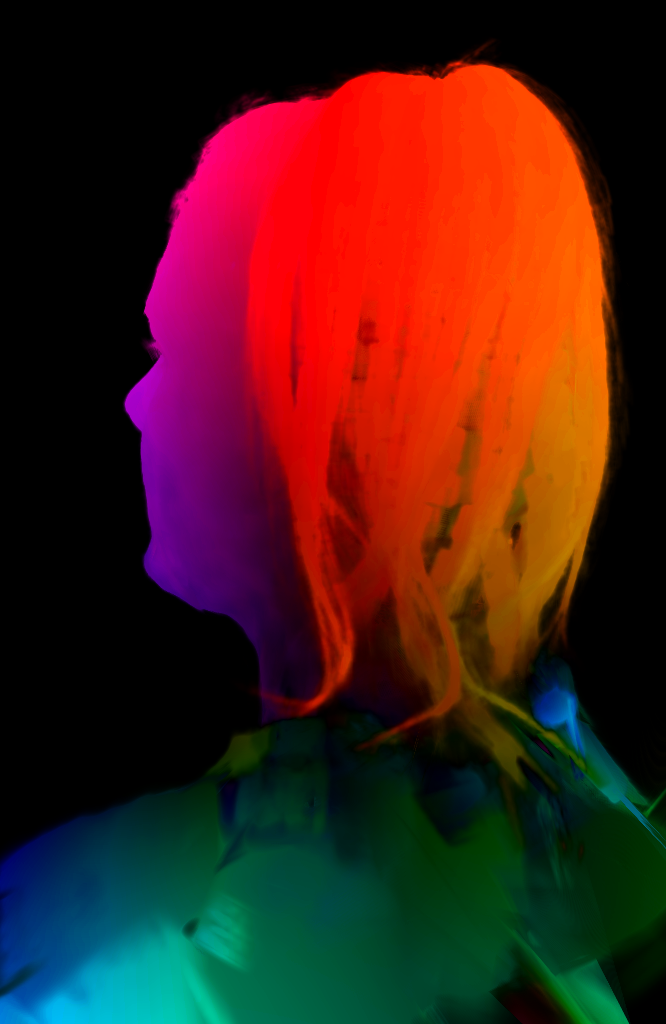} &
 \adjincludegraphics[width=0.16\textwidth]{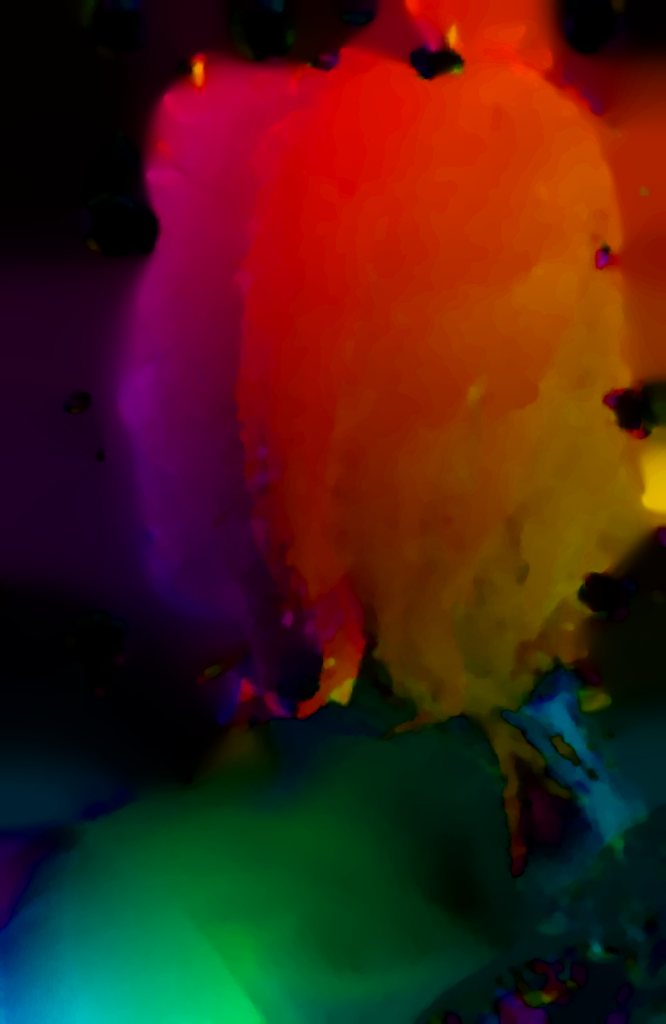} &
 \adjincludegraphics[width=0.16\textwidth]{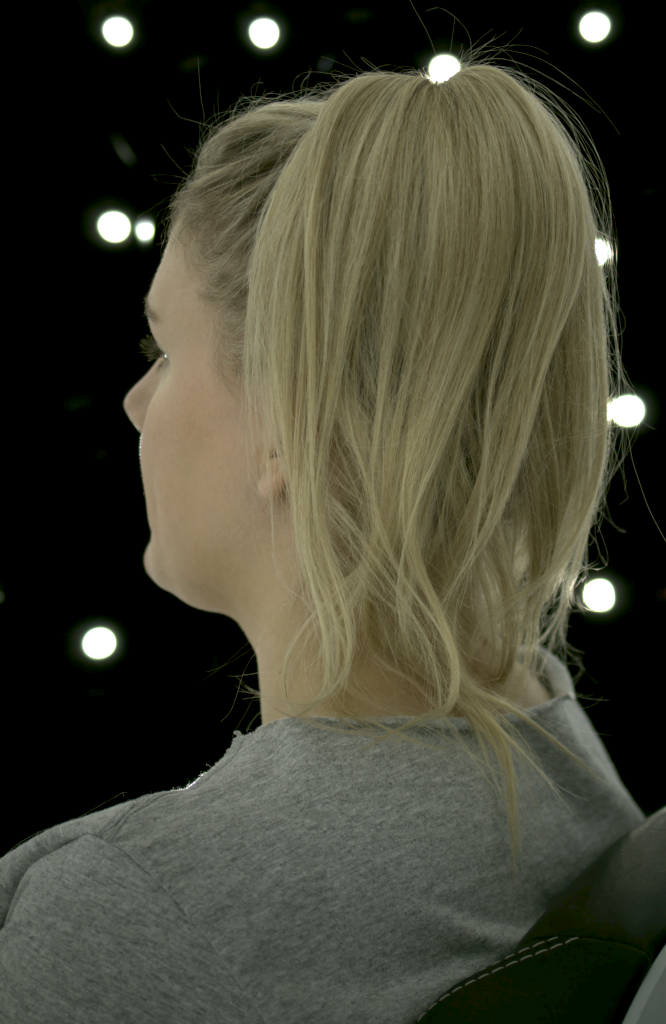} \\
\end{tabular}
\caption{\label{fig:iflow_vis}\textbf{Visualization of flow.} We show the rendered 3D scene flow into 2D flow in the first column and the openCV optical flow~\cite{kroeger2016disof} in the second column. The last column shows the ground truth image as reference.}
\end{figure}

Please see Figure~\ref{fig:iflow_vis} for a visualization of the rendered flow from our representation. Compared to the optical flow from ~\cite{kroeger2016disof}, our rendered 2D flow has less noise on the background. This is because that we only define our 3D scene flow on the volumetric primitives instead of the whole space. With the help of the coarse level geometry like the hair strands and head tracked mesh, the scene flow of most part of the empty space will naturally be zero. This could help us eliminate the noise from the background optical flow to certain degree.

\noindent\textbf{Run Time Analysis.} We report the rendering time of one iamge at resolution $1024\times667$ for each methods here. MVP~\cite{steve_mvp} takes 0.223s. Ours takes 0.254s. NSFF takes 28.68s. NRNeRF~\cite{tretschk2021nrnerf} takes 41.29s. All tests are conducted under a single Nvidia Tesla V100 GPU.

\subsection{Hair Tracking Analysis}
\begin{figure}[h!]
    \centering
    \includegraphics[width=0.5\textwidth]{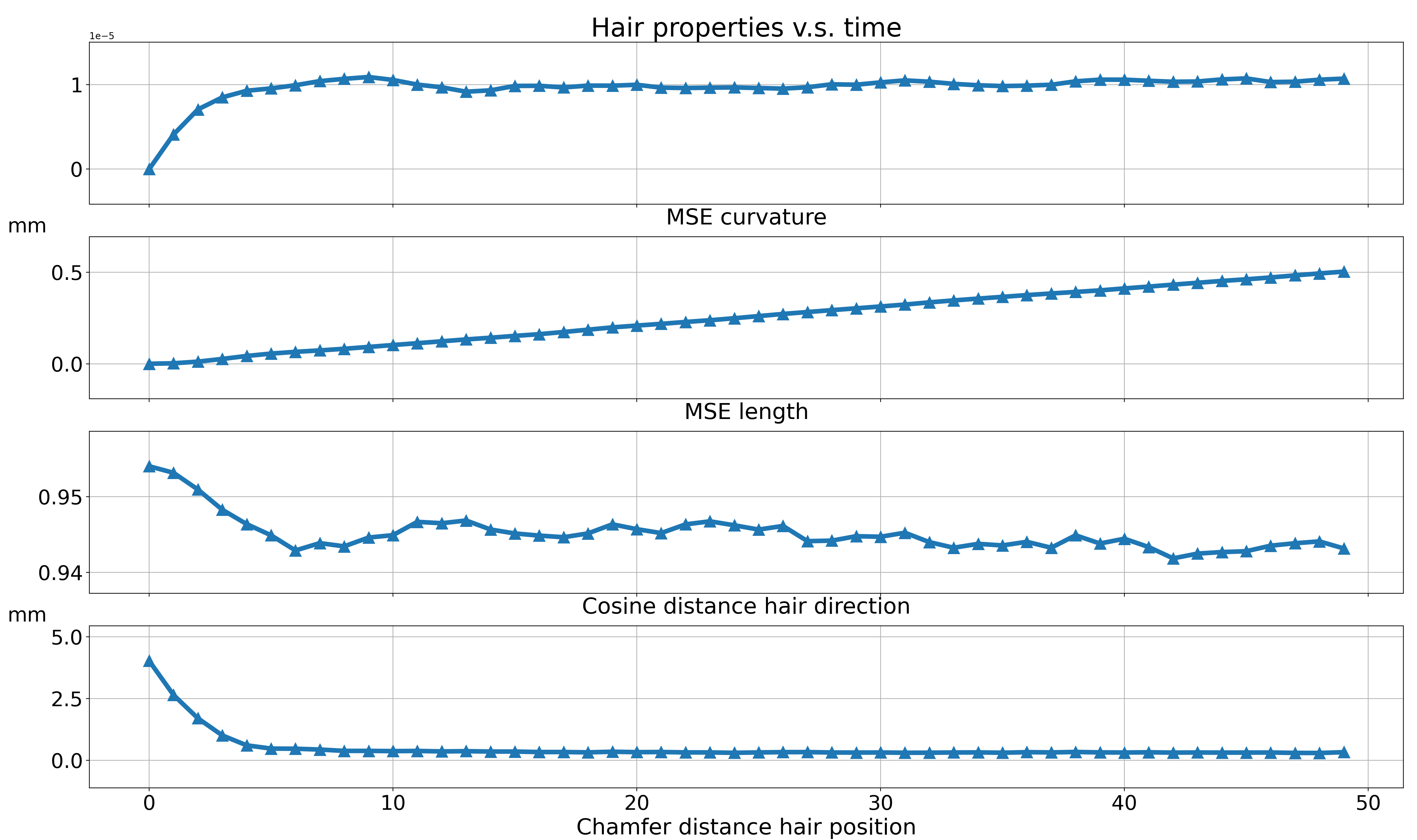}
    \caption{\label{fig:track_plot}\textbf{Plot of tracked hair properties v.s. time.} As we can see, the hair properties like length and curvature are not changing too much across time and hair Chamfer distance are relatively small.}
\end{figure}

In Figure~\ref{fig:track_plot}, we plot different hair properties over time. We report four different metrics describing how well the tracked hairs fit the per-frame reconstruction and how well it preserves its length and curvature. In the first two rows, we report the MSE between the tracked hair and the tracked hair at first frame in terms of curvature and length. In the last two rows, we report the cosine distance between the direction of each nodes on the tracked guide hair and the direction of its neighbor from the reconstruction and the Chamfer distance between the tracked guide hair nodes and the reconstruction. As we can see the length and curvature are relatively preserved across frames and the affinity between the per-frame reconstruction and the tracked guide hair is relatively high. 

\section{Video Results}

Please see all the video results at appended html page: $./video\_navigation/index.html$